%% file: main.tex
\documentclass[sigconf,nonacm]{acmart}

\usepackage{amsmath}
\usepackage{booktabs}
\usepackage{float}
\usepackage{graphicx}
\usepackage{subcaption}
\usepackage{tabularx}
\usepackage{bbm}

\renewcommand\footnotetextcopyrightpermission[1]{}

\title[EviRec]{EviRec: Continual Evidence Learning for Dual Cold-Start
POI Recommendation}

\author{Rongchao Xu}
\email{rx21a@fsu.edu}
\affiliation{
  \institution{Florida State University}
  \city{Tallahassee}
  \state{Florida}
  \country{USA}
}

\author{Lin Jiang}
\email{lj23d@fsu.edu}
\affiliation{
  \institution{Florida State University}
  \city{Tallahassee}
  \state{Florida}
  \country{USA}
}

\author{Guang Wang}
\email{guang.wang@fsu.edu}
\affiliation{
  \institution{Florida State University}
  \city{Tallahassee}
  \state{Florida}
  \country{USA}
}

\begin{document}

\input{sections/abstract}


\maketitle

\input{sections/introNew}

\input{sections/problemNew}

\input{sections/dataanalysis}

\input{sections/methodNew}

\input{sections/newEvaluation}

\input{sections/related_work}

\input{sections/conclusion}

\bibliographystyle{ACM-Reference-Format}
\bibliography{main}

\clearpage
\appendix

\input{sections/appendix_benchmark_provenance}
\input{sections/appendix_metric_definitions}

\input{sections/appendix_current_experimental_boundary}

\end{document}

%% file: sections/abstract.tex
\begin{abstract}
Point-of-Interest (POI) recommendation is a core task in location-based services, yet most existing methods assume a fixed user population and POI catalog. Through a large-scale data-driven analysis of 10 U.S. cities, we identify substantial POI churn, user turnover, category drift, and decay in static POI memory, motivating the study of continual dual cold-start POI recommendation. To address this setting, we propose \textbf{EviRec}, a continual evidence-learning framework that estimates how much historical evidence should be trusted separately for each candidate POI. EviRec scores each visible candidate from three complementary views: a matching view based on the user's recent mobility profile, a transition-memory view that captures repeated mobility routines, and a lifecycle view that reflects candidate maturity. Because a near-zero transition score may indicate either irrelevance or insufficient observation, EviRec qualifies the evidence using each candidate's observation state and applies a reliability gate to adaptively route between transition-memory and lifecycle evidence. We evaluate EviRec on a full-year, five-city POI check-in dataset containing more than 30,000 users and 684,200 trajectories. Experimental results show that EviRec consistently outperforms state-of-the-art baselines, with the largest gains concentrated on cold-start queries. In particular, EviRec improves NDCG@10 by 20.4\% on Dual-New cases over the strongest baseline. In-depth analyses further confirm that these gains arise primarily from candidate-specific reliability gating while largely preserving previously learned mobility routines.

\end{abstract}

%% file: sections/introNew.tex
\section{Introduction}

% Next Point-of-Interest (POI) recommendation aims to predict and provide personalized suggestions for a user's next visit based on their historical mobility behavior and contextual information. As a fundamental task in location-based services, it supports a wide range of real-world applications, including local search, venue recommendation, trip planning~\cite{xu2026geogen,xu2026synhat}, mobile advertising, and urban service discovery~\cite{yu2025uqgnn,yu2026trustenergy,shen2025learning,10446624}. Hence, it has attracted substantial attention from both academia and industry.

Next Point-of-Interest (POI) recommendation aims to predict a user's next visit and provide personalized suggestions based on their historical mobility behavior and contextual information. As a fundamental task in location-based services, it supports a wide range of real-world applications, including local search~\cite{lane2010hapori,bennett2011inferring}, venue recommendation~\cite{li2024large,wang2025gnprsid}, trip planning~\cite{xu2026geogen,xu2026synhat}, and mobile advertising~\cite{rafieian2020targeting,bernritter2021behaviorally}. Hence, it has attracted substantial attention from both academia and industry.

Despite substantial progress in POI recommendation, existing methods~\cite{tan2016gru4rec, kang2018sasrec, sun2019bert4rec, luo2021stan, yang2023getnext, wang2024bigsl, lei2025cagnn} are typically developed and evaluated under a largely static user--POI interaction space and largely overlook the inherently dynamic and evolving nature of urban environments. However, in real-world settings, new POIs continually open, existing POIs close or become inactive, new users come to a city, and previously active users may also leave. These changes continually reshape the available user--POI interaction space and reduce the reliability of historical mobility patterns. Consequently, existing methods provide limited support for newly observed users and POIs and do not adequately capture how recommendation systems should adapt as urban environments evolve over time.

Motivated by this gap, we study \emph{continual dual cold-start POI recommendation}, where the model considers both newly observed POIs and first-time users while continually updating as new mobility data arrive. This setting presents two key challenges, each supported by our data-driven analysis across 10 cities and 90 city-month observations. First, \textbf{historical mobility evidence is unevenly reliable}. Repeated transitions involving well-observed POIs provide strong signals, whereas newly observed POIs have little transition history and first-time users offer limited personalized evidence. In our data, newly observed POIs account for 5.66\% of visits on average and up to 20.20\% in the most dynamic city-month, while newly observed users contribute 36.22\% of visits on average. Moreover, the coverage of POIs observed during training declines from 90.22\% one month later to 76.24\% after eight months, showing that even previously reliable evidence becomes less representative over time. Second, \textbf{continual updates should capture emerging mobility patterns without overwriting useful routines learned from earlier periods}. On average, 13.57\% of POIs active in one month become inactive in the next, while the remaining well-observed POIs continue to support recurring mobility routines. 

To address these challenges, we propose \textbf{EviRec}, a continual evidence learning framework for dual cold-start POI recommendation. Its key idea is to estimate, for each candidate POI, how much its historical evidence can be trusted, rather than assigning the same reliability to all candidates. For every POI visible at prediction time, EviRec constructs three complementary evidence views: a \textit{matching view} that captures its compatibility with the user's recent mobility history, a \textit{transition-memory view} that measures how frequently the user's recent POIs have previously transitioned to it, and a \textit{lifecycle view} that characterizes its maturity based on when it was first observed. A central difficulty is that a near-zero transition score is ambiguous: it may indicate that the candidate is irrelevant, or simply that the candidate is newly observed and has not accumulated sufficient transition evidence. EviRec resolves this ambiguity by conditioning the evidence views on the candidate's observation state, including whether any incoming transition has been observed and how long the candidate has been visible. A reliability gate is then designed to adaptively route each candidate between transition-memory and lifecycle evidence, favoring transition memory when historical support is sufficient and shifting toward lifecycle evidence when that support is sparse or unavailable. This candidate-specific routing addresses uneven evidence reliability, while continual updates allow both the transition memory and routing decisions to evolve as new interactions arrive without overwriting still-useful mobility routines.

We evaluate EviRec on real-world POI check-in data from five cities in 2025, comprising more than 30,000 users, 111,296 POIs, and 684,200 check-in trajectories. Extensive experiments show that EviRec substantially outperforms state-of-the-art methods in the dual cold-start setting, improving NDCG@10 by 20.4\% in the dual cold-start setting. Component ablations, stratified routing analysis, and an in-depth case study further demonstrate that this improvement primarily stems from reliability-gated routing between transition-memory and lifecycle evidence.

The key contributions of this paper are as follows:

\begin{itemize}
\item \textbf{Conceptually,} we formulate continual dual cold-start POI recommendation as a dynamic recommendation problem in which both users and candidate POIs may be newly observed and the model updates as mobility data arrive. A 10-city data-driven analysis reveals substantial POI churn, user turnover, and decay in historical POI coverage, motivating a time-aware evaluation protocol for this setting.

\item \textbf{Technically,} we propose \textbf{EviRec}, a continual evidence learning framework that estimates the reliability of historical evidence separately for each candidate POI. EviRec integrates matching evidence, transition-memory evidence, and lifecycle evidence, conditions them on the candidate's observation state, and uses a reliability gate to shift from transition memory to lifecycle evidence when historical support is sparse or unavailable.

\item \textbf{Empirically,} we conduct extensive experiments on a full-year, five-city dataset containing more than 30,000 users and 684,200 check-in trajectories. EviRec substantially outperforms state-of-the-art methods in the dual cold-start setting, improving NDCG@10 by 20.4\% on Dual-New cases. Code is available at \url{https://anonymous.4open.science/r/EviRec-0B61}.

\end{itemize}

%% file: sections/problemNew.tex
\vspace{-8pt}
\section{Problem Formulation}
\label{sec:Problem}
We study dual cold-start next-POI recommendation over a chronological data stream that is partitioned into ordered periods $t=\{1,\ldots,T\}$.
Let $\mathcal{D}_t$ denote the visits observed in period $t$, $\mathcal{U}_t$ the active users, and $\mathcal{P}_t$ the active POIs; let $\mathcal{P}=\bigcup_t\mathcal{P}_t$ denote the complete POI set observed across the stream.
Each visit is an event $e_i=(u_i,p_i,\mathbf{x}_i,t_i)$, where $u_i$ is the user, $p_i$ is the visited POI, $\mathbf{x}_i$ contains contextual features such as category, spatial, and temporal attributes, and $t_i$ is the period in which the visit occurs.
For a query event $i$, the user's history is
\vspace{-8pt}
\begin{equation}
    H_i=\{(p_j,\mathbf{x}_j,t_j):j<i,\ u_j=u_i\},
    \label{eq:user-history}
\end{equation}
that is, the sequence of visits made by this user prior to the target event.
Conditioned on $H_i$, the model ranks candidate POIs to recommend next POI $p_i$.
Unlike the standard fixed-catalog formulation, both sides of the problem are open: the set of recommendable POIs and the user pool both evolve over time.

Let $a(p)$ be the first period in which POI $p$ is observed.
At period $t$, the visible POI candidate set is
\begin{equation}
    \mathcal{C}_t=\{p\in\mathcal{P}:a(p)\leq t\}.
    \label{eq:candidate-set}
\end{equation}
A POI becomes recommendable at period $t$ only once it has appeared in the stream.
This rule prevents the model from accessing future POIs, while allowing newly observed POIs to enter the recommendation space as soon as they become visible.

Given a query $(u_i,H_i,\mathbf{x}_i,t_i)$, a model $f_\theta$ assigns a score to each visible POI,
\begin{equation}
    s_i(p)=f_\theta(u_i,H_i,p,\mathbf{x}_i,t_i),
    \quad p\in\mathcal{C}_{t_i},
    \label{eq:ranking-score}
\end{equation}
and ranks the visible candidates in $\mathcal{C}_{t_i}$ by this score to predict and recommend the next POI $p_i$.
The dual cold-start setting concerns queries whose target POI or querying user is new to the stream.

Let $b(u)$ be the first period in which user $u$ is observed, defined analogously to $a(p)$ for POIs.
We define three evaluation subsets over the query set,
        $\mathcal{Q}^{\mathrm{POI}} =\{i:a(p_i)=t_i\}$,
        $\mathcal{Q}^{\mathrm{User}} =\{i:b(u_i)=t_i\}$, and
        $\mathcal{Q}^{\mathrm{Dual}} =\mathcal{Q}^{\mathrm{POI}}\cap\mathcal{Q}^{\mathrm{User}}$,
which we refer to as POI-New, User-New, and Dual-New, respectively.
In a POI-New query, the target POI first becomes visible during period $t_i$ and therefore carries no transition evidence from earlier periods.
In a User-New query, the querying user first appears during period $t_i$ and thus has no personalized history from preceding periods.
A Dual-New query satisfies both conditions simultaneously.
These subsets isolate the dynamic recommendation cases in which transition and personalization evidence is scarcest.

%% file: sections/dataanalysis.tex
\vspace{-6pt}
\section{Data-Driven Insights and Motivation}

\begin{figure}[t]
    \centering
    \begin{subfigure}[t]{0.485\linewidth}
        \centering
        \includegraphics[width=\linewidth]{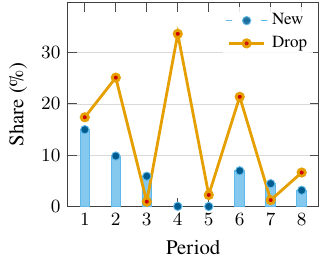}\vspace{-5pt}
        \caption{POI-New demand}
    \end{subfigure}
    \begin{subfigure}[t]{0.485\linewidth}
        \centering
        \includegraphics[width=\linewidth]{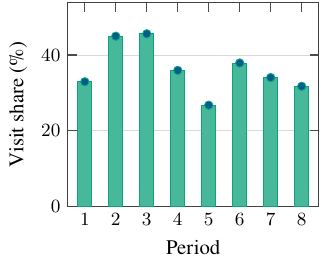}\vspace{-5pt}
        \caption{User-New demand}
    \end{subfigure}
    \vspace{0.25em}
    \begin{subfigure}[t]{0.485\linewidth}
        \centering
        \includegraphics[width=\linewidth]{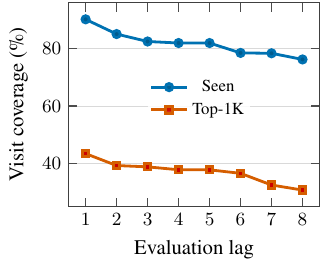}\vspace{-5pt}
        \caption{Memory decay}
    \end{subfigure}
    \begin{subfigure}[t]{0.485\linewidth}
        \centering
        \includegraphics[width=\linewidth]{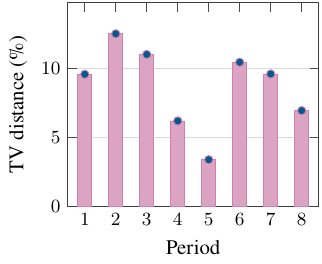}\vspace{-5pt}
        \caption{Category drift}
    \end{subfigure}\vspace{-10pt}
    \caption{Dynamic human activities. (a) POI-New visit share measures the fraction of visits whose target POI is first observed in the current period, and Drop measures the previously active POIs absent from the current period. (b) User-New visit share measures the fraction of visits from users first observed in the current period. (c) Visit coverage measures how much future demand is covered by POIs observed in a past training period. (d) Category drift is the adjacent-period total variation between category visit distributions.}
    \label{fig:data-anatomy}\vspace{-10pt}
    \Description{A four-panel figure showing POI catalog churn, user turnover, historical memory decay, and category demand drift.}
\end{figure}

We first conduct a data-driven analysis to show the motivation of our work. Our analysis covers 10 U.S. cities, 9 observed periods, 90 city-period aggregates, and 2.36 billion represented visits.
Let $\mathcal{D}_t^{\mathrm{newP}}$ denote the visits whose target POI first appears at $t$, and $\mathcal{D}_t^{\mathrm{newU}}$ the visits from users first observed at $t$.
We report visit-level shares because they directly measure the proportion of recommendation traffic affected by newly observed POIs or users.
A newly observed POI receiving many visits creates greater deployment pressure than multiple new POIs that are rarely visited.
The POI-New and User-New visit shares are
\begin{equation}
    \mathrm{share}_{\mathrm{newP}}(t)=\frac{|\mathcal{D}_t^{\mathrm{newP}}|}{|\mathcal{D}_t|},
    \quad
    \mathrm{share}_{\mathrm{newU}}(t)=\frac{|\mathcal{D}_t^{\mathrm{newU}}|}{|\mathcal{D}_t|}.
\end{equation}
The dropped-POI rate in Figure~\ref{fig:data-anatomy} is $|\mathcal{P}_{t-1}\setminus\mathcal{P}_t|/|\mathcal{P}_t|$, i.e., the number of previously active POIs that are no longer active, normalized by the size of the current active catalog.
We exclude the first observed period of each city from the churn statistics, since every POI and user is new by construction in that period.
Figure~\ref{fig:data-anatomy} shows that both sides of the recommendation problem change substantially over time.
The POI catalog evolves in both directions.
Newly observed POIs account for 5.66\% of visits on average after the first observed period and reach 20.20\% in the most dynamic city-period, while the rate of POIs dropped from the previous period averages 13.57\%.
The user population is even more dynamic, with newly observed users contributing 36.22\% of visits on average and 57.84\% in the most dynamic city-period.
An evaluation restricted to fixed users and POIs would therefore overlook a substantial portion of the deployed recommendation workload.
Historical evidence remains valuable, but its coverage decays as the mobility stream evolves.
For a training period $s$ and an evaluation period $t=s+k$, let $\mathcal{D}_t(S_s)$ be the visits in $t$ whose target POI belongs to a historical POI set $S_s$.
We define its visit coverage as
\begin{equation}
    \mathrm{coverage}(S_s,t) = \frac{|\mathcal{D}_t(S_s)|}{|\mathcal{D}_t|}.
\end{equation}
In Figure~\ref{fig:data-anatomy}(c), $S_s$ is either the set of all POIs seen in the training period or the top-1K most visited training POIs.
Coverage by POIs observed during training declines from 90.22\% at a one-period lag to 76.24\% at an eight-period lag, while coverage by the top-1K POIs declines from 43.42\% to 30.73\%.
Thus, historical memory remains too valuable to discard but becomes too incomplete to trust uniformly.
Figure~\ref{fig:data-anatomy}(d) further reveals category-level demand drift.
If $q_t(c)$ is the visit distribution over categories in period $t$, then adjacent-period total variation is
\begin{equation}
    \mathrm{TV}(q_t,q_{t+1}) = \frac{1}{2}\sum_c |q_t(c)-q_{t+1}(c)|.
\end{equation}
The mean adjacent-period TV is 0.0872, indicating that mobility demand shifts not only at the POI identity level but also at the behavioral category level.

These observations motivate two design choices.
First, the evaluation protocol should be time-aware, ensuring that future POIs cannot enter the candidate set before they are observed, and should report performance on the POI-New, User-New, and Dual-New subsets rather than relying solely on aggregate accuracy.
Second, the model should not rely exclusively on either historical memory or adaptive evidence for newly observed POIs.
Historical POIs continue to cover a substantial portion of future demand, whereas new POIs, new users, and shifting category preferences create cases with limited historical support.
These findings motivate EviRec's continual evidence learning design, in which a reliability gate preserves transition memory when it is informative and assigns greater weight to lifecycle evidence when historical support is weak.

%% file: sections/methodNew.tex
\vspace{-6pt}
\section{Methodology}

In an evolving POI catalog, the same near-zero transition score can have two fundamentally different meanings: the candidate POI may be irrelevant, or it may be newly visible and have insufficient observations to accumulate reliable transition evidence.
A fusion rule that fails to distinguish these cases will either systematically suppress new POIs or weaken reliable mobility routines by introducing noisy evidence.
The central principle of EviRec is therefore to separate two questions for each candidate POI: how should the candidate be scored from different evidence sources, and how much should each source be trusted?
We refer to the latter as evidence \emph{reliability}.
Conventional fixed, global, or query-level fusion mechanisms primarily address the first question and therefore cannot resolve the ambiguity of near-zero transition scores.
EviRec addresses both by estimating evidence reliability separately for each candidate from its own observation state.

The remainder of this section develops EviRec from evidence construction to continual recommendation.
Section~\ref{sec:method-values} scores each visible candidate from multiple complementary evidence views.
Section~\ref{sec:method-reliability} characterizes the candidate's observation state and estimates the reliability of each evidence source, thereby resolving the ambiguity of near-zero transition scores.
Section~\ref{sec:method-routing} uses these reliability estimates to gate evidence routing and combine the view-specific scores into the final ranking.

\vspace{-6pt}
\subsection{Multi-view Candidate Scoring}
\label{sec:method-values}

For each candidate POI $p\in\mathcal{C}_{t_i}$, EviRec constructs three complementary scores from a matching view, a transition-memory view, and a lifecycle view. Each score represents how highly the candidate should be ranked when the corresponding evidence source is considered reliable.

\subsubsection{Matching view.}

For query event $i$, a sequence encoder maps the user history $H_i$ to a context vector $\mathbf{h}_i$ that summarizes the user's ordered mobility intent.
EviRec also constructs an attention profile over the non-padding positions in the history.
Let $\mathbf{z}_j$ denote the token embedding of historical visit $j\in H_i$.
The attention weights and pooled history profile are defined as
\vspace{-3pt}
\begin{equation}
    \beta_{ij}=\operatorname{softmax}_{j\in H_i}\bigl(a_\theta(\mathbf{h}_i,\mathbf{z}_j,t_i-t_j)\bigr),\quad \boldsymbol{\rho}_i=\sum_{j\in H_i}\beta_{ij}\mathbf{z}_j,
    \label{eq:history-profile}
\end{equation}
where $a_\theta(\cdot)$ scores each historical visit against the current query state and its recency.
Thus, $\boldsymbol{\rho}_i$ provides a relevance-weighted representation of the user's recent semantic and local context rather than treating all past visits equally.

Let $\mathbf{e}_p$ and $b_p$ denote the embedding and bias of candidate $p$, respectively.
EviRec compares $p$ against both the sequence vector and the attention profile through the same candidate embedding.
The identity term measures compatibility with the user's ordered mobility intent:
\begin{equation}
    s_{\mathrm{base}}(i,p)=\langle \mathbf{h}_i,\mathbf{e}_p\rangle+b_p,
    \label{eq:base-score}
\end{equation}
while the profile term measures semantic and local compatibility,
\begin{equation}
    s_{\mathrm{prof}}(i,p)=\gamma(\mathbf{h}_i)\,\langle \boldsymbol{\rho}_i,\mathbf{e}_p\rangle .
    \label{eq:profile-score}
\end{equation}
The query-dependent gate $\gamma(\mathbf{h}_i)\in(0,1)$ calibrates how much the current query should rely on profile matching.
This term serves as a semantic and local retrieval channel, allowing EviRec to promote candidates that align with the user's recent profile even when their exact POI identities have limited historical support.

The resulting matching-view score $s_{\mathrm{match}}(i,p)=s_{\mathrm{base}}(i,p)+s_{\mathrm{prof}}(i,p)$ is always active, because it does not depend on transition history and therefore gives signal for both established and newly visible candidates.

\subsubsection{Transition-memory view.}
Transition memory records repeated mobility routines.
Let $p_i^{-}$ denote the most recent POI in $H_i$, and let $M_t(q,p)$ aggregate the reciprocal-rank scores of $p$ among the destinations recorded from $q$ using visits observed up to period $t$.
For query $i$, the memory score of candidate $p$ is
\begin{equation}
    m_i(p)=M_{t_i}(p_i^{-},p).
    \label{eq:memory-shorthand}
\end{equation}
Because reciprocal-rank evidence is already bounded and naturally emphasizes highly ranked transitions, no additional compression is required.
EviRec calibrates this evidence according to the current query context:
\begin{equation}
    s_{\mathrm{mem}}(i,p)=\lambda_{\mathrm{mem}}(\mathbf{h}_i)\,m_i(p),
    \label{eq:memory-score}
\end{equation}
where $\lambda_{\mathrm{mem}}(\mathbf{h}_i)\in(0,\infty)$ controls the overall importance of transition evidence for the current query.

\subsubsection{Lifecycle view.}
The lifecycle view characterizes candidate maturity rather than repeated transition behavior.
Using the first-observed period $a(p)$ from Eq.~\eqref{eq:candidate-set}, EviRec assigns $p$ to an age bucket,
\begin{equation}
    A_t(p)=\operatorname{clip}\bigl(t-a(p),0,A_{\max}\bigr).
    \label{eq:age-bucket}
\end{equation}
A learnable scalar bias $b_A$ is associated with each age bucket.
The lifecycle-view score is then defined as
\begin{equation}
    s_{\mathrm{life}}(i,p)=\lambda_{\mathrm{life}}(\mathbf{h}_i)\,b_{A_{t_i}(p)},
    \label{eq:lifecycle-score}
\end{equation}
with a query-dependent scale $\lambda_{\mathrm{life}}(\mathbf{h}_i)\in(0,\infty)$.
This view provides an explicit maturity signal for newly visible or sparsely observed candidates without replacing transition memory when reliable transition evidence is available.

\vspace{-6pt}
\subsection{Evidence-aware Reliability Qualification}
\label{sec:method-reliability}

The scores above indicate how strongly each evidence view supports a candidate, but they do not reveal how much the underlying evidence should be trusted.
EviRec therefore qualifies each view using the candidate's own observation state rather than relying on the score value alone.

This distinction is especially important for transition memory.
When $m_i(p)$ is near zero, the resulting memory-view score $s_{\mathrm{mem}}(i,p)$ is also near zero, but this value may arise for fundamentally different reasons.
The candidate may be irrelevant to the current transition, or it may have appeared too recently to accumulate sufficient transition evidence.
The score alone cannot distinguish between these cases, so applying the same interpretation to every near-zero value can either suppress newly visible POIs or overemphasize unsupported candidates.

EviRec resolves this ambiguity by exposing the candidate's observation state to the subsequent routing mechanism.
The primary reliability signal is a memory-presence indicator that records whether any transition evidence is available:
\begin{equation}
    \pi_i(p)=\mathbbm{1}\{m_i(p)>0\}.
    \label{eq:presence}
\end{equation}
Together with the candidate's age bucket, this indicator distinguishes a mature candidate with no observed transition from a newly visible candidate that has had limited opportunity to accumulate transition evidence.
The full model includes two additional reliability signals.
The first is a first-appearance indicator,
\begin{equation}
\zeta_i(p)
=
\mathbbm{1}\{A_{t_i}(p)=0\},
\label{eq:zeta}
\end{equation}
which explicitly identifies candidates appearing for the first time in the current period, when transition memory is least informative.
The second captures disagreement between transition-memory and lifecycle evidence:
\begin{equation}
\delta_i(p)
=
\operatorname{sg}
\bigl(
s_{\mathrm{mem}}(i,p)-s_{\mathrm{life}}(i,p)
\bigr),
\label{eq:deltap}
\end{equation}
where $\operatorname{sg}(\cdot)$ denotes the stop-gradient operator.
This signal allows the reliability mechanism to respond cautiously when the two evidence sources provide conflicting assessments without allowing the routing objective to alter their scores directly.

The resulting candidate-specific observation state is
\begin{equation}
\mathbf{c}_i(p)
=
\bigl[
\mathbf{h}i,;
A{t_i}(p),;
m_i(p),;
\pi_i(p),;
\zeta_i(p),;
\delta_i(p)
\bigr].
\label{eq:routing-state}
\end{equation}
Except for the shared query representation $\mathbf{h}_i$, all components describe the observation state of the individual candidate.
This candidate-specific construction enables EviRec to estimate evidence reliability separately for every visible POI.

\subsection{Reliability-Gated Evidence Routing}
\label{sec:method-routing}

EviRec combines the three routed evidence views through a reliability gate, implemented as a lightweight mixture-of-experts router~\cite{jacobs1991moe,shazeer2017moe,fedus2022switch} in which the transition-memory and lifecycle views serve as experts. The gate is driven by the candidate-specific reliability qualification introduced in Section~\ref{sec:method-reliability}.
Unlike a standard gate conditioned only on the query, EviRec converts each candidate's observation state into a candidate-specific reliability weight for transition memory.
A fixed or global gate would assign the same weight to every candidate and therefore remain unable to resolve the ambiguity of near-zero transition scores.
EviRec instead estimates reliability from each candidate's own state:
\begin{equation}\footnotesize
    \ell_i(p)=R_{\mathrm{ctx}}(\mathbf{h}_i)+\beta^{\mathrm{age}}_{A_{t_i}(p)}+w_m\,m_i(p)+w_\pi\bigl(1-\pi_i(p)\bigr)+w_\zeta\,\zeta_i(p)+w_\delta\,\delta_i(p),
    \label{eq:router}
\end{equation}
\begin{equation}
    r_i(p)=\sigma\bigl(\ell_i(p)\bigr)\in(0,1),
    \label{eq:route-weight}
\end{equation}
where $R_{\mathrm{ctx}}$ is a context term over $\mathbf{h}_i$, $\beta^{\mathrm{age}}$ is a learnable per-bucket gate bias, and $w_m$, $w\pi$, $w_\zeta$, and $w_\delta$ are learnable scalar parameters.
The presence and first-appearance coefficients $w_\pi$ and $w_\zeta$ are initialized to negative values, providing an initial prior against transition memory for candidates with no observed support or appearing for the first time.
Training can subsequently relax or adjust this prior according to the observed recommendation outcomes.
We interpret $r_i(p)$ as the reliability weight assigned to transition memory and define $r_{i,\mathrm{mem}}(p)=r_i(p)$ and $r_{i,\mathrm{life}}(p)=1-r_i(p)$.
A high value indicates that transition evidence is reliable, whereas a low value shifts the candidate toward lifecycle evidence because its transition support is sparse or unavailable.

The final score adds the always-active matching-view score to the reliability-weighted combination of the routed evidence views:
\begin{equation}
    s_i(p)=s_{\mathrm{match}}(i,p)+r_i(p)\,s_{\mathrm{mem}}(i,p)+\bigl(1-r_i(p)\bigr)\,s_{\mathrm{life}}(i,p).
    \label{eq:final-score}
\end{equation}
The gate therefore performs a candidate-specific reliability assessment rather than applying a global correction for newly observed POIs.
Within the same query, EviRec can preserve transition-based ranking for well-supported candidates while shifting newly visible or under-observed candidates toward the lifecycle view.
The appropriate regime is determined separately for each candidate from its own observation state.

\vspace{-6pt}
\subsection{Training and Online Inference}
\label{sec:method-training}

EviRec is trained only over POIs that are visible during the target period.
All scores outside $\mathcal{C}_{t_i}$ are masked, ensuring that the model never ranks a POI before it appears in the stream.
The training objective is the cross-entropy loss over the visible candidate set:
\begin{equation}
    \mathcal{L}=-\sum_i \log \frac{\exp s_i(p_i)}{\sum_{p\in\mathcal{C}_{t_i}}\exp s_i(p)}.
    \label{eq:training-loss}
\end{equation}
The scoring, reliability-qualification, and gating parameters are optimized jointly under this objective, allowing the reliability weights to be learned from recommendation outcomes rather than determined by a handcrafted threshold.
During online inference, each newly observed period updates the transition memory $M_t$ and the first-observed metadata $a(p)$ before the subsequent period is processed.
Consequently, the observation state $\mathbf{c}_i(p)$ evolves with the mobility stream, while the same reliability gate continually determines how much each candidate should rely on transition-memory versus lifecycle evidence.
As new interactions accumulate, EviRec increases its reliance on transition memory for candidates that acquire sufficient support while continuing to use lifecycle evidence for candidates that remain under-observed.
This process allows candidate-specific reliability decisions to evolve over time without treating all POIs as uniformly established or uniformly new.

%% file: sections/newEvaluation.tex
\section{Evaluation}

\subsection{Experimental Setting}

\subsubsection{Dataset}
We evaluate EviRec on real-world POI check-in data collected by Veraset from five U.S. cities in 2025: Boston, Chicago, Houston, Los Angeles, and New York.
Table~\ref{tab:dewey-stats} reports the per-city statistics averaged across monthly snapshots. The final three columns show the proportions of test queries involving a newly visible target POI (POI-New), a first-time user (User-New), or both simultaneously (Dual-New). POI-New queries account for 18\% to 28\% of the test stream, while User-New queries account for 23\% to 38\%, demonstrating that dynamic cold-start cases are both prevalent and heterogeneous across cities.

We adopt a chronological predict-then-update protocol with three consecutive temporal splits. The model is initialized using data from January through April, tuned on a held-out validation month, and evaluated sequentially on the remaining reporting months. Each reporting month is evaluated before its interactions are used for model updating, ensuring that no future information is available at prediction time. Before evaluating the next month, the model performs one online adaptation pass using the most recently observed data to incorporate the latest stream state. At inference time, every method ranks only the POIs in the temporally visible candidate set $\mathcal{C}_t$ defined in Eq.~\eqref{eq:candidate-set}. Consequently, a newly observed POI enters the recommendation space only in the month in which it first appears.

\begin{table}[t]
\centering
\small
\caption{Per-city dataset statistics (monthly averages).}\vspace{-10pt}
\label{tab:dewey-stats}
\resizebox{\columnwidth}{!}{%
\begin{tabular}{lcccccc}
\toprule
City & \# Users & \# POIs & \# Cat. & POI-New & User-New & Dual-New \\
\midrule
Boston & 1.48M & 7.3K & 168 & 18.4\% & 32.4\% & 8.3\% \\
Chicago & 5.10M & 38.0K & 222 & 21.6\% & 26.8\% & 7.4\% \\
Houston & 6.57M & 59.3K & 245 & 21.3\% & 23.3\% & 6.5\% \\
Los Angeles & 5.16M & 34.2K & 213 & 24.7\% & 26.4\% & 8.6\% \\
New York & 3.79M & 26.8K & 192 & 28.0\% & 38.0\% & 12.9\% \\
\bottomrule
\end{tabular}}\vspace{-10pt}
\end{table}

\subsubsection{Metrics}
\begin{sloppypar}
Because EviRec is designed for dual cold-start recommendation, we focus primarily on the three cold-start query subsets defined in Sec.~\ref{sec:Problem}, while reporting performance over all queries as a reference.
We use two standard top-$K$ ranking metrics: Recall@$K$ denoted as R@$K$, and NDCG@$K$ denoted as N@$K$.
R@$K$ is 1 when the ground-truth next POI appears among the top-$K$ recommendations and 0 otherwise, measuring whether the correct POI is successfully retrieved. N@$K$ is $1/\log_2(r+1)$ when the ground-truth POI is ranked at position $r\leq K$ and 0 otherwise, additionally rewarding methods that place the correct POI closer to the top of the ranking. We report both metrics on four query subsets: All, POI-New, User-New, and Dual-New.
\begin{itemize}
\item \textbf{POI-New} ($\mathcal{Q}^{\mathrm{POI}}$): queries whose target POI first becomes visible in the current period and therefore has no transition evidence from earlier periods.
\item \textbf{User-New} ($\mathcal{Q}^{\mathrm{User}}$): queries from users who first appear in the current period and therefore have no personalized history from earlier periods.
\item \textbf{Dual-New} ($\mathcal{Q}^{\mathrm{Dual}}$): the intersection of POI-New and User-New. This is the most challenging subset because both the target POI and the querying user lack historical support from prior periods.
\item \textbf{All}: all test queries in the evaluation periods, reported as an aggregate reference over the complete stream.
\end{itemize}
We compute both metrics at $K\in{10,20,30}$. Table~\ref{tab:main-full-year} reports N@10 and R@10 for the three cold-start subsets and the All subset, while results at the remaining cutoffs are provided in the appendix.
\end{sloppypar}

\begin{table*}[!t]
\centering
\scriptsize 
\caption{Full-year average performance across all cities. We report NDCG@10 and Recall@10 on the three cold-start subsets and on the all-query aggregate. Best is bold and the second is underlined.}\vspace{-8pt}
\label{tab:main-full-year}
\resizebox{\textwidth}{!}{%
\begin{tabular}{lcccccccc}
\toprule
& \multicolumn{2}{c}{POI-New} & \multicolumn{2}{c}{User-New} & \multicolumn{2}{c}{Dual-New} & \multicolumn{2}{c}{All} \\
\cmidrule(lr){2-3}\cmidrule(lr){4-5}\cmidrule(lr){6-7}\cmidrule(lr){8-9}
Method & N@10 & R@10 & N@10 & R@10 & N@10 & R@10 & N@10 & R@10 \\
\midrule
SASRec   & 0.055$\pm$0.006 & 0.064$\pm$0.006 & 0.094$\pm$0.006 & 0.103$\pm$0.006 & 0.066$\pm$0.003 & 0.074$\pm$0.002 & 0.133$\pm$0.004 & 0.147$\pm$0.003 \\
GETNext  & 0.129$\pm$0.000 & 0.169$\pm$0.000 & 0.196$\pm$0.000 & 0.236$\pm$0.000 & 0.156$\pm$0.000 & 0.194$\pm$0.000 & 0.226$\pm$0.000 & 0.275$\pm$0.000 \\
KBGNN    & 0.131$\pm$0.002 & 0.170$\pm$0.001 & 0.193$\pm$0.003 & 0.232$\pm$0.004 & 0.152$\pm$0.004 & 0.190$\pm$0.005 & 0.225$\pm$0.001 & 0.272$\pm$0.003 \\
DisenPOI & 0.145$\pm$0.005 & 0.182$\pm$0.004 & 0.203$\pm$0.004 & 0.242$\pm$0.004 & 0.160$\pm$0.006 & 0.197$\pm$0.004 & 0.236$\pm$0.004 & 0.282$\pm$0.004 \\
Diff-POI & \underline{0.145$\pm$0.002} & \underline{0.186$\pm$0.003} & \underline{0.213$\pm$0.001} & \underline{0.253$\pm$0.001} & \underline{0.172$\pm$0.001} & \underline{0.211$\pm$0.001} & \underline{0.241$\pm$0.001} & \underline{0.290$\pm$0.001} \\
BiGSL    & 0.136$\pm$0.002 & 0.176$\pm$0.002 & 0.200$\pm$0.004 & 0.241$\pm$0.004 & 0.156$\pm$0.004 & 0.196$\pm$0.005 & 0.231$\pm$0.003 & 0.281$\pm$0.004 \\
GNPR-SID & 0.133$\pm$0.000 & 0.172$\pm$0.001 & 0.196$\pm$0.001 & 0.235$\pm$0.001 & 0.155$\pm$0.003 & 0.193$\pm$0.003 & 0.226$\pm$0.001 & 0.274$\pm$0.001 \\
CAGNN    & 0.128$\pm$0.004 & 0.167$\pm$0.003 & 0.192$\pm$0.002 & 0.231$\pm$0.003 & 0.150$\pm$0.003 & 0.188$\pm$0.003 & 0.222$\pm$0.003 & 0.270$\pm$0.003 \\
\textbf{EviRec} & \textbf{0.185$\pm$0.005} & \textbf{0.239$\pm$0.005} & \textbf{0.222$\pm$0.002} & \textbf{0.267$\pm$0.003} & \textbf{0.207$\pm$0.004} & \textbf{0.259$\pm$0.005} & \textbf{0.252$\pm$0.004} & \textbf{0.305$\pm$0.004} \\
\bottomrule
\end{tabular}}\vspace{-8pt}
\end{table*}
% \vspace{-10pt}

\subsubsection{Baselines}

We compare EviRec with eight state-of-the-art POI recommendation baselines including SASRec~\cite{kang2018sasrec}, GETNext~\cite{yang2023getnext}, 
KBGNN~\cite{ju2022kbgnn},
DisenPOI~\cite{qin2023disenpoi}, 
Diff-POI~\cite{qin2023diffpoi}, 
BiGSL~\cite{wang2024bigsl}, 
GNPR-SID~\cite{wang2025gnprsid},
and CAGNN~\cite{lei2025cagnn}, together with their monthly updated variants when applicable. These methods cover sequence-based, trajectory-flow, graph-based, context-adaptive, disentangled, and diffusion-based next-POI recommendation.
More details of these baselines will be shown in the Appendix.

% We compare EviRec with eight POI recommendation baselines.
% \begin{itemize}
%     \item \textbf{SASRec} \cite{kang2018sasrec}: a unidirectional self-attention encoder over the visited-POI sequence.
%     \item \textbf{GETNext} \cite{yang2023getnext}: a global trajectory-flow map fused with a Transformer over the personal check-in sequence.
%     \item \textbf{KBGNN} \cite{ju2022kbgnn}: kernel-based message passing over separate geographical and sequential graphs.
%     \item \textbf{DisenPOI} \cite{qin2023disenpoi}: contrastive disentangling of sequential-transition and geographical preference from two parallel graphs.
%     \item \textbf{Diff-POI} \cite{qin2023diffpoi}: a graph-encoded diffusion process over latent spatial preference that samples the next-POI distribution, the strongest baseline on our dataset.
%     \item \textbf{BiGSL} \cite{wang2024bigsl}: bi-level transition-graph learning over fine-grained POI nodes and coarse prototype clusters.
%     \item \textbf{GNPR-SID} \cite{wang2025gnprsid}: hierarchical semantic-ID quantization of POIs with generative decoding, adapted to our full-catalog ranking protocol.
%     \item \textbf{CAGNN} \cite{lei2025cagnn}: spatiotemporal-context-conditioned graph propagation and attention.
% \end{itemize}
% When supported, we report the strongest available protocol, and all neural baselines are updated month by month on past data only, under the same temporal candidate visibility rule as EviRec.
% \end{sloppypar}
\vspace{-6pt}
\subsection{Main Performance}

Table~\ref{tab:main-full-year} reports the full-year performance averaged across all five cities. Each result is presented as the mean $\pm$ standard deviation across available random seeds, with zero deviation reported for methods evaluated using a single seed. Diff-POI is the strongest baseline across all query subsets and metrics and is therefore used as the primary reference for comparison.

EviRec delivers its largest improvements in the settings where historical transition evidence is weakest. Compared with Diff-POI, which achieves 0.145 N@10 on POI-New and 0.172 N@10 on Dual-New, EviRec improves POI-New performance by 27.3\% in N@10 and 28.7\% in R@10. On the most challenging Dual-New subset, it yields gains of 20.4\% in N@10 and 22.6\% in R@10. 
% The improvements on User-New are smaller, at 4.2\% in N@10 and 5.4\% in R@10, because a newly observed user may still benefit from transition evidence associated with an already observed target POI. 
EviRec improves the All-query aggregate by 4.6\% in N@10 and 5.2\% in R@10. These results show that EviRec not only improves overall recommendation quality but also concentrates its gains on the cold-start cases.

\begin{figure}[h]
\centering
\includegraphics[height=0.46\columnwidth]{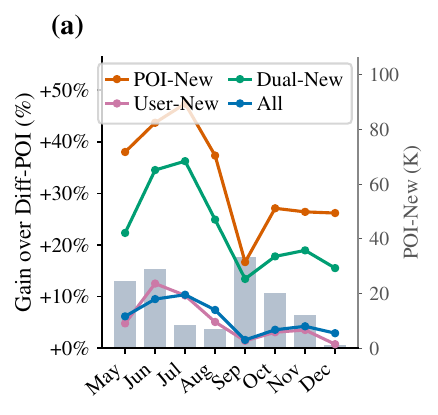}\hfill
\includegraphics[height=0.46\columnwidth]{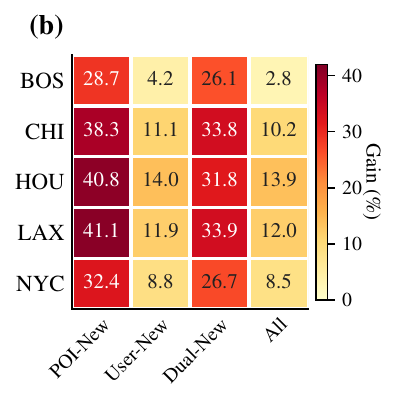}\vspace{-6pt}
\caption{EviRec performance compared with the strongest baseline (Diff-POI) across time (a) and cities (b).}\vspace{-10pt}
\label{fig:eval-per-city}
\end{figure}

Figure~\ref{fig:eval-per-city} further examines the temporal and geographic consistency of these improvements.
Figure~\ref{fig:eval-per-city}(a) presents the monthly relative NDCG@10 gains over Diff-POI for each query subset. EviRec consistently improves POI-New and Dual-New performance in every reporting month, indicating that its advantage is not driven by a single period.
Figure~\ref{fig:eval-per-city}(b) 
reports the corresponding per-city gains. EviRec improves every city-subset combination, with macro-average gains of 9.5\% on All, 36.3\% on POI-New, 10.0\% on User-New, and 30.4\% on Dual-New in N@10. Even in Boston, which exhibits the smallest aggregate improvement of 2.8\%, EviRec achieves substantial gains of 28.7\% on POI-New and 26.1\% on Dual-New. These results demonstrate that EviRec's advantage is consistent across both time and cities and is most pronounced in the dynamic cold-start settings targeted by the method.

\vspace{-10pt}

\vspace{-10pt}
\subsection{Ablation Study}

Table~\ref{tab:ablation} isolates the contribution of EviRec's reliability-gated routing by ablating the design choices introduced in Sections~\ref{sec:method-values}--\ref{sec:method-routing}.
All variants use the same matching, transition-memory, and lifecycle views and differ only in how the latter two are combined. The comparison therefore focuses specifically on the routing mechanism. Because transition memory is generally reliable for well-observed POIs, we concentrate on the cold-start subsets and omit the All-query aggregate, where performance is already high.

The \textbf{Memory-only} variant fixes the routing weight to rely exclusively on transition memory and removes lifecycle evidence. Its performance drops substantially on the cold-start subsets, reaching 0.158 N@10 on POI-New and 0.183 on Dual-New, because newly visible targets have little or no prior transition support.
The \textbf{Fixed blend} variant replaces the learned gate with a uniform combination of transition-memory and lifecycle evidence. It reaches 0.182 N@10 on POI-New and 0.204 on Dual-New, outperforming all reduced-gate variants. This result demonstrates that incorporating lifecycle evidence is already beneficial. However, a fixed blend cannot adjust the balance between the two views according to each candidate's observation state.
The next two variants retain both evidence views but remove candidate-specific routing information. \textbf{Global gate} learns a single weight shared by all queries and candidates, whereas \textbf{Context-only gate} conditions the weight on the user-context representation $\mathbf{h}_i$ without using candidate-specific features. Both achieve 0.160 N@10 on POI-New and 0.185 on Dual-New, indicating that query context alone provides little benefit for distinguishing newly visible candidates from those with reliable transition support.
The \textbf{w/o observation state} variant conditions the gate on query context and candidate age but removes the remaining candidate-specific signals in Eq.~\eqref{eq:routing-state}, including the memory score, memory-presence indicator, first-appearance indicator, and disagreement between the two evidence views. It improves only slightly, reaching 0.163 N@10 on POI-New and 0.188 on Dual-New. This result shows that candidate age alone is insufficient to determine whether weak transition support reflects irrelevance or insufficient observation.
Full EviRec achieves the best performance in every column. The consistent pattern across variants validates EviRec's central design principle: effective arbitration between transition-memory and lifecycle evidence requires a reliability gate conditioned on each candidate's own observation state rather than on global or query-level information alone.

\begin{table}[t]
\centering
\scriptsize 
\caption{Ablation of EviRec's reliability gate. All variants differ only in the gating rule. }
\label{tab:ablation}\vspace{-10pt}
\resizebox{\columnwidth}{!}{%
\begin{tabular}{lcccccc}
\toprule
& \multicolumn{2}{c}{POI-New} & \multicolumn{2}{c}{User-New} & \multicolumn{2}{c}{Dual-New} \\
\cmidrule(lr){2-3}\cmidrule(lr){4-5}\cmidrule(lr){6-7}
Variant & N@10 & R@10 & N@10 & R@10 & N@10 & R@10 \\
\midrule
Memory-only & 0.158 & 0.202 & 0.224 & 0.267 & 0.183 & 0.225 \\
Fixed blend & 0.182 & 0.235 & 0.223 & 0.268 & 0.204 & 0.255 \\
Global gate & 0.160 & 0.205 & 0.224 & 0.268 & 0.185 & 0.228 \\
Context-only gate & 0.160 & 0.205 & 0.224 & 0.268 & 0.185 & 0.228 \\
w/o observation state & 0.163 & 0.208 & \textbf{0.225} & \textbf{0.268} & 0.188 & 0.231 \\
\textbf{EviRec} & \textbf{0.190} & \textbf{0.247} & 0.221 & 0.267 & \textbf{0.212} & \textbf{0.266} \\
\bottomrule
\end{tabular}}\vspace{-10pt}
\end{table}

\subsection{In-depth Analysis}

\begin{figure}[t]
\centering
\includegraphics[width=\columnwidth]{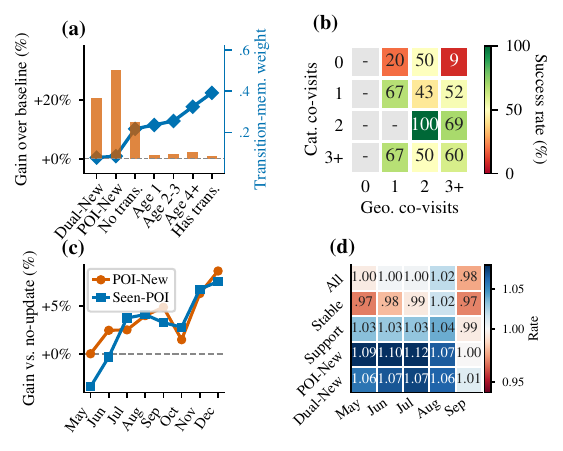}
\vspace{-20pt}
\caption{In-depth analysis.
  (a)~For each target group, the gain over Diff-POI (bars) and the weight EviRec puts on transition memory (line), sorted by that weight, showing that the gate relies least on transition memory where it helps most.
 (b)~For age-0 targets ranked outside the top 10 by transition memory alone, we compare 120 cases recovered by EviRec with 120 cases still missed. Each cell reports the recovery rate for targets grouped by the number of past visits to the same category and area.
(c)~Monthly NDCG@10 gain of the continually updated model over its frozen counterpart on POI-New and Seen-POI queries.
(d)~Forgetting analysis. For each earlier month, we report the ratio between its NDCG@10 under a later checkpoint and its NDCG@10 immediately after learning that month. Values near or above 1 indicate that earlier performance is retained.}\vspace{-10pt}
\label{fig:in-depth}
\end{figure}

\subsubsection{How Does the Gate Adapt to POI Maturity?}
\label{sec:router-adapts}

A central question is whether the learned gate adapts its evidence allocation to each candidate's observation state or simply maintains a fixed preference for lifecycle evidence.
Figure~\ref{fig:in-depth}(a) answers this by comparing EviRec with Diff-POI across seven target groups.
The groups are defined as follows.
\textbf{POI-New} contains targets appearing in the catalog for the first time at age~0 and so having no transition evidence from earlier periods. \textbf{Dual-New} further requires the querying user to be newly observed, yielding the most challenging dual cold-start setting.
\textbf{No transition} contains targets for which no prior transition from the user's most recently visited POI has been recorded. It is a superset of POI-New and also includes mature POIs that have never been reached from the current source POI. \textbf{Has transition} contains targets with at least one previously recorded transition from that source. The remaining groups partition targets by the number of months since their first appearance, namely age~1, age~2--3, and age~4+. For each group, the line reports the average transition-memory weight $r_i(p^\star)$ assigned to the ground-truth target. A lower value indicates greater reliance on lifecycle evidence.

The routing pattern changes consistently with POI maturity and transition support. Dual-New and POI-New, which achieve the largest gains over Diff-POI at 20.7\% and 30.4\%, receive the lowest transition-memory weights, 0.0773 and 0.0843, respectively. This shows that the gate suppresses transition memory when the target lacks historical support. As targets mature from age~1 to age~2--3 and age~4+, the transition-memory weight gradually increases toward 0.32, while the gain decreases to below 2.5\%. This trend indicates that EviRec increasingly relies on transition memory as candidates accumulate more reliable historical evidence.

The \textbf{Has transition} group provides the clearest contrast. Diff-POI already achieves an NDCG@10 of 0.6412 on these targets, leaving EviRec with a gain of only 0.8\%, while the average transition-memory weight rises to 0.3924. Thus, the gate does not indiscriminately replace transition memory with lifecycle evidence. Instead, it restores reliance on memory when prior transitions provide reliable support. This candidate-specific calibration is learned end to end from the observation-state vector Eq.~\eqref{eq:routing-state}, without relying on a handcrafted threshold.

\subsubsection{What Distinguishes a Successful Recovery from a Failure?}
We investigate what enables EviRec to recover newly observed POIs when transition memory fails. Specifically, we consider only age-0 targets that the transition-memory view ranks outside the top 10 and construct two equally sized groups: 120 \emph{successful} cases in which full EviRec recovers the target within the top 10, and 120 \emph{failed} cases in which it does not. Because transition memory fails in both groups and all targets have the same age, systematic differences between them reveal which signals in the user's recent history are associated with successful recovery, particularly the semantic and spatial compatibility captured by the profile-matching term $s_{\mathrm{prof}}$ in Eq.~\eqref{eq:profile-score}.

Figure~\ref{fig:in-depth}(b) reports the success rate in each category-by-geospatial co-visit cell across the two pools.
Category depth, not geospatial depth, separates the outcomes.
The zero-category row stays weak, with 20\% success at $n=5$ and 9\% at $n=43$, while every row with at least one same-category co-visit reaches 43--100\% regardless of the geospatial column.
Aggregated over the grid, zero same-category co-visits succeed 12.0\% of the time (6/50) versus 60.0\% (114/190) with at least one, a five-fold gap driven by category depth alone, since varying geospatial co-visits within a fixed category row changes the rate far less than moving across rows.
This confirms that $s_{\mathrm{prof}}$ recovers age-0 targets mainly through semantic compatibility rather than spatial proximity, and this advantage naturally tapers off once transition support exists, since in the has-transition group of Figure~\ref{fig:in-depth}(a) Diff-POI already reaches 0.6412 N@10 and EviRec gains only +0.8\% at a memory weight of 0.3924, so the gate hands ranking back to transition memory.

\subsubsection{Do Online Updates Help Without Causing Catastrophic Forgetting?}
We examine whether monthly online updates improve adaptation beyond the initial checkpoint without degrading previously learned behavior. The frozen control uses the same initialization and candidate-visibility protocol as EviRec but disables monthly adaptation with \texttt{update\_epochs=0} and no replay. Figure~\ref{fig:in-depth}(c) shows that online updates improve POI-New NDCG@10 in all eight reporting months, with an average gain of 3.8\%. They also improve Seen-POI performance in six of eight months, yielding an average gain of 3.1\%, with the largest decline limited to 3.4\% in May. These results indicate that continual adaptation improves responsiveness to newly observed POIs while largely preserving performance on established ones~\cite{yoo2025continualrec}.

We further test forgetting by evaluating fixed queries from each earlier month $m$ using both the checkpoint obtained immediately after updating on $m$ and a later checkpoint. We report the ratio of later-checkpoint NDCG@10 to immediate-checkpoint NDCG@10, where values near or above 1 indicate retention or improvement. Figure~\ref{fig:in-depth}(d) shows little evidence of catastrophic forgetting. Across three seeds and five evaluated months, the retention ratios are $0.999{\pm}0.015$ on All queries and $0.986{\pm}0.019$ on stable returning-user queries. Performance improves on memory-supported queries at $1.026{\pm}0.016$, POI-New at $1.074{\pm}0.045$, and Dual-New at $1.055{\pm}0.026$. Overall, EviRec adapts to newly observed POIs while largely retaining previously learned mobility patterns.

\vspace{-6pt}
\subsection{Computational Efficiency}

Online continual POI recommendation also requires practical inference cost.
Table~\ref{tab:efficiency} reports evaluate-only wall-clock time on the same test split of 600,150 examples, averaged over three seeds on a single GPU under an identical batch-ranking protocol (batch size 512). The reported time measures only the cost of generating ranked POI lists after loading a trained checkpoint and excludes data loading and monthly model adaptation.

EviRec is not the fastest method because it jointly computes profile matching, transition-memory, and lifecycle scores before applying candidate-specific reliability gating.
Nevertheless, its inference overhead remains modest. 
EviRec processes the complete test split in 5.6 seconds on average, compared with 6.1 seconds for Diff-POI, the strongest competing method in recommendation accuracy. 
This result reflects a favorable accuracy-efficiency trade-off: although reliability-gated evidence routing introduces additional computation, EviRec remains within the same practical inference range as competitive neural POI recommendation systems.

\vspace{-6pt}
\begin{table}[h]
\centering
\scriptsize
\caption{Inference efficiency on the full test split. It measures only the cost of generating ranked POI lists after a checkpoint is loaded, excluding data loading and adaptation.}\vspace{-10pt}
\label{tab:efficiency}
\resizebox{\columnwidth}{!}{%
\begin{tabular}{lccc}
\toprule
Method & Trainable params & Trainable frac. & Inference time (s) \\
\midrule
SASRec & 516.2K & 3.47\% & 2.6 \\
GETNext & 621.7K & 4.15\% & 4.9 \\
KBGNN & 655.5K & 4.37\% & 6.5 \\
DisenPOI & 656.8K & 4.38\% & 4.4 \\
Diff-POI & 822.7K & 5.42\% & 6.1 \\
BiGSL & 654.9K & 4.36\% & 5.3 \\
GNPR-SID & 621.7K & 4.13\% & 5.5 \\
CAGNN & 621.7K & 4.12\% & 5.1 \\
\textbf{EviRec} & 631.9K & 4.22\% & 5.6 \\
\bottomrule
\end{tabular}}\vspace{-6pt}
\end{table}

%% file: sections/related_work.tex
\section{Related Work}

\subsection{POI Recommendation}
Different from item recommendation in e-commerce,
POI recommendation needs to jointly model sequential transitions, geographic proximity, temporal regularity, and user-specific mobility patterns. Early studies focus on recurrent modeling, self-attention, and spatiotemporal context~\cite{tan2016gru4rec,kang2018sasrec,luo2021stan}. More recent methods incorporate graph structures, disentangled spatial and sequential preferences, temporal regularities, hierarchical POI representations, context-adaptive graph propagation, and language-model reasoning~\cite{yang2023getnext,ju2022kbgnn,qin2023disenpoi,qin2023diffpoi,deng2024replay,wang2024bigsl,lei2025cagnn,liu2025gallm}. Although these methods substantially improve recommendation accuracy, most are developed and evaluated under offline protocols with a fixed or implicitly stable user and POI universe. A separate line of work targets the cold-start problem directly, learning to produce useful representations for users or items with little interaction history through meta-learning or embedding warm-up~\cite{lee2019melu,pan2019warmup}. These methods, however, typically assume a static catalog and treat cold start as a one-off initialization step rather than a recurring property of a live stream. In contrast, we study continual dual cold-start recommendation, where new users and POIs enter over time and temporal candidate visibility is part of the task definition.

\vspace{-2pt}
\subsection{Sequential and Continual Recommendation}

Sequential recommenders such as GRU4Rec, SASRec, and BERT4Rec encode ordered user interactions, while graph-based methods such as LightGCN exploit collaborative user-item structure~\cite{tan2016gru4rec,kang2018sasrec,sun2019bert4rec,he2020lightgcn}. Continual recommendation considers evolving user preferences and item catalogs, with the goal of incorporating new evidence without forgetting useful historical knowledge~\cite{yoo2025continualrec}. It inherits the central stability-plasticity trade-off studied in continual learning, where a model must absorb new information without catastrophically forgetting previously acquired knowledge~\cite{kirkpatrick2017ewc,delange2022clsurvey}. Classical remedies constrain important parameters~\cite{kirkpatrick2017ewc,li2017lwf} or replay stored exemplars and gradients~\cite{rebuffi2017icarl,lopezpaz2017gem}, and recommendation-specific approaches address the same trade-off through dynamic model expansion and graph prompt tuning~\cite{he2023degc,zhang2024gpt4rec}. Rather than expanding the model or adapting graph prompts, EviRec performs candidate-level reliability routing between transition-memory and lifecycle evidence, preserving established routines while adapting to newly visible and under-observed POIs.

%% file: sections/conclusion.tex
\vspace{-3pt}
\section{Conclusion}

In this paper, motivated by a 10-city analysis revealing substantial changes in POI catalogs, user populations, and category distributions over time, we study continual dual cold-start next-POI recommendation in evolving urban environments. We propose \textbf{EviRec}, a continual evidence learning framework that estimates the reliability of historical evidence separately for each candidate POI. EviRec integrates matching, transition-memory, and lifecycle views, qualifies their reliability using candidate-specific observation states, and adaptively routes between transition-memory and lifecycle evidence. This design preserves reliable mobility routines while reducing dependence on sparse or unavailable transition history for newly visible POIs. Experiments on a full-year, five-city POI check-in dataset show that EviRec consistently outperforms state-of-the-art baselines, with the largest gains concentrated on cold-start queries. In particular, EviRec improves NDCG@10 by 20.4\% on Dual-New queries over the strongest baseline, and monthly updates improve responsiveness to newly observed POIs without substantially degrading previously learned behavior. 

% Ablation and routing analyses confirm that these gains primarily arise from candidate-specific reliability gating, while continual adaptation and forgetting analyses show that monthly updates improve responsiveness to newly observed POIs without substantially degrading previously learned behavior.

%% file: sections/appendix_benchmark_provenance.tex
% \section{Dataset Provenance}

% The full-year dataset spans January through December 2025 and covers five cities: Boston, Chicago, Houston, Los Angeles, and New York. It includes all 60 expected city-month sources with no missing data. January and May through December are stored as fast Parquet sources, while February through April use legacy CSV sources. Before sequence construction, the dataset contains 17,633,454 sampled event records, yielding 684,200 sequence examples: 58,382 for training, 25,668 for validation, and 600,150 for testing. The validation and test sets together contain 148,513 POI-New examples and 167,418 User-New examples.

\section{Ethical Use of Data and Informed Consent}
This study uses de-identified mobility data (POI check-in trajectories) collected by Veraset and provided by Dewey under the applicable data-use agreement. The research team did not directly recruit participants or collect personal information, and any required user notification or consent was handled by the original data provider under its privacy policies and legal requirements. We use the data solely for research, restrict access to authorized researchers, and report results only in aggregate. We do not attempt to identify individuals, link mobility records to external personal information, infer sensitive personal attributes, or disclose individual trajectories. IRB is not required for this research.

%% file: sections/appendix_metric_definitions.tex
% \section{Metric Definitions}

% NDCG@$K$ discounts the rank of the held-out target POI within the top-$K$ predictions, assigning higher scores when the target appears closer to the top. Recall@$K$ equals 1 if the target POI appears among the top-$K$ predictions and 0 otherwise, and is averaged across all evaluation examples. POI-New metrics are computed only on examples whose target POI is first observed in the target month or after the training window, according to the dataset's first-observation metadata. User-New metrics are computed only on examples whose users are identified as newly observed under the temporal evaluation protocol. The Dual-New subset is the intersection of the POI-New and User-New subsets.

%% file: sections/appendix_current_experimental_boundary.tex
\section{Experimental Details}

\subsection{Dataset}

The full-year dataset spans January through December 2025 and covers five cities: Boston, Chicago, Houston, Los Angeles, and New York. It includes all 60 expected city-month sources with no missing data. January and May through December are stored as fast Parquet sources, while February through April use legacy CSV sources. Before sequence construction, the dataset contains 17,633,454 sampled event records, yielding 684,200 sequence examples: 58,382 for training, 25,668 for validation, and 600,150 for testing. The validation and test sets together contain 148,513 POI-New examples and 167,418 User-New examples.
The dataset statistics are summarized in Table~\ref{tab:dewey-stats}. 
The early months are used for model training, the subsequent month is used for validation and warm-start adaptation, and the remaining months are evaluated sequentially.

At each evaluation period, the candidate POIs are restricted according to the period-level visibility rule in Eq.~\eqref{eq:candidate-set}.
This constraint preserves causality by preventing methods from ranking POIs before they appear in the mobility stream, while allowing newly visible POIs to be evaluated as soon as they become observable.

\subsection{Metrics}

We report NDCG@$K$ and Recall@$K$ for $K\in{10,20,30}$.
Recall@$K$ measures whether the ground-truth next POI appears in the top-$K$ ranking, while NDCG@$K$ additionally rewards methods that place the ground-truth POI closer to the top.
NDCG@$K$ discounts the rank of the held-out target POI within the top-$K$ predictions, assigning higher scores when the target appears closer to the top. Recall@$K$ equals 1 if the target POI appears among the top-$K$ predictions and 0 otherwise, and is averaged across all evaluation examples. POI-New metrics are computed only on examples whose target POI is first observed in the target month or after the training window, according to the dataset's first-observation metadata. User-New metrics are computed only on examples whose users are identified as newly observed under the temporal evaluation protocol. The Dual-New subset is the intersection of the POI-New and User-New subsets.

In addition to the aggregate performance over all queries, we report three dynamic subsets that reflect the motivation of the task.
POI-New contains examples whose target POI first becomes visible in the current period, User-New contains examples from users first observed under the temporal protocol, and Dual-New is the intersection of the two subsets.

\subsection{Baselines}

\begin{sloppypar}
% \begin{itemize}
% \item \textbf{SASRec}~\cite{kang2018sasrec}: a self-attentive sequential recommender.
% It is the sequence-only baseline without explicit graph, geography, or online lifecycle modeling.
% \item \textbf{GETNext}~\cite{yang2023getnext}: a trajectory-flow Transformer for next-POI recommendation.
% It tests whether transition-flow modeling is enough under dynamic POI visibility.
% \item \textbf{BiGSL}~\cite{wang2024bigsl}: a bi-level graph learning baseline over POI/prototype structure.
% It is a strong graph baseline for sparse POI relations.
% \item \textbf{CAGNN}~\cite{lei2025cagnn}: a context-adaptive graph neural baseline.
% It tests edge-specific context adaptation without EviRec's lifecycle-memory router.
% \item \textbf{DisenPOI}~\cite{qin2023disenpoi}: a model that disentangles sequential and geographical influence.
% It tests whether separating routine and spatial preference closes the gap.
% \item \textbf{Diff-POI}~\cite{qin2023diffpoi}: a diffusion-style POI recommender.
% It is the strongest non-EviRec baseline in the full-year online protocol.
% \item \textbf{KBGNN}~\cite{ju2022kbgnn}: a kernel graph neural baseline for next-POI recommendation.
% It tests high-order graph substructure and kernelized candidate scoring.
\begin{itemize}
    \item \textbf{SASRec} \cite{kang2018sasrec}: a unidirectional self-attention encoder over the visited-POI sequence.
    \item \textbf{GETNext} \cite{yang2023getnext}: a global trajectory-flow map fused with a Transformer over the personal check-in sequence.
    \item \textbf{KBGNN} \cite{ju2022kbgnn}: kernel-based message passing over separate geographical and sequential graphs.
    \item \textbf{DisenPOI} \cite{qin2023disenpoi}: contrastive disentangling of sequential-transition and geographical preference from two parallel graphs.
    \item \textbf{Diff-POI} \cite{qin2023diffpoi}: a graph-encoded diffusion process over latent spatial preference that samples the next-POI distribution, the strongest baseline on our dataset.
    \item \textbf{BiGSL} \cite{wang2024bigsl}: bi-level transition-graph learning over fine-grained POI nodes and coarse prototype clusters.
    \item \textbf{GNPR-SID} \cite{wang2025gnprsid}: hierarchical semantic-ID quantization of POIs with generative decoding, adapted to our full-catalog ranking protocol.
    \item \textbf{CAGNN} \cite{lei2025cagnn}: spatiotemporal-context-conditioned graph propagation and attention.
\item \textbf{Monthly updated variants}: updated baseline rows use the same monthly sequential update protocol as EviRec and update using prior data only.
\end{itemize}
\end{sloppypar}

\section{Extended Evaluation Results}

Tables~\ref{tab:appendix-boston-part1}--\ref{tab:appendix-new-york-part1} report the month-by-month results for each of the five cities across the full evaluation period (2025-05 to 2025-12).
Results are reported as the mean $\pm$ standard deviation across available seeds.
Within each table, bold and underlined values indicate the best and second-best reported results, respectively.

\begin{table*}[p]
\centering
\tiny
\caption{Monthly performance in Boston.}
\label{tab:appendix-boston-part1}
\resizebox{\textwidth}{!}{%
\begin{tabular}{llccccccccc}
\toprule
Month & Method & All N@10 & All R@10 & All N@20 & All R@20 & All N@30 & All R@30 & POI-New N@10 & User-New N@10 & Dual-New N@10 \\
\midrule
2025-05 & SASRec & 0.131 $\pm$ 0.000 & 0.158 $\pm$ 0.000 & 0.137 $\pm$ 0.000 & 0.183 $\pm$ 0.000 & 0.139 $\pm$ 0.000 & 0.193 $\pm$ 0.000 & 0.039 $\pm$ 0.000 & 0.149 $\pm$ 0.000 & 0.069 $\pm$ 0.000 \\
 & GETNext & 0.236 $\pm$ 0.000 & 0.291 $\pm$ 0.000 & 0.246 $\pm$ 0.000 & 0.331 $\pm$ 0.000 & 0.251 $\pm$ 0.000 & 0.356 $\pm$ 0.000 & 0.100 $\pm$ 0.000 & 0.239 $\pm$ 0.000 & 0.150 $\pm$ 0.000 \\
 & KBGNN & 0.237 $\pm$ 0.008 & 0.289 $\pm$ 0.003 & 0.246 $\pm$ 0.007 & 0.326 $\pm$ 0.002 & 0.252 $\pm$ 0.007 & 0.353 $\pm$ 0.003 & 0.107 $\pm$ 0.006 & 0.241 $\pm$ 0.012 & 0.157 $\pm$ 0.002 \\
 & DisenPOI & 0.228 $\pm$ 0.002 & 0.281 $\pm$ 0.005 & 0.238 $\pm$ 0.002 & 0.320 $\pm$ 0.005 & 0.244 $\pm$ 0.002 & 0.348 $\pm$ 0.007 & 0.087 $\pm$ 0.010 & 0.227 $\pm$ 0.008 & 0.122 $\pm$ 0.023 \\
 & Diff-POI & 0.240 $\pm$ 0.003 & 0.293 $\pm$ 0.004 & 0.249 $\pm$ 0.003 & 0.329 $\pm$ 0.005 & 0.254 $\pm$ 0.004 & 0.356 $\pm$ 0.007 & 0.101 $\pm$ 0.004 & \underline{0.251 $\pm$ 0.008} & 0.148 $\pm$ 0.005 \\
 & BiGSL & \underline{0.246 $\pm$ 0.002} & \underline{0.303 $\pm$ 0.007} & \underline{0.257 $\pm$ 0.002} & \underline{0.343 $\pm$ 0.005} & \underline{0.263 $\pm$ 0.001} & \underline{0.373 $\pm$ 0.002} & 0.103 $\pm$ 0.004 & 0.248 $\pm$ 0.003 & 0.156 $\pm$ 0.002 \\
 & GNPR-SID & 0.239 $\pm$ 0.012 & 0.289 $\pm$ 0.008 & 0.248 $\pm$ 0.010 & 0.324 $\pm$ 0.002 & 0.253 $\pm$ 0.010 & 0.349 $\pm$ 0.003 & \underline{0.108 $\pm$ 0.003} & 0.246 $\pm$ 0.015 & \underline{0.159 $\pm$ 0.002} \\
 & CAGNN & 0.231 $\pm$ 0.011 & 0.282 $\pm$ 0.007 & 0.241 $\pm$ 0.012 & 0.322 $\pm$ 0.012 & 0.247 $\pm$ 0.012 & 0.351 $\pm$ 0.011 & 0.100 $\pm$ 0.004 & 0.235 $\pm$ 0.008 & 0.151 $\pm$ 0.004 \\
 & \textbf{EviRec} & \textbf{0.258 $\pm$ 0.004} & \textbf{0.312 $\pm$ 0.004} & \textbf{0.268 $\pm$ 0.004} & \textbf{0.351 $\pm$ 0.006} & \textbf{0.274 $\pm$ 0.004} & \textbf{0.379 $\pm$ 0.008} & \textbf{0.150 $\pm$ 0.009} & \textbf{0.266 $\pm$ 0.007} & \textbf{0.206 $\pm$ 0.006} \\
\midrule
2025-06 & SASRec & 0.112 $\pm$ 0.000 & 0.134 $\pm$ 0.000 & 0.116 $\pm$ 0.000 & 0.148 $\pm$ 0.000 & 0.118 $\pm$ 0.000 & 0.158 $\pm$ 0.000 & 0.038 $\pm$ 0.000 & 0.098 $\pm$ 0.000 & 0.059 $\pm$ 0.000 \\
 & GETNext & 0.189 $\pm$ 0.000 & 0.245 $\pm$ 0.000 & 0.197 $\pm$ 0.000 & 0.278 $\pm$ 0.000 & 0.202 $\pm$ 0.000 & 0.302 $\pm$ 0.000 & 0.080 $\pm$ 0.000 & 0.161 $\pm$ 0.000 & 0.107 $\pm$ 0.000 \\
 & KBGNN & 0.189 $\pm$ 0.004 & 0.239 $\pm$ 0.007 & 0.198 $\pm$ 0.004 & 0.276 $\pm$ 0.006 & 0.204 $\pm$ 0.004 & 0.304 $\pm$ 0.008 & 0.083 $\pm$ 0.005 & 0.168 $\pm$ 0.007 & 0.112 $\pm$ 0.008 \\
 & DisenPOI & 0.200 $\pm$ 0.006 & 0.245 $\pm$ 0.010 & 0.209 $\pm$ 0.006 & 0.280 $\pm$ 0.010 & 0.215 $\pm$ 0.006 & 0.308 $\pm$ 0.010 & 0.078 $\pm$ 0.011 & 0.156 $\pm$ 0.010 & 0.098 $\pm$ 0.011 \\
 & Diff-POI & 0.203 $\pm$ 0.004 & 0.251 $\pm$ 0.005 & 0.212 $\pm$ 0.003 & 0.286 $\pm$ 0.004 & 0.217 $\pm$ 0.003 & 0.311 $\pm$ 0.004 & 0.082 $\pm$ 0.006 & 0.166 $\pm$ 0.010 & 0.108 $\pm$ 0.011 \\
 & BiGSL & \textbf{0.211 $\pm$ 0.004} & \underline{0.261 $\pm$ 0.003} & \textbf{0.221 $\pm$ 0.004} & \underline{0.300 $\pm$ 0.003} & \textbf{0.226 $\pm$ 0.004} & \underline{0.327 $\pm$ 0.003} & \underline{0.090 $\pm$ 0.012} & \underline{0.178 $\pm$ 0.006} & \underline{0.122 $\pm$ 0.014} \\
 & GNPR-SID & 0.191 $\pm$ 0.007 & 0.243 $\pm$ 0.010 & 0.200 $\pm$ 0.007 & 0.279 $\pm$ 0.010 & 0.206 $\pm$ 0.007 & 0.306 $\pm$ 0.008 & 0.085 $\pm$ 0.011 & 0.172 $\pm$ 0.010 & 0.120 $\pm$ 0.011 \\
 & CAGNN & 0.188 $\pm$ 0.014 & 0.239 $\pm$ 0.014 & 0.198 $\pm$ 0.013 & 0.277 $\pm$ 0.010 & 0.203 $\pm$ 0.013 & 0.301 $\pm$ 0.012 & 0.081 $\pm$ 0.012 & 0.164 $\pm$ 0.008 & 0.108 $\pm$ 0.010 \\
 & \textbf{EviRec} & \underline{0.210 $\pm$ 0.003} & \textbf{0.264 $\pm$ 0.010} & \underline{0.220 $\pm$ 0.004} & \textbf{0.303 $\pm$ 0.012} & \underline{0.225 $\pm$ 0.005} & \textbf{0.329 $\pm$ 0.015} & \textbf{0.122 $\pm$ 0.008} & \textbf{0.184 $\pm$ 0.005} & \textbf{0.147 $\pm$ 0.013} \\
\midrule
2025-07 & SASRec & 0.082 $\pm$ 0.000 & 0.109 $\pm$ 0.000 & 0.087 $\pm$ 0.000 & 0.126 $\pm$ 0.000 & 0.088 $\pm$ 0.000 & 0.132 $\pm$ 0.000 & 0.000 $\pm$ 0.000 & 0.067 $\pm$ 0.000 & 0.000 $\pm$ 0.000 \\
 & GETNext & 0.217 $\pm$ 0.000 & 0.263 $\pm$ 0.000 & 0.228 $\pm$ 0.000 & 0.305 $\pm$ 0.000 & 0.235 $\pm$ 0.000 & 0.339 $\pm$ 0.000 & 0.028 $\pm$ 0.000 & 0.284 $\pm$ 0.000 & 0.098 $\pm$ 0.000 \\
 & KBGNN & 0.214 $\pm$ 0.005 & 0.259 $\pm$ 0.004 & 0.222 $\pm$ 0.005 & 0.293 $\pm$ 0.006 & 0.228 $\pm$ 0.007 & 0.321 $\pm$ 0.014 & 0.038 $\pm$ 0.010 & 0.287 $\pm$ 0.004 & 0.129 $\pm$ 0.031 \\
 & DisenPOI & 0.215 $\pm$ 0.003 & 0.261 $\pm$ 0.004 & 0.225 $\pm$ 0.004 & 0.303 $\pm$ 0.008 & 0.232 $\pm$ 0.004 & 0.334 $\pm$ 0.010 & 0.040 $\pm$ 0.015 & 0.267 $\pm$ 0.018 & 0.129 $\pm$ 0.047 \\
 & Diff-POI & 0.216 $\pm$ 0.002 & 0.270 $\pm$ 0.003 & 0.225 $\pm$ 0.002 & 0.305 $\pm$ 0.004 & 0.231 $\pm$ 0.003 & 0.332 $\pm$ 0.008 & \textbf{0.054 $\pm$ 0.014} & 0.276 $\pm$ 0.010 & \textbf{0.159 $\pm$ 0.040} \\
 & BiGSL & \underline{0.227 $\pm$ 0.006} & \textbf{0.277 $\pm$ 0.009} & \textbf{0.238 $\pm$ 0.006} & \textbf{0.320 $\pm$ 0.009} & \textbf{0.245 $\pm$ 0.006} & \textbf{0.352 $\pm$ 0.009} & 0.041 $\pm$ 0.012 & 0.287 $\pm$ 0.010 & 0.122 $\pm$ 0.034 \\
 & GNPR-SID & 0.210 $\pm$ 0.003 & 0.253 $\pm$ 0.003 & 0.219 $\pm$ 0.004 & 0.287 $\pm$ 0.007 & 0.224 $\pm$ 0.005 & 0.312 $\pm$ 0.013 & 0.040 $\pm$ 0.006 & \underline{0.292 $\pm$ 0.009} & 0.134 $\pm$ 0.024 \\
 & CAGNN & 0.210 $\pm$ 0.008 & 0.258 $\pm$ 0.008 & 0.219 $\pm$ 0.009 & 0.293 $\pm$ 0.010 & 0.224 $\pm$ 0.009 & 0.318 $\pm$ 0.011 & 0.038 $\pm$ 0.010 & 0.280 $\pm$ 0.025 & 0.128 $\pm$ 0.034 \\
 & \textbf{EviRec} & \textbf{0.228 $\pm$ 0.008} & \underline{0.274 $\pm$ 0.013} & \underline{0.237 $\pm$ 0.008} & \underline{0.313 $\pm$ 0.013} & \underline{0.243 $\pm$ 0.007} & \underline{0.340 $\pm$ 0.010} & \underline{0.050 $\pm$ 0.002} & \textbf{0.298 $\pm$ 0.010} & \underline{0.155 $\pm$ 0.014} \\
\midrule
2025-08 & SASRec & 0.098 $\pm$ 0.000 & 0.119 $\pm$ 0.000 & 0.101 $\pm$ 0.000 & 0.134 $\pm$ 0.000 & 0.102 $\pm$ 0.000 & 0.139 $\pm$ 0.000 & 0.004 $\pm$ 0.000 & 0.141 $\pm$ 0.000 & 0.000 $\pm$ 0.000 \\
 & GETNext & 0.178 $\pm$ 0.000 & 0.221 $\pm$ 0.000 & 0.189 $\pm$ 0.000 & 0.264 $\pm$ 0.000 & 0.193 $\pm$ 0.000 & 0.286 $\pm$ 0.000 & 0.065 $\pm$ 0.000 & 0.256 $\pm$ 0.000 & 0.125 $\pm$ 0.000 \\
 & KBGNN & 0.175 $\pm$ 0.001 & 0.218 $\pm$ 0.003 & 0.185 $\pm$ 0.001 & 0.257 $\pm$ 0.005 & 0.191 $\pm$ 0.001 & 0.285 $\pm$ 0.007 & 0.074 $\pm$ 0.008 & 0.265 $\pm$ 0.012 & \underline{0.149 $\pm$ 0.020} \\
 & DisenPOI & 0.185 $\pm$ 0.003 & 0.231 $\pm$ 0.003 & 0.197 $\pm$ 0.003 & \underline{0.277 $\pm$ 0.003} & 0.203 $\pm$ 0.003 & \textbf{0.305 $\pm$ 0.002} & 0.064 $\pm$ 0.006 & 0.252 $\pm$ 0.015 & 0.116 $\pm$ 0.004 \\
 & Diff-POI & 0.192 $\pm$ 0.000 & 0.236 $\pm$ 0.005 & 0.201 $\pm$ 0.000 & 0.271 $\pm$ 0.006 & 0.207 $\pm$ 0.000 & 0.298 $\pm$ 0.004 & 0.079 $\pm$ 0.009 & \underline{0.275 $\pm$ 0.009} & 0.146 $\pm$ 0.022 \\
 & BiGSL & \underline{0.197 $\pm$ 0.006} & \underline{0.240 $\pm$ 0.002} & \underline{0.205 $\pm$ 0.006} & 0.273 $\pm$ 0.004 & \underline{0.211 $\pm$ 0.006} & 0.300 $\pm$ 0.006 & \underline{0.081 $\pm$ 0.005} & 0.269 $\pm$ 0.012 & 0.142 $\pm$ 0.034 \\
 & GNPR-SID & 0.176 $\pm$ 0.002 & 0.218 $\pm$ 0.004 & 0.184 $\pm$ 0.001 & 0.252 $\pm$ 0.001 & 0.191 $\pm$ 0.002 & 0.282 $\pm$ 0.006 & 0.074 $\pm$ 0.005 & 0.266 $\pm$ 0.004 & 0.146 $\pm$ 0.020 \\
 & CAGNN & 0.177 $\pm$ 0.010 & 0.216 $\pm$ 0.012 & 0.186 $\pm$ 0.009 & 0.251 $\pm$ 0.011 & 0.192 $\pm$ 0.010 & 0.280 $\pm$ 0.011 & 0.067 $\pm$ 0.014 & 0.261 $\pm$ 0.019 & 0.135 $\pm$ 0.025 \\
 & \textbf{EviRec} & \textbf{0.198 $\pm$ 0.005} & \textbf{0.244 $\pm$ 0.010} & \textbf{0.208 $\pm$ 0.006} & \textbf{0.281 $\pm$ 0.016} & \textbf{0.212 $\pm$ 0.007} & \underline{0.304 $\pm$ 0.021} & \textbf{0.113 $\pm$ 0.016} & \textbf{0.278 $\pm$ 0.008} & \textbf{0.192 $\pm$ 0.033} \\
\midrule
 2025-09 & SASRec & 0.163 $\pm$ 0.000 & 0.208 $\pm$ 0.000 & 0.170 $\pm$ 0.000 & 0.236 $\pm$ 0.000 & 0.174 $\pm$ 0.000 & 0.254 $\pm$ 0.000 & 0.061 $\pm$ 0.000 & 0.169 $\pm$ 0.000 & 0.067 $\pm$ 0.000 \\
 & GETNext & 0.412 $\pm$ 0.000 & 0.483 $\pm$ 0.000 & 0.422 $\pm$ 0.000 & 0.523 $\pm$ 0.000 & 0.427 $\pm$ 0.000 & 0.547 $\pm$ 0.000 & 0.293 $\pm$ 0.000 & 0.381 $\pm$ 0.000 & 0.278 $\pm$ 0.000 \\
 & KBGNN & 0.404 $\pm$ 0.003 & 0.479 $\pm$ 0.002 & 0.414 $\pm$ 0.003 & 0.522 $\pm$ 0.001 & 0.420 $\pm$ 0.002 & 0.550 $\pm$ 0.004 & 0.290 $\pm$ 0.004 & 0.377 $\pm$ 0.000 & 0.278 $\pm$ 0.001 \\
 & DisenPOI & \textbf{0.454 $\pm$ 0.006} & \textbf{0.522 $\pm$ 0.003} & \textbf{0.463 $\pm$ 0.005} & \textbf{0.559 $\pm$ 0.006} & \textbf{0.468 $\pm$ 0.005} & \textbf{0.582 $\pm$ 0.006} & \underline{0.328 $\pm$ 0.012} & \textbf{0.421 $\pm$ 0.009} & \underline{0.307 $\pm$ 0.021} \\
 & Diff-POI & 0.429 $\pm$ 0.006 & 0.502 $\pm$ 0.007 & 0.439 $\pm$ 0.005 & 0.541 $\pm$ 0.006 & 0.444 $\pm$ 0.005 & 0.566 $\pm$ 0.006 & 0.297 $\pm$ 0.002 & 0.400 $\pm$ 0.006 & 0.279 $\pm$ 0.010 \\
 & BiGSL & \underline{0.439 $\pm$ 0.011} & 0.507 $\pm$ 0.012 & \underline{0.449 $\pm$ 0.011} & 0.544 $\pm$ 0.012 & \underline{0.455 $\pm$ 0.011} & 0.571 $\pm$ 0.012 & 0.312 $\pm$ 0.015 & 0.409 $\pm$ 0.015 & 0.290 $\pm$ 0.015 \\
 & GNPR-SID & 0.406 $\pm$ 0.002 & 0.481 $\pm$ 0.002 & 0.417 $\pm$ 0.001 & 0.523 $\pm$ 0.003 & 0.423 $\pm$ 0.001 & 0.550 $\pm$ 0.006 & 0.293 $\pm$ 0.004 & 0.383 $\pm$ 0.002 & 0.281 $\pm$ 0.008 \\
 & CAGNN & 0.396 $\pm$ 0.006 & 0.474 $\pm$ 0.006 & 0.407 $\pm$ 0.006 & 0.516 $\pm$ 0.007 & 0.412 $\pm$ 0.005 & 0.541 $\pm$ 0.008 & 0.283 $\pm$ 0.015 & 0.374 $\pm$ 0.005 & 0.275 $\pm$ 0.007 \\
 & \textbf{EviRec} & 0.433 $\pm$ 0.002 & \underline{0.509 $\pm$ 0.004} & 0.444 $\pm$ 0.003 & \underline{0.554 $\pm$ 0.007} & 0.450 $\pm$ 0.003 & \underline{0.580 $\pm$ 0.008} & \textbf{0.367 $\pm$ 0.015} & \underline{0.409 $\pm$ 0.002} & \textbf{0.347 $\pm$ 0.017} \\
\midrule
2025-10 & SASRec & 0.090 $\pm$ 0.000 & 0.114 $\pm$ 0.000 & 0.093 $\pm$ 0.000 & 0.129 $\pm$ 0.000 & 0.095 $\pm$ 0.000 & 0.136 $\pm$ 0.000 & 0.028 $\pm$ 0.000 & 0.077 $\pm$ 0.000 & 0.053 $\pm$ 0.000 \\
 & GETNext & 0.260 $\pm$ 0.000 & 0.303 $\pm$ 0.000 & 0.269 $\pm$ 0.000 & 0.337 $\pm$ 0.000 & 0.275 $\pm$ 0.000 & 0.364 $\pm$ 0.000 & 0.145 $\pm$ 0.000 & 0.236 $\pm$ 0.000 & 0.253 $\pm$ 0.000 \\
 & KBGNN & 0.261 $\pm$ 0.000 & 0.308 $\pm$ 0.004 & 0.270 $\pm$ 0.000 & 0.342 $\pm$ 0.005 & 0.275 $\pm$ 0.000 & 0.366 $\pm$ 0.005 & 0.156 $\pm$ 0.005 & 0.235 $\pm$ 0.006 & 0.251 $\pm$ 0.015 \\
 & DisenPOI & 0.277 $\pm$ 0.001 & 0.325 $\pm$ 0.001 & 0.285 $\pm$ 0.002 & 0.357 $\pm$ 0.001 & 0.290 $\pm$ 0.002 & 0.380 $\pm$ 0.004 & 0.158 $\pm$ 0.005 & 0.241 $\pm$ 0.002 & 0.245 $\pm$ 0.007 \\
 & Diff-POI & 0.276 $\pm$ 0.002 & 0.323 $\pm$ 0.006 & 0.284 $\pm$ 0.002 & 0.355 $\pm$ 0.003 & 0.289 $\pm$ 0.001 & 0.376 $\pm$ 0.001 & 0.160 $\pm$ 0.009 & 0.240 $\pm$ 0.001 & 0.251 $\pm$ 0.007 \\
 & BiGSL & \underline{0.281 $\pm$ 0.004} & \underline{0.327 $\pm$ 0.004} & \underline{0.290 $\pm$ 0.005} & \underline{0.364 $\pm$ 0.006} & \underline{0.296 $\pm$ 0.005} & \textbf{0.389 $\pm$ 0.008} & \underline{0.175 $\pm$ 0.004} & \underline{0.248 $\pm$ 0.007} & \underline{0.273 $\pm$ 0.009} \\
 & GNPR-SID & 0.263 $\pm$ 0.002 & 0.309 $\pm$ 0.005 & 0.272 $\pm$ 0.002 & 0.341 $\pm$ 0.005 & 0.277 $\pm$ 0.002 & 0.365 $\pm$ 0.007 & 0.156 $\pm$ 0.008 & 0.238 $\pm$ 0.006 & 0.254 $\pm$ 0.006 \\
 & CAGNN & 0.260 $\pm$ 0.006 & 0.308 $\pm$ 0.010 & 0.268 $\pm$ 0.007 & 0.341 $\pm$ 0.013 & 0.273 $\pm$ 0.007 & 0.364 $\pm$ 0.013 & 0.154 $\pm$ 0.009 & 0.235 $\pm$ 0.003 & 0.251 $\pm$ 0.003 \\
 & \textbf{EviRec} & \textbf{0.284 $\pm$ 0.001} & \textbf{0.335 $\pm$ 0.004} & \textbf{0.291 $\pm$ 0.001} & \textbf{0.366 $\pm$ 0.005} & \textbf{0.296 $\pm$ 0.001} & \underline{0.388 $\pm$ 0.008} & \textbf{0.194 $\pm$ 0.004} & \textbf{0.258 $\pm$ 0.003} & \textbf{0.291 $\pm$ 0.006} \\
\midrule
2025-11 & SASRec & 0.053 $\pm$ 0.000 & 0.069 $\pm$ 0.000 & 0.056 $\pm$ 0.000 & 0.081 $\pm$ 0.000 & 0.057 $\pm$ 0.000 & 0.088 $\pm$ 0.000 & 0.003 $\pm$ 0.000 & 0.046 $\pm$ 0.000 & 0.007 $\pm$ 0.000 \\
 & GETNext & 0.188 $\pm$ 0.000 & 0.233 $\pm$ 0.000 & 0.198 $\pm$ 0.000 & 0.269 $\pm$ 0.000 & 0.202 $\pm$ 0.000 & 0.289 $\pm$ 0.000 & 0.071 $\pm$ 0.000 & 0.151 $\pm$ 0.000 & 0.080 $\pm$ 0.000 \\
 & KBGNN & 0.188 $\pm$ 0.007 & 0.237 $\pm$ 0.003 & 0.195 $\pm$ 0.006 & 0.267 $\pm$ 0.001 & 0.200 $\pm$ 0.006 & 0.289 $\pm$ 0.002 & 0.069 $\pm$ 0.007 & 0.155 $\pm$ 0.012 & 0.089 $\pm$ 0.012 \\
 & DisenPOI & 0.206 $\pm$ 0.007 & 0.253 $\pm$ 0.007 & 0.214 $\pm$ 0.006 & 0.284 $\pm$ 0.005 & 0.219 $\pm$ 0.006 & 0.308 $\pm$ 0.004 & 0.078 $\pm$ 0.006 & 0.166 $\pm$ 0.005 & 0.094 $\pm$ 0.006 \\
 & Diff-POI & 0.203 $\pm$ 0.004 & 0.253 $\pm$ 0.004 & 0.211 $\pm$ 0.004 & 0.286 $\pm$ 0.004 & 0.217 $\pm$ 0.004 & 0.311 $\pm$ 0.003 & 0.079 $\pm$ 0.005 & 0.171 $\pm$ 0.007 & 0.092 $\pm$ 0.005 \\
 & BiGSL & \underline{0.207 $\pm$ 0.011} & \underline{0.256 $\pm$ 0.011} & \underline{0.216 $\pm$ 0.011} & \underline{0.292 $\pm$ 0.009} & \underline{0.221 $\pm$ 0.010} & \underline{0.316 $\pm$ 0.007} & \underline{0.080 $\pm$ 0.004} & \underline{0.171 $\pm$ 0.013} & \underline{0.101 $\pm$ 0.015} \\
 & GNPR-SID & 0.192 $\pm$ 0.005 & 0.237 $\pm$ 0.005 & 0.200 $\pm$ 0.005 & 0.267 $\pm$ 0.004 & 0.204 $\pm$ 0.005 & 0.288 $\pm$ 0.004 & 0.074 $\pm$ 0.005 & 0.160 $\pm$ 0.008 & 0.100 $\pm$ 0.012 \\
 & CAGNN & 0.186 $\pm$ 0.007 & 0.233 $\pm$ 0.007 & 0.195 $\pm$ 0.007 & 0.266 $\pm$ 0.011 & 0.199 $\pm$ 0.007 & 0.287 $\pm$ 0.010 & 0.078 $\pm$ 0.003 & 0.158 $\pm$ 0.009 & 0.097 $\pm$ 0.010 \\
 & \textbf{EviRec} & \textbf{0.216 $\pm$ 0.002} & \textbf{0.264 $\pm$ 0.004} & \textbf{0.224 $\pm$ 0.002} & \textbf{0.295 $\pm$ 0.005} & \textbf{0.228 $\pm$ 0.003} & \textbf{0.318 $\pm$ 0.008} & \textbf{0.112 $\pm$ 0.002} & \textbf{0.180 $\pm$ 0.006} & \textbf{0.131 $\pm$ 0.013} \\
\midrule
2025-12 & SASRec & 0.091 $\pm$ 0.000 & 0.120 $\pm$ 0.000 & 0.096 $\pm$ 0.000 & 0.141 $\pm$ 0.000 & 0.098 $\pm$ 0.000 & 0.152 $\pm$ 0.000 & 0.000 $\pm$ 0.000 & 0.087 $\pm$ 0.000 & 0.000 $\pm$ 0.000 \\
 & GETNext & 0.292 $\pm$ 0.000 & 0.340 $\pm$ 0.000 & 0.301 $\pm$ 0.000 & 0.377 $\pm$ 0.000 & 0.307 $\pm$ 0.000 & 0.401 $\pm$ 0.000 & 0.030 $\pm$ 0.000 & 0.248 $\pm$ 0.000 & \underline{0.053 $\pm$ 0.000} \\
 & KBGNN & 0.293 $\pm$ 0.003 & 0.341 $\pm$ 0.002 & 0.301 $\pm$ 0.003 & 0.371 $\pm$ 0.004 & 0.305 $\pm$ 0.002 & 0.391 $\pm$ 0.005 & 0.040 $\pm$ 0.020 & 0.239 $\pm$ 0.007 & 0.053 $\pm$ 0.000 \\
 & DisenPOI & 0.305 $\pm$ 0.004 & 0.355 $\pm$ 0.005 & 0.313 $\pm$ 0.003 & 0.388 $\pm$ 0.003 & 0.318 $\pm$ 0.004 & 0.412 $\pm$ 0.003 & 0.039 $\pm$ 0.018 & 0.240 $\pm$ 0.015 & 0.053 $\pm$ 0.000 \\
 & Diff-POI & \underline{0.309 $\pm$ 0.002} & 0.359 $\pm$ 0.004 & \underline{0.317 $\pm$ 0.001} & 0.389 $\pm$ 0.005 & \underline{0.322 $\pm$ 0.002} & \underline{0.412 $\pm$ 0.005} & 0.038 $\pm$ 0.016 & \textbf{0.256 $\pm$ 0.009} & 0.053 $\pm$ 0.000 \\
 & BiGSL & 0.308 $\pm$ 0.003 & \underline{0.361 $\pm$ 0.006} & 0.316 $\pm$ 0.003 & \textbf{0.392 $\pm$ 0.003} & 0.321 $\pm$ 0.002 & \textbf{0.418 $\pm$ 0.003} & 0.036 $\pm$ 0.012 & 0.247 $\pm$ 0.001 & 0.053 $\pm$ 0.000 \\
 & GNPR-SID & 0.289 $\pm$ 0.008 & 0.338 $\pm$ 0.012 & 0.297 $\pm$ 0.009 & 0.368 $\pm$ 0.015 & 0.302 $\pm$ 0.007 & 0.393 $\pm$ 0.007 & \underline{0.040 $\pm$ 0.013} & 0.236 $\pm$ 0.013 & 0.053 $\pm$ 0.000 \\
 & CAGNN & 0.291 $\pm$ 0.010 & 0.337 $\pm$ 0.011 & 0.299 $\pm$ 0.009 & 0.366 $\pm$ 0.008 & 0.303 $\pm$ 0.010 & 0.386 $\pm$ 0.013 & 0.035 $\pm$ 0.019 & 0.247 $\pm$ 0.014 & 0.046 $\pm$ 0.011 \\
 & \textbf{EviRec} & \textbf{0.317 $\pm$ 0.003} & \textbf{0.366 $\pm$ 0.002} & \textbf{0.324 $\pm$ 0.004} & \underline{0.392 $\pm$ 0.002} & \textbf{0.328 $\pm$ 0.004} & 0.411 $\pm$ 0.003 & \textbf{0.078 $\pm$ 0.003} & \underline{0.254 $\pm$ 0.009} & \textbf{0.058 $\pm$ 0.010} \\
\bottomrule
\end{tabular}}
\end{table*}

\begin{table*}[p]
\centering
\tiny
\caption{Monthly performance in Chicago.}
\label{tab:appendix-chicago-part1}
\resizebox{\textwidth}{!}{%
\begin{tabular}{llccccccccc}
\toprule
Month & Method & All N@10 & All R@10 & All N@20 & All R@20 & All N@30 & All R@30 & POI-New N@10 & User-New N@10 & Dual-New N@10 \\
\midrule
2025-05 & SASRec & 0.089 $\pm$ 0.000 & 0.111 $\pm$ 0.000 & 0.093 $\pm$ 0.000 & 0.126 $\pm$ 0.000 & 0.094 $\pm$ 0.000 & 0.134 $\pm$ 0.000 & 0.023 $\pm$ 0.000 & 0.107 $\pm$ 0.000 & 0.047 $\pm$ 0.000 \\
 & GETNext & \underline{0.210 $\pm$ 0.000} & \underline{0.257 $\pm$ 0.000} & \underline{0.219 $\pm$ 0.000} & \underline{0.289 $\pm$ 0.000} & \underline{0.223 $\pm$ 0.000} & 0.308 $\pm$ 0.000 & \underline{0.099 $\pm$ 0.000} & \underline{0.221 $\pm$ 0.000} & \underline{0.151 $\pm$ 0.000} \\
 & KBGNN & 0.203 $\pm$ 0.006 & 0.249 $\pm$ 0.007 & 0.211 $\pm$ 0.006 & 0.278 $\pm$ 0.007 & 0.215 $\pm$ 0.006 & 0.299 $\pm$ 0.007 & 0.094 $\pm$ 0.004 & 0.215 $\pm$ 0.006 & 0.148 $\pm$ 0.007 \\
 & DisenPOI & 0.182 $\pm$ 0.006 & 0.232 $\pm$ 0.003 & 0.190 $\pm$ 0.005 & 0.265 $\pm$ 0.002 & 0.195 $\pm$ 0.005 & 0.289 $\pm$ 0.002 & 0.078 $\pm$ 0.003 & 0.188 $\pm$ 0.005 & 0.114 $\pm$ 0.006 \\
 & Diff-POI & 0.196 $\pm$ 0.000 & 0.246 $\pm$ 0.002 & 0.204 $\pm$ 0.001 & 0.278 $\pm$ 0.004 & 0.209 $\pm$ 0.001 & 0.299 $\pm$ 0.004 & 0.093 $\pm$ 0.003 & 0.210 $\pm$ 0.001 & 0.146 $\pm$ 0.006 \\
 & BiGSL & 0.202 $\pm$ 0.004 & 0.255 $\pm$ 0.004 & 0.210 $\pm$ 0.004 & 0.288 $\pm$ 0.002 & 0.215 $\pm$ 0.004 & \underline{0.312 $\pm$ 0.003} & 0.089 $\pm$ 0.003 & 0.210 $\pm$ 0.007 & 0.137 $\pm$ 0.004 \\
 & GNPR-SID & 0.203 $\pm$ 0.003 & 0.248 $\pm$ 0.003 & 0.210 $\pm$ 0.003 & 0.279 $\pm$ 0.002 & 0.215 $\pm$ 0.003 & 0.300 $\pm$ 0.002 & 0.094 $\pm$ 0.001 & 0.215 $\pm$ 0.002 & 0.147 $\pm$ 0.005 \\
 & CAGNN & 0.197 $\pm$ 0.002 & 0.243 $\pm$ 0.004 & 0.205 $\pm$ 0.002 & 0.275 $\pm$ 0.005 & 0.210 $\pm$ 0.002 & 0.297 $\pm$ 0.007 & 0.087 $\pm$ 0.001 & 0.208 $\pm$ 0.004 & 0.139 $\pm$ 0.005 \\
 & \textbf{EviRec} & \textbf{0.225 $\pm$ 0.004} & \textbf{0.274 $\pm$ 0.006} & \textbf{0.233 $\pm$ 0.004} & \textbf{0.305 $\pm$ 0.007} & \textbf{0.237 $\pm$ 0.003} & \textbf{0.325 $\pm$ 0.007} & \textbf{0.142 $\pm$ 0.008} & \textbf{0.242 $\pm$ 0.005} & \textbf{0.198 $\pm$ 0.007} \\
\midrule
2025-06 & SASRec & 0.068 $\pm$ 0.000 & 0.083 $\pm$ 0.000 & 0.070 $\pm$ 0.000 & 0.092 $\pm$ 0.000 & 0.072 $\pm$ 0.000 & 0.098 $\pm$ 0.000 & 0.021 $\pm$ 0.000 & 0.051 $\pm$ 0.000 & 0.029 $\pm$ 0.000 \\
 & GETNext & 0.178 $\pm$ 0.000 & 0.221 $\pm$ 0.000 & 0.185 $\pm$ 0.000 & 0.248 $\pm$ 0.000 & 0.189 $\pm$ 0.000 & 0.268 $\pm$ 0.000 & \underline{0.089 $\pm$ 0.000} & 0.127 $\pm$ 0.000 & 0.103 $\pm$ 0.000 \\
 & KBGNN & 0.175 $\pm$ 0.003 & 0.217 $\pm$ 0.004 & 0.182 $\pm$ 0.003 & 0.243 $\pm$ 0.005 & 0.186 $\pm$ 0.003 & 0.262 $\pm$ 0.005 & 0.085 $\pm$ 0.004 & 0.123 $\pm$ 0.003 & 0.102 $\pm$ 0.002 \\
 & DisenPOI & 0.160 $\pm$ 0.005 & 0.201 $\pm$ 0.006 & 0.167 $\pm$ 0.005 & 0.228 $\pm$ 0.005 & 0.171 $\pm$ 0.005 & 0.249 $\pm$ 0.005 & 0.070 $\pm$ 0.004 & 0.102 $\pm$ 0.003 & 0.077 $\pm$ 0.002 \\
 & Diff-POI & 0.167 $\pm$ 0.003 & 0.209 $\pm$ 0.003 & 0.174 $\pm$ 0.004 & 0.238 $\pm$ 0.004 & 0.178 $\pm$ 0.003 & 0.257 $\pm$ 0.003 & 0.077 $\pm$ 0.005 & 0.111 $\pm$ 0.004 & 0.085 $\pm$ 0.003 \\
 & BiGSL & \underline{0.185 $\pm$ 0.003} & \underline{0.230 $\pm$ 0.003} & \underline{0.192 $\pm$ 0.002} & \underline{0.260 $\pm$ 0.001} & \underline{0.197 $\pm$ 0.002} & \underline{0.280 $\pm$ 0.001} & 0.089 $\pm$ 0.001 & \underline{0.130 $\pm$ 0.002} & 0.105 $\pm$ 0.005 \\
 & GNPR-SID & 0.178 $\pm$ 0.003 & 0.221 $\pm$ 0.002 & 0.185 $\pm$ 0.003 & 0.246 $\pm$ 0.002 & 0.189 $\pm$ 0.003 & 0.264 $\pm$ 0.002 & 0.089 $\pm$ 0.001 & 0.129 $\pm$ 0.001 & \underline{0.107 $\pm$ 0.001} \\
 & CAGNN & 0.176 $\pm$ 0.001 & 0.220 $\pm$ 0.001 & 0.183 $\pm$ 0.001 & 0.246 $\pm$ 0.001 & 0.187 $\pm$ 0.001 & 0.265 $\pm$ 0.000 & 0.087 $\pm$ 0.001 & 0.127 $\pm$ 0.002 & 0.106 $\pm$ 0.003 \\
 & \textbf{EviRec} & \textbf{0.196 $\pm$ 0.004} & \textbf{0.242 $\pm$ 0.005} & \textbf{0.203 $\pm$ 0.004} & \textbf{0.271 $\pm$ 0.006} & \textbf{0.207 $\pm$ 0.004} & \textbf{0.292 $\pm$ 0.006} & \textbf{0.126 $\pm$ 0.003} & \textbf{0.145 $\pm$ 0.005} & \textbf{0.145 $\pm$ 0.006} \\
\midrule
2025-07 & SASRec & 0.069 $\pm$ 0.000 & 0.086 $\pm$ 0.000 & 0.072 $\pm$ 0.000 & 0.097 $\pm$ 0.000 & 0.073 $\pm$ 0.000 & 0.103 $\pm$ 0.000 & 0.012 $\pm$ 0.000 & 0.048 $\pm$ 0.000 & 0.030 $\pm$ 0.000 \\
 & GETNext & 0.201 $\pm$ 0.000 & 0.242 $\pm$ 0.000 & 0.208 $\pm$ 0.000 & 0.271 $\pm$ 0.000 & 0.213 $\pm$ 0.000 & 0.293 $\pm$ 0.000 & 0.066 $\pm$ 0.000 & 0.152 $\pm$ 0.000 & 0.093 $\pm$ 0.000 \\
 & KBGNN & 0.200 $\pm$ 0.003 & 0.240 $\pm$ 0.003 & 0.207 $\pm$ 0.003 & 0.269 $\pm$ 0.003 & 0.211 $\pm$ 0.003 & 0.286 $\pm$ 0.002 & 0.074 $\pm$ 0.007 & 0.154 $\pm$ 0.001 & 0.101 $\pm$ 0.010 \\
 & DisenPOI & 0.182 $\pm$ 0.001 & 0.223 $\pm$ 0.001 & 0.188 $\pm$ 0.001 & 0.248 $\pm$ 0.001 & 0.192 $\pm$ 0.001 & 0.267 $\pm$ 0.001 & 0.068 $\pm$ 0.005 & 0.124 $\pm$ 0.002 & 0.095 $\pm$ 0.010 \\
 & Diff-POI & 0.185 $\pm$ 0.003 & 0.226 $\pm$ 0.004 & 0.192 $\pm$ 0.003 & 0.254 $\pm$ 0.005 & 0.196 $\pm$ 0.003 & 0.273 $\pm$ 0.006 & 0.066 $\pm$ 0.006 & 0.133 $\pm$ 0.003 & 0.092 $\pm$ 0.016 \\
 & BiGSL & \underline{0.204 $\pm$ 0.007} & \underline{0.249 $\pm$ 0.010} & \underline{0.211 $\pm$ 0.007} & \underline{0.276 $\pm$ 0.009} & \underline{0.215 $\pm$ 0.008} & \underline{0.295 $\pm$ 0.011} & \underline{0.080 $\pm$ 0.007} & \underline{0.158 $\pm$ 0.009} & \underline{0.112 $\pm$ 0.004} \\
 & GNPR-SID & 0.197 $\pm$ 0.002 & 0.237 $\pm$ 0.002 & 0.204 $\pm$ 0.002 & 0.265 $\pm$ 0.001 & 0.207 $\pm$ 0.002 & 0.282 $\pm$ 0.001 & 0.073 $\pm$ 0.006 & 0.149 $\pm$ 0.005 & 0.098 $\pm$ 0.010 \\
 & CAGNN & 0.193 $\pm$ 0.001 & 0.233 $\pm$ 0.001 & 0.199 $\pm$ 0.001 & 0.261 $\pm$ 0.002 & 0.204 $\pm$ 0.001 & 0.280 $\pm$ 0.002 & 0.072 $\pm$ 0.002 & 0.144 $\pm$ 0.005 & 0.105 $\pm$ 0.011 \\
 & \textbf{EviRec} & \textbf{0.216 $\pm$ 0.003} & \textbf{0.262 $\pm$ 0.004} & \textbf{0.223 $\pm$ 0.003} & \textbf{0.290 $\pm$ 0.006} & \textbf{0.228 $\pm$ 0.003} & \textbf{0.309 $\pm$ 0.004} & \textbf{0.105 $\pm$ 0.005} & \textbf{0.172 $\pm$ 0.005} & \textbf{0.145 $\pm$ 0.011} \\
\midrule
2025-08 & SASRec & 0.079 $\pm$ 0.000 & 0.095 $\pm$ 0.000 & 0.081 $\pm$ 0.000 & 0.105 $\pm$ 0.000 & 0.082 $\pm$ 0.000 & 0.111 $\pm$ 0.000 & 0.010 $\pm$ 0.000 & 0.088 $\pm$ 0.000 & 0.026 $\pm$ 0.000 \\
 & GETNext & 0.194 $\pm$ 0.000 & 0.237 $\pm$ 0.000 & 0.202 $\pm$ 0.000 & 0.268 $\pm$ 0.000 & 0.206 $\pm$ 0.000 & 0.287 $\pm$ 0.000 & 0.080 $\pm$ 0.000 & 0.176 $\pm$ 0.000 & 0.083 $\pm$ 0.000 \\
 & KBGNN & 0.196 $\pm$ 0.004 & 0.242 $\pm$ 0.007 & 0.203 $\pm$ 0.003 & 0.270 $\pm$ 0.005 & 0.208 $\pm$ 0.003 & 0.290 $\pm$ 0.006 & 0.082 $\pm$ 0.007 & 0.177 $\pm$ 0.001 & 0.090 $\pm$ 0.007 \\
 & DisenPOI & 0.182 $\pm$ 0.004 & 0.221 $\pm$ 0.005 & 0.189 $\pm$ 0.003 & 0.250 $\pm$ 0.004 & 0.193 $\pm$ 0.004 & 0.271 $\pm$ 0.005 & 0.077 $\pm$ 0.002 & 0.174 $\pm$ 0.004 & 0.094 $\pm$ 0.010 \\
 & Diff-POI & 0.190 $\pm$ 0.004 & 0.234 $\pm$ 0.004 & 0.198 $\pm$ 0.004 & 0.265 $\pm$ 0.005 & 0.202 $\pm$ 0.004 & 0.286 $\pm$ 0.007 & 0.076 $\pm$ 0.003 & 0.180 $\pm$ 0.002 & \underline{0.099 $\pm$ 0.005} \\
 & BiGSL & \underline{0.210 $\pm$ 0.004} & \underline{0.258 $\pm$ 0.006} & \underline{0.218 $\pm$ 0.003} & \underline{0.290 $\pm$ 0.004} & \underline{0.222 $\pm$ 0.003} & \underline{0.309 $\pm$ 0.005} & \underline{0.093 $\pm$ 0.010} & \underline{0.184 $\pm$ 0.001} & 0.093 $\pm$ 0.002 \\
 & GNPR-SID & 0.198 $\pm$ 0.005 & 0.244 $\pm$ 0.007 & 0.205 $\pm$ 0.005 & 0.273 $\pm$ 0.006 & 0.210 $\pm$ 0.005 & 0.293 $\pm$ 0.004 & 0.081 $\pm$ 0.010 & 0.179 $\pm$ 0.001 & 0.094 $\pm$ 0.007 \\
 & CAGNN & 0.196 $\pm$ 0.007 & 0.241 $\pm$ 0.009 & 0.203 $\pm$ 0.007 & 0.269 $\pm$ 0.011 & 0.207 $\pm$ 0.007 & 0.288 $\pm$ 0.012 & 0.083 $\pm$ 0.000 & 0.178 $\pm$ 0.006 & 0.097 $\pm$ 0.006 \\
 & \textbf{EviRec} & \textbf{0.226 $\pm$ 0.002} & \textbf{0.281 $\pm$ 0.005} & \textbf{0.234 $\pm$ 0.002} & \textbf{0.310 $\pm$ 0.004} & \textbf{0.238 $\pm$ 0.002} & \textbf{0.332 $\pm$ 0.005} & \textbf{0.130 $\pm$ 0.012} & \textbf{0.201 $\pm$ 0.002} & \textbf{0.126 $\pm$ 0.005} \\
\midrule
2025-09 & SASRec & 0.095 $\pm$ 0.000 & 0.123 $\pm$ 0.000 & 0.100 $\pm$ 0.000 & 0.143 $\pm$ 0.000 & 0.102 $\pm$ 0.000 & 0.155 $\pm$ 0.000 & 0.026 $\pm$ 0.000 & 0.115 $\pm$ 0.000 & 0.043 $\pm$ 0.000 \\
 & GETNext & 0.331 $\pm$ 0.000 & 0.397 $\pm$ 0.000 & 0.341 $\pm$ 0.000 & 0.435 $\pm$ 0.000 & 0.346 $\pm$ 0.000 & 0.459 $\pm$ 0.000 & 0.204 $\pm$ 0.000 & 0.300 $\pm$ 0.000 & 0.219 $\pm$ 0.000 \\
 & KBGNN & 0.340 $\pm$ 0.008 & 0.405 $\pm$ 0.007 & 0.349 $\pm$ 0.008 & 0.442 $\pm$ 0.006 & 0.354 $\pm$ 0.008 & 0.465 $\pm$ 0.006 & 0.215 $\pm$ 0.012 & 0.302 $\pm$ 0.003 & 0.219 $\pm$ 0.007 \\
 & DisenPOI & \textbf{0.387 $\pm$ 0.005} & \textbf{0.450 $\pm$ 0.003} & \textbf{0.395 $\pm$ 0.004} & \textbf{0.483 $\pm$ 0.001} & \textbf{0.399 $\pm$ 0.004} & \textbf{0.503 $\pm$ 0.001} & \underline{0.272 $\pm$ 0.006} & \underline{0.326 $\pm$ 0.003} & \underline{0.245 $\pm$ 0.008} \\
 & Diff-POI & 0.350 $\pm$ 0.004 & 0.419 $\pm$ 0.003 & 0.359 $\pm$ 0.003 & 0.456 $\pm$ 0.001 & 0.364 $\pm$ 0.003 & 0.479 $\pm$ 0.001 & 0.231 $\pm$ 0.007 & 0.307 $\pm$ 0.002 & 0.224 $\pm$ 0.006 \\
 & BiGSL & 0.363 $\pm$ 0.003 & 0.429 $\pm$ 0.002 & 0.372 $\pm$ 0.003 & 0.464 $\pm$ 0.002 & 0.377 $\pm$ 0.003 & 0.487 $\pm$ 0.003 & 0.240 $\pm$ 0.002 & 0.316 $\pm$ 0.003 & 0.231 $\pm$ 0.010 \\
 & GNPR-SID & 0.338 $\pm$ 0.007 & 0.403 $\pm$ 0.006 & 0.347 $\pm$ 0.007 & 0.438 $\pm$ 0.005 & 0.352 $\pm$ 0.006 & 0.461 $\pm$ 0.004 & 0.214 $\pm$ 0.008 & 0.302 $\pm$ 0.002 & 0.219 $\pm$ 0.002 \\
 & CAGNN & 0.329 $\pm$ 0.007 & 0.396 $\pm$ 0.006 & 0.338 $\pm$ 0.007 & 0.433 $\pm$ 0.006 & 0.343 $\pm$ 0.007 & 0.457 $\pm$ 0.005 & 0.208 $\pm$ 0.011 & 0.297 $\pm$ 0.003 & 0.215 $\pm$ 0.012 \\
 & \textbf{EviRec} & \underline{0.374 $\pm$ 0.004} & \underline{0.444 $\pm$ 0.005} & \underline{0.383 $\pm$ 0.004} & \underline{0.479 $\pm$ 0.004} & \underline{0.388 $\pm$ 0.004} & \underline{0.501 $\pm$ 0.003} & \textbf{0.292 $\pm$ 0.004} & \textbf{0.329 $\pm$ 0.002} & \textbf{0.275 $\pm$ 0.003} \\
\midrule
2025-10 & SASRec & 0.042 $\pm$ 0.000 & 0.054 $\pm$ 0.000 & 0.043 $\pm$ 0.000 & 0.061 $\pm$ 0.000 & 0.044 $\pm$ 0.000 & 0.065 $\pm$ 0.000 & 0.009 $\pm$ 0.000 & 0.037 $\pm$ 0.000 & 0.020 $\pm$ 0.000 \\
 & GETNext & 0.168 $\pm$ 0.000 & 0.207 $\pm$ 0.000 & 0.175 $\pm$ 0.000 & 0.233 $\pm$ 0.000 & 0.178 $\pm$ 0.000 & 0.251 $\pm$ 0.000 & 0.083 $\pm$ 0.000 & 0.137 $\pm$ 0.000 & 0.093 $\pm$ 0.000 \\
 & KBGNN & 0.168 $\pm$ 0.004 & 0.205 $\pm$ 0.007 & 0.174 $\pm$ 0.004 & 0.230 $\pm$ 0.007 & 0.178 $\pm$ 0.004 & 0.248 $\pm$ 0.008 & 0.081 $\pm$ 0.003 & 0.136 $\pm$ 0.004 & 0.091 $\pm$ 0.006 \\
 & DisenPOI & 0.171 $\pm$ 0.004 & 0.207 $\pm$ 0.007 & 0.178 $\pm$ 0.004 & 0.232 $\pm$ 0.008 & 0.181 $\pm$ 0.004 & 0.250 $\pm$ 0.008 & 0.086 $\pm$ 0.008 & 0.140 $\pm$ 0.002 & 0.095 $\pm$ 0.000 \\
 & Diff-POI & 0.167 $\pm$ 0.000 & 0.203 $\pm$ 0.002 & 0.174 $\pm$ 0.000 & 0.229 $\pm$ 0.002 & 0.177 $\pm$ 0.000 & 0.246 $\pm$ 0.003 & 0.080 $\pm$ 0.002 & 0.140 $\pm$ 0.003 & 0.094 $\pm$ 0.008 \\
 & BiGSL & \underline{0.180 $\pm$ 0.001} & \underline{0.220 $\pm$ 0.002} & \underline{0.186 $\pm$ 0.001} & \underline{0.247 $\pm$ 0.003} & \underline{0.190 $\pm$ 0.001} & \underline{0.265 $\pm$ 0.003} & \underline{0.092 $\pm$ 0.001} & \underline{0.145 $\pm$ 0.003} & \underline{0.098 $\pm$ 0.002} \\
 & GNPR-SID & 0.166 $\pm$ 0.004 & 0.203 $\pm$ 0.006 & 0.172 $\pm$ 0.004 & 0.228 $\pm$ 0.007 & 0.175 $\pm$ 0.004 & 0.244 $\pm$ 0.007 & 0.083 $\pm$ 0.002 & 0.135 $\pm$ 0.001 & 0.092 $\pm$ 0.004 \\
 & CAGNN & 0.162 $\pm$ 0.005 & 0.200 $\pm$ 0.007 & 0.168 $\pm$ 0.005 & 0.225 $\pm$ 0.008 & 0.172 $\pm$ 0.005 & 0.243 $\pm$ 0.009 & 0.076 $\pm$ 0.004 & 0.132 $\pm$ 0.004 & 0.087 $\pm$ 0.003 \\
 & \textbf{EviRec} & \textbf{0.184 $\pm$ 0.004} & \textbf{0.226 $\pm$ 0.006} & \textbf{0.190 $\pm$ 0.004} & \textbf{0.251 $\pm$ 0.006} & \textbf{0.194 $\pm$ 0.004} & \textbf{0.268 $\pm$ 0.005} & \textbf{0.118 $\pm$ 0.002} & \textbf{0.152 $\pm$ 0.001} & \textbf{0.126 $\pm$ 0.002} \\
\midrule
2025-11 & SASRec & 0.038 $\pm$ 0.000 & 0.047 $\pm$ 0.000 & 0.040 $\pm$ 0.000 & 0.054 $\pm$ 0.000 & 0.041 $\pm$ 0.000 & 0.058 $\pm$ 0.000 & 0.009 $\pm$ 0.000 & 0.037 $\pm$ 0.000 & 0.020 $\pm$ 0.000 \\
 & GETNext & 0.163 $\pm$ 0.000 & 0.203 $\pm$ 0.000 & 0.169 $\pm$ 0.000 & 0.229 $\pm$ 0.000 & 0.174 $\pm$ 0.000 & 0.248 $\pm$ 0.000 & 0.075 $\pm$ 0.000 & 0.149 $\pm$ 0.000 & 0.111 $\pm$ 0.000 \\
 & KBGNN & 0.165 $\pm$ 0.003 & 0.204 $\pm$ 0.003 & 0.172 $\pm$ 0.003 & 0.230 $\pm$ 0.003 & 0.176 $\pm$ 0.003 & 0.248 $\pm$ 0.003 & 0.078 $\pm$ 0.006 & 0.147 $\pm$ 0.001 & 0.115 $\pm$ 0.008 \\
 & DisenPOI & 0.170 $\pm$ 0.004 & 0.208 $\pm$ 0.005 & 0.177 $\pm$ 0.003 & 0.233 $\pm$ 0.004 & 0.181 $\pm$ 0.003 & 0.252 $\pm$ 0.003 & 0.083 $\pm$ 0.004 & 0.147 $\pm$ 0.005 & 0.116 $\pm$ 0.005 \\
 & Diff-POI & 0.168 $\pm$ 0.002 & 0.206 $\pm$ 0.002 & 0.174 $\pm$ 0.002 & 0.232 $\pm$ 0.003 & 0.178 $\pm$ 0.002 & 0.250 $\pm$ 0.004 & 0.080 $\pm$ 0.002 & 0.150 $\pm$ 0.004 & 0.117 $\pm$ 0.006 \\
 & BiGSL & \underline{0.176 $\pm$ 0.006} & \underline{0.216 $\pm$ 0.007} & \underline{0.183 $\pm$ 0.006} & \underline{0.243 $\pm$ 0.008} & \underline{0.187 $\pm$ 0.006} & \underline{0.262 $\pm$ 0.008} & \underline{0.084 $\pm$ 0.007} & \underline{0.155 $\pm$ 0.008} & \underline{0.119 $\pm$ 0.012} \\
 & GNPR-SID & 0.166 $\pm$ 0.003 & 0.205 $\pm$ 0.002 & 0.173 $\pm$ 0.003 & 0.230 $\pm$ 0.002 & 0.176 $\pm$ 0.003 & 0.247 $\pm$ 0.002 & 0.082 $\pm$ 0.004 & 0.149 $\pm$ 0.002 & 0.118 $\pm$ 0.003 \\
 & CAGNN & 0.165 $\pm$ 0.005 & 0.204 $\pm$ 0.005 & 0.171 $\pm$ 0.005 & 0.230 $\pm$ 0.005 & 0.175 $\pm$ 0.005 & 0.247 $\pm$ 0.005 & 0.078 $\pm$ 0.004 & 0.148 $\pm$ 0.004 & 0.117 $\pm$ 0.006 \\
 & \textbf{EviRec} & \textbf{0.187 $\pm$ 0.002} & \textbf{0.230 $\pm$ 0.002} & \textbf{0.194 $\pm$ 0.002} & \textbf{0.257 $\pm$ 0.001} & \textbf{0.198 $\pm$ 0.002} & \textbf{0.275 $\pm$ 0.002} & \textbf{0.120 $\pm$ 0.003} & \textbf{0.167 $\pm$ 0.001} & \textbf{0.158 $\pm$ 0.003} \\
\midrule
2025-12 & SASRec & 0.080 $\pm$ 0.000 & 0.099 $\pm$ 0.000 & 0.083 $\pm$ 0.000 & 0.111 $\pm$ 0.000 & 0.084 $\pm$ 0.000 & 0.118 $\pm$ 0.000 & 0.011 $\pm$ 0.000 & 0.068 $\pm$ 0.000 & 0.027 $\pm$ 0.000 \\
 & GETNext & 0.269 $\pm$ 0.000 & 0.320 $\pm$ 0.000 & 0.276 $\pm$ 0.000 & 0.348 $\pm$ 0.000 & 0.280 $\pm$ 0.000 & 0.367 $\pm$ 0.000 & 0.091 $\pm$ 0.000 & 0.150 $\pm$ 0.000 & 0.088 $\pm$ 0.000 \\
 & KBGNN & 0.265 $\pm$ 0.006 & 0.316 $\pm$ 0.006 & 0.272 $\pm$ 0.006 & 0.346 $\pm$ 0.005 & 0.276 $\pm$ 0.006 & 0.363 $\pm$ 0.006 & 0.094 $\pm$ 0.004 & 0.146 $\pm$ 0.003 & 0.088 $\pm$ 0.004 \\
 & DisenPOI & 0.271 $\pm$ 0.006 & 0.320 $\pm$ 0.007 & 0.278 $\pm$ 0.006 & 0.348 $\pm$ 0.005 & 0.281 $\pm$ 0.005 & 0.366 $\pm$ 0.004 & 0.097 $\pm$ 0.008 & 0.148 $\pm$ 0.004 & 0.086 $\pm$ 0.002 \\
 & Diff-POI & 0.266 $\pm$ 0.004 & 0.318 $\pm$ 0.004 & 0.273 $\pm$ 0.004 & 0.348 $\pm$ 0.004 & 0.277 $\pm$ 0.003 & 0.367 $\pm$ 0.001 & \underline{0.100 $\pm$ 0.006} & 0.148 $\pm$ 0.002 & 0.088 $\pm$ 0.012 \\
 & BiGSL & \underline{0.273 $\pm$ 0.002} & \underline{0.326 $\pm$ 0.003} & \underline{0.281 $\pm$ 0.002} & \underline{0.355 $\pm$ 0.003} & \underline{0.285 $\pm$ 0.002} & \underline{0.375 $\pm$ 0.003} & 0.097 $\pm$ 0.005 & \underline{0.150 $\pm$ 0.003} & 0.084 $\pm$ 0.008 \\
 & GNPR-SID & 0.263 $\pm$ 0.006 & 0.313 $\pm$ 0.008 & 0.271 $\pm$ 0.006 & 0.342 $\pm$ 0.006 & 0.274 $\pm$ 0.005 & 0.359 $\pm$ 0.002 & 0.100 $\pm$ 0.008 & 0.148 $\pm$ 0.004 & \underline{0.090 $\pm$ 0.009} \\
 & CAGNN & 0.261 $\pm$ 0.003 & 0.314 $\pm$ 0.006 & 0.269 $\pm$ 0.003 & 0.343 $\pm$ 0.007 & 0.273 $\pm$ 0.004 & 0.362 $\pm$ 0.008 & 0.099 $\pm$ 0.009 & 0.143 $\pm$ 0.005 & 0.088 $\pm$ 0.012 \\
 & \textbf{EviRec} & \textbf{0.286 $\pm$ 0.003} & \textbf{0.341 $\pm$ 0.004} & \textbf{0.294 $\pm$ 0.002} & \textbf{0.370 $\pm$ 0.004} & \textbf{0.297 $\pm$ 0.003} & \textbf{0.388 $\pm$ 0.008} & \textbf{0.114 $\pm$ 0.008} & \textbf{0.155 $\pm$ 0.002} & \textbf{0.110 $\pm$ 0.010} \\
\bottomrule
\end{tabular}}
\end{table*}

\begin{table*}[p]
\centering
\tiny
\caption{Monthly performance in Houston.}
\label{tab:appendix-houston-part1}
\resizebox{\textwidth}{!}{%
\begin{tabular}{llccccccccc}
\toprule
Month & Method & All N@10 & All R@10 & All N@20 & All R@20 & All N@30 & All R@30 & POI-New N@10 & User-New N@10 & Dual-New N@10 \\
\midrule
2025-05 & SASRec & 0.057 $\pm$ 0.000 & 0.073 $\pm$ 0.000 & 0.059 $\pm$ 0.000 & 0.082 $\pm$ 0.000 & 0.061 $\pm$ 0.000 & 0.088 $\pm$ 0.000 & 0.016 $\pm$ 0.000 & 0.062 $\pm$ 0.000 & 0.027 $\pm$ 0.000 \\
 & GETNext & \underline{0.176 $\pm$ 0.000} & \underline{0.223 $\pm$ 0.000} & \underline{0.184 $\pm$ 0.000} & \underline{0.254 $\pm$ 0.000} & \underline{0.189 $\pm$ 0.000} & 0.276 $\pm$ 0.000 & 0.084 $\pm$ 0.000 & \underline{0.147 $\pm$ 0.000} & \underline{0.110 $\pm$ 0.000} \\
 & KBGNN & 0.173 $\pm$ 0.005 & 0.217 $\pm$ 0.008 & 0.181 $\pm$ 0.006 & 0.248 $\pm$ 0.009 & 0.185 $\pm$ 0.006 & 0.269 $\pm$ 0.010 & \underline{0.085 $\pm$ 0.002} & 0.143 $\pm$ 0.007 & 0.106 $\pm$ 0.008 \\
 & DisenPOI & 0.148 $\pm$ 0.005 & 0.195 $\pm$ 0.004 & 0.157 $\pm$ 0.004 & 0.230 $\pm$ 0.003 & 0.163 $\pm$ 0.004 & 0.257 $\pm$ 0.003 & 0.072 $\pm$ 0.000 & 0.129 $\pm$ 0.004 & 0.090 $\pm$ 0.002 \\
 & Diff-POI & 0.163 $\pm$ 0.001 & 0.210 $\pm$ 0.002 & 0.172 $\pm$ 0.001 & 0.243 $\pm$ 0.004 & 0.177 $\pm$ 0.002 & 0.267 $\pm$ 0.004 & 0.082 $\pm$ 0.002 & 0.141 $\pm$ 0.001 & 0.107 $\pm$ 0.003 \\
 & BiGSL & 0.168 $\pm$ 0.002 & 0.217 $\pm$ 0.002 & 0.177 $\pm$ 0.002 & 0.254 $\pm$ 0.003 & 0.183 $\pm$ 0.002 & \underline{0.280 $\pm$ 0.003} & 0.081 $\pm$ 0.004 & 0.141 $\pm$ 0.002 & 0.100 $\pm$ 0.005 \\
 & GNPR-SID & 0.171 $\pm$ 0.007 & 0.214 $\pm$ 0.009 & 0.179 $\pm$ 0.007 & 0.244 $\pm$ 0.010 & 0.183 $\pm$ 0.007 & 0.265 $\pm$ 0.011 & 0.085 $\pm$ 0.005 & 0.141 $\pm$ 0.008 & 0.105 $\pm$ 0.009 \\
 & CAGNN & 0.169 $\pm$ 0.005 & 0.213 $\pm$ 0.005 & 0.176 $\pm$ 0.005 & 0.243 $\pm$ 0.006 & 0.181 $\pm$ 0.005 & 0.264 $\pm$ 0.007 & 0.081 $\pm$ 0.003 & 0.137 $\pm$ 0.006 & 0.100 $\pm$ 0.003 \\
 & \textbf{EviRec} & \textbf{0.193 $\pm$ 0.008} & \textbf{0.242 $\pm$ 0.010} & \textbf{0.202 $\pm$ 0.008} & \textbf{0.277 $\pm$ 0.010} & \textbf{0.207 $\pm$ 0.008} & \textbf{0.300 $\pm$ 0.011} & \textbf{0.128 $\pm$ 0.008} & \textbf{0.162 $\pm$ 0.007} & \textbf{0.145 $\pm$ 0.008} \\
\midrule
2025-06 & SASRec & 0.052 $\pm$ 0.000 & 0.065 $\pm$ 0.000 & 0.054 $\pm$ 0.000 & 0.073 $\pm$ 0.000 & 0.055 $\pm$ 0.000 & 0.079 $\pm$ 0.000 & 0.016 $\pm$ 0.000 & 0.045 $\pm$ 0.000 & 0.036 $\pm$ 0.000 \\
 & GETNext & \underline{0.177 $\pm$ 0.000} & \underline{0.222 $\pm$ 0.000} & \underline{0.184 $\pm$ 0.000} & 0.252 $\pm$ 0.000 & \underline{0.189 $\pm$ 0.000} & 0.273 $\pm$ 0.000 & 0.102 $\pm$ 0.000 & 0.155 $\pm$ 0.000 & 0.152 $\pm$ 0.000 \\
 & KBGNN & 0.169 $\pm$ 0.006 & 0.212 $\pm$ 0.008 & 0.177 $\pm$ 0.007 & 0.242 $\pm$ 0.010 & 0.181 $\pm$ 0.007 & 0.263 $\pm$ 0.009 & 0.100 $\pm$ 0.001 & 0.151 $\pm$ 0.005 & 0.147 $\pm$ 0.006 \\
 & DisenPOI & 0.145 $\pm$ 0.006 & 0.185 $\pm$ 0.006 & 0.152 $\pm$ 0.006 & 0.216 $\pm$ 0.006 & 0.158 $\pm$ 0.006 & 0.240 $\pm$ 0.006 & 0.081 $\pm$ 0.003 & 0.122 $\pm$ 0.003 & 0.108 $\pm$ 0.003 \\
 & Diff-POI & 0.154 $\pm$ 0.001 & 0.196 $\pm$ 0.002 & 0.161 $\pm$ 0.001 & 0.228 $\pm$ 0.003 & 0.166 $\pm$ 0.001 & 0.250 $\pm$ 0.003 & 0.089 $\pm$ 0.001 & 0.137 $\pm$ 0.000 & 0.129 $\pm$ 0.004 \\
 & BiGSL & 0.175 $\pm$ 0.003 & 0.221 $\pm$ 0.003 & 0.184 $\pm$ 0.003 & \underline{0.254 $\pm$ 0.003} & 0.188 $\pm$ 0.003 & \underline{0.276 $\pm$ 0.003} & 0.105 $\pm$ 0.004 & 0.155 $\pm$ 0.002 & 0.147 $\pm$ 0.006 \\
 & GNPR-SID & 0.175 $\pm$ 0.006 & 0.219 $\pm$ 0.007 & 0.182 $\pm$ 0.005 & 0.249 $\pm$ 0.006 & 0.186 $\pm$ 0.005 & 0.269 $\pm$ 0.006 & \underline{0.107 $\pm$ 0.005} & \underline{0.157 $\pm$ 0.005} & \underline{0.154 $\pm$ 0.009} \\
 & CAGNN & 0.172 $\pm$ 0.002 & 0.215 $\pm$ 0.002 & 0.179 $\pm$ 0.002 & 0.245 $\pm$ 0.001 & 0.184 $\pm$ 0.002 & 0.266 $\pm$ 0.001 & 0.103 $\pm$ 0.003 & 0.152 $\pm$ 0.004 & 0.148 $\pm$ 0.005 \\
 & \textbf{EviRec} & \textbf{0.185 $\pm$ 0.003} & \textbf{0.233 $\pm$ 0.003} & \textbf{0.193 $\pm$ 0.003} & \textbf{0.264 $\pm$ 0.004} & \textbf{0.198 $\pm$ 0.003} & \textbf{0.286 $\pm$ 0.005} & \textbf{0.142 $\pm$ 0.005} & \textbf{0.167 $\pm$ 0.004} & \textbf{0.185 $\pm$ 0.006} \\
\midrule
2025-07 & SASRec & 0.052 $\pm$ 0.000 & 0.066 $\pm$ 0.000 & 0.054 $\pm$ 0.000 & 0.076 $\pm$ 0.000 & 0.055 $\pm$ 0.000 & 0.082 $\pm$ 0.000 & 0.010 $\pm$ 0.000 & 0.048 $\pm$ 0.000 & 0.024 $\pm$ 0.000 \\
 & GETNext & \underline{0.215 $\pm$ 0.000} & \underline{0.269 $\pm$ 0.000} & \underline{0.224 $\pm$ 0.000} & \underline{0.303 $\pm$ 0.000} & \underline{0.228 $\pm$ 0.000} & \underline{0.325 $\pm$ 0.000} & \underline{0.094 $\pm$ 0.000} & 0.168 $\pm$ 0.000 & \underline{0.165 $\pm$ 0.000} \\
 & KBGNN & 0.211 $\pm$ 0.004 & 0.263 $\pm$ 0.005 & 0.220 $\pm$ 0.004 & 0.296 $\pm$ 0.005 & 0.224 $\pm$ 0.004 & 0.318 $\pm$ 0.006 & 0.090 $\pm$ 0.007 & 0.168 $\pm$ 0.003 & 0.156 $\pm$ 0.011 \\
 & DisenPOI & 0.178 $\pm$ 0.005 & 0.224 $\pm$ 0.005 & 0.187 $\pm$ 0.005 & 0.258 $\pm$ 0.006 & 0.192 $\pm$ 0.005 & 0.284 $\pm$ 0.005 & 0.077 $\pm$ 0.004 & 0.142 $\pm$ 0.004 & 0.115 $\pm$ 0.008 \\
 & Diff-POI & 0.189 $\pm$ 0.001 & 0.238 $\pm$ 0.002 & 0.198 $\pm$ 0.001 & 0.272 $\pm$ 0.003 & 0.203 $\pm$ 0.001 & 0.296 $\pm$ 0.004 & 0.082 $\pm$ 0.006 & 0.154 $\pm$ 0.003 & 0.133 $\pm$ 0.003 \\
 & BiGSL & 0.209 $\pm$ 0.005 & 0.261 $\pm$ 0.005 & 0.218 $\pm$ 0.004 & 0.297 $\pm$ 0.004 & 0.223 $\pm$ 0.005 & 0.320 $\pm$ 0.006 & 0.091 $\pm$ 0.004 & 0.162 $\pm$ 0.006 & 0.147 $\pm$ 0.011 \\
 & GNPR-SID & 0.207 $\pm$ 0.002 & 0.258 $\pm$ 0.001 & 0.215 $\pm$ 0.002 & 0.289 $\pm$ 0.002 & 0.220 $\pm$ 0.002 & 0.311 $\pm$ 0.002 & 0.093 $\pm$ 0.005 & \underline{0.169 $\pm$ 0.005} & 0.159 $\pm$ 0.012 \\
 & CAGNN & 0.207 $\pm$ 0.005 & 0.259 $\pm$ 0.005 & 0.216 $\pm$ 0.005 & 0.292 $\pm$ 0.005 & 0.220 $\pm$ 0.005 & 0.315 $\pm$ 0.006 & 0.087 $\pm$ 0.008 & 0.163 $\pm$ 0.005 & 0.150 $\pm$ 0.017 \\
 & \textbf{EviRec} & \textbf{0.226 $\pm$ 0.003} & \textbf{0.282 $\pm$ 0.003} & \textbf{0.235 $\pm$ 0.003} & \textbf{0.315 $\pm$ 0.005} & \textbf{0.239 $\pm$ 0.003} & \textbf{0.336 $\pm$ 0.005} & \textbf{0.130 $\pm$ 0.003} & \textbf{0.183 $\pm$ 0.007} & \textbf{0.194 $\pm$ 0.009} \\
\midrule
2025-08 & SASRec & 0.073 $\pm$ 0.000 & 0.094 $\pm$ 0.000 & 0.076 $\pm$ 0.000 & 0.105 $\pm$ 0.000 & 0.078 $\pm$ 0.000 & 0.114 $\pm$ 0.000 & 0.021 $\pm$ 0.000 & 0.079 $\pm$ 0.000 & 0.051 $\pm$ 0.000 \\
 & GETNext & 0.220 $\pm$ 0.000 & 0.270 $\pm$ 0.000 & 0.229 $\pm$ 0.000 & 0.305 $\pm$ 0.000 & 0.233 $\pm$ 0.000 & 0.326 $\pm$ 0.000 & 0.104 $\pm$ 0.000 & 0.209 $\pm$ 0.000 & 0.131 $\pm$ 0.000 \\
 & KBGNN & 0.219 $\pm$ 0.006 & 0.270 $\pm$ 0.006 & 0.227 $\pm$ 0.005 & 0.302 $\pm$ 0.005 & 0.232 $\pm$ 0.006 & 0.323 $\pm$ 0.006 & 0.106 $\pm$ 0.001 & 0.213 $\pm$ 0.010 & 0.138 $\pm$ 0.008 \\
 & DisenPOI & 0.209 $\pm$ 0.006 & 0.256 $\pm$ 0.006 & 0.217 $\pm$ 0.005 & 0.288 $\pm$ 0.004 & 0.222 $\pm$ 0.005 & 0.310 $\pm$ 0.005 & 0.105 $\pm$ 0.005 & 0.198 $\pm$ 0.003 & 0.122 $\pm$ 0.008 \\
 & Diff-POI & 0.215 $\pm$ 0.003 & 0.266 $\pm$ 0.005 & 0.224 $\pm$ 0.003 & 0.298 $\pm$ 0.004 & 0.228 $\pm$ 0.003 & 0.320 $\pm$ 0.003 & 0.109 $\pm$ 0.005 & 0.213 $\pm$ 0.006 & \underline{0.145 $\pm$ 0.017} \\
 & BiGSL & \underline{0.226 $\pm$ 0.001} & \underline{0.279 $\pm$ 0.003} & \underline{0.235 $\pm$ 0.001} & \underline{0.312 $\pm$ 0.001} & \underline{0.239 $\pm$ 0.001} & \underline{0.334 $\pm$ 0.000} & \underline{0.113 $\pm$ 0.001} & \underline{0.219 $\pm$ 0.005} & 0.135 $\pm$ 0.004 \\
 & GNPR-SID & 0.221 $\pm$ 0.004 & 0.271 $\pm$ 0.004 & 0.229 $\pm$ 0.004 & 0.303 $\pm$ 0.004 & 0.233 $\pm$ 0.004 & 0.324 $\pm$ 0.004 & 0.107 $\pm$ 0.003 & 0.218 $\pm$ 0.006 & 0.138 $\pm$ 0.003 \\
 & CAGNN & 0.215 $\pm$ 0.007 & 0.265 $\pm$ 0.008 & 0.223 $\pm$ 0.007 & 0.296 $\pm$ 0.008 & 0.227 $\pm$ 0.007 & 0.317 $\pm$ 0.010 & 0.105 $\pm$ 0.004 & 0.215 $\pm$ 0.011 & 0.142 $\pm$ 0.019 \\
 & \textbf{EviRec} & \textbf{0.245 $\pm$ 0.003} & \textbf{0.305 $\pm$ 0.005} & \textbf{0.253 $\pm$ 0.003} & \textbf{0.337 $\pm$ 0.005} & \textbf{0.258 $\pm$ 0.003} & \textbf{0.360 $\pm$ 0.006} & \textbf{0.161 $\pm$ 0.010} & \textbf{0.242 $\pm$ 0.010} & \textbf{0.197 $\pm$ 0.018} \\
\midrule
2025-09 & SASRec & 0.064 $\pm$ 0.000 & 0.085 $\pm$ 0.000 & 0.068 $\pm$ 0.000 & 0.101 $\pm$ 0.000 & 0.070 $\pm$ 0.000 & 0.111 $\pm$ 0.000 & 0.026 $\pm$ 0.000 & 0.063 $\pm$ 0.000 & 0.045 $\pm$ 0.000 \\
 & GETNext & 0.287 $\pm$ 0.000 & 0.349 $\pm$ 0.000 & 0.296 $\pm$ 0.000 & 0.383 $\pm$ 0.000 & 0.300 $\pm$ 0.000 & 0.404 $\pm$ 0.000 & 0.221 $\pm$ 0.000 & 0.245 $\pm$ 0.000 & 0.223 $\pm$ 0.000 \\
 & KBGNN & 0.291 $\pm$ 0.004 & 0.351 $\pm$ 0.002 & 0.299 $\pm$ 0.004 & 0.384 $\pm$ 0.002 & 0.304 $\pm$ 0.004 & 0.405 $\pm$ 0.002 & 0.228 $\pm$ 0.006 & 0.245 $\pm$ 0.000 & 0.221 $\pm$ 0.004 \\
 & DisenPOI & \textbf{0.317 $\pm$ 0.004} & \underline{0.371 $\pm$ 0.003} & \textbf{0.325 $\pm$ 0.004} & \underline{0.401 $\pm$ 0.003} & \textbf{0.329 $\pm$ 0.004} & \underline{0.420 $\pm$ 0.003} & \underline{0.262 $\pm$ 0.005} & \underline{0.266 $\pm$ 0.006} & \underline{0.246 $\pm$ 0.008} \\
 & Diff-POI & 0.290 $\pm$ 0.002 & 0.350 $\pm$ 0.002 & 0.299 $\pm$ 0.002 & 0.384 $\pm$ 0.001 & 0.303 $\pm$ 0.002 & 0.405 $\pm$ 0.002 & 0.232 $\pm$ 0.006 & 0.243 $\pm$ 0.000 & 0.222 $\pm$ 0.002 \\
 & BiGSL & 0.302 $\pm$ 0.003 & 0.361 $\pm$ 0.003 & 0.310 $\pm$ 0.003 & 0.394 $\pm$ 0.002 & 0.315 $\pm$ 0.003 & 0.415 $\pm$ 0.001 & 0.244 $\pm$ 0.004 & 0.253 $\pm$ 0.004 & 0.233 $\pm$ 0.006 \\
 & GNPR-SID & 0.291 $\pm$ 0.003 & 0.350 $\pm$ 0.003 & 0.299 $\pm$ 0.003 & 0.383 $\pm$ 0.003 & 0.304 $\pm$ 0.003 & 0.403 $\pm$ 0.002 & 0.233 $\pm$ 0.003 & 0.245 $\pm$ 0.002 & 0.225 $\pm$ 0.001 \\
 & CAGNN & 0.282 $\pm$ 0.003 & 0.342 $\pm$ 0.002 & 0.291 $\pm$ 0.003 & 0.376 $\pm$ 0.002 & 0.295 $\pm$ 0.002 & 0.397 $\pm$ 0.002 & 0.221 $\pm$ 0.004 & 0.238 $\pm$ 0.002 & 0.215 $\pm$ 0.008 \\
 & \textbf{EviRec} & \underline{0.316 $\pm$ 0.003} & \textbf{0.380 $\pm$ 0.002} & \underline{0.324 $\pm$ 0.003} & \textbf{0.413 $\pm$ 0.001} & \underline{0.328 $\pm$ 0.003} & \textbf{0.433 $\pm$ 0.001} & \textbf{0.293 $\pm$ 0.005} & \textbf{0.267 $\pm$ 0.003} & \textbf{0.276 $\pm$ 0.001} \\
\midrule
2025-10 & SASRec & 0.028 $\pm$ 0.000 & 0.037 $\pm$ 0.000 & 0.029 $\pm$ 0.000 & 0.044 $\pm$ 0.000 & 0.030 $\pm$ 0.000 & 0.048 $\pm$ 0.000 & 0.007 $\pm$ 0.000 & 0.019 $\pm$ 0.000 & 0.016 $\pm$ 0.000 \\
 & GETNext & 0.162 $\pm$ 0.000 & 0.203 $\pm$ 0.000 & 0.169 $\pm$ 0.000 & 0.230 $\pm$ 0.000 & 0.173 $\pm$ 0.000 & 0.250 $\pm$ 0.000 & 0.116 $\pm$ 0.000 & 0.137 $\pm$ 0.000 & 0.191 $\pm$ 0.000 \\
 & KBGNN & 0.158 $\pm$ 0.007 & 0.197 $\pm$ 0.009 & 0.165 $\pm$ 0.007 & 0.224 $\pm$ 0.009 & 0.169 $\pm$ 0.007 & 0.242 $\pm$ 0.010 & 0.112 $\pm$ 0.007 & 0.132 $\pm$ 0.010 & 0.177 $\pm$ 0.016 \\
 & DisenPOI & 0.153 $\pm$ 0.003 & 0.190 $\pm$ 0.005 & 0.160 $\pm$ 0.003 & 0.217 $\pm$ 0.005 & 0.164 $\pm$ 0.004 & 0.237 $\pm$ 0.006 & 0.111 $\pm$ 0.003 & 0.125 $\pm$ 0.006 & 0.169 $\pm$ 0.011 \\
 & Diff-POI & 0.154 $\pm$ 0.001 & 0.193 $\pm$ 0.002 & 0.161 $\pm$ 0.001 & 0.220 $\pm$ 0.001 & 0.166 $\pm$ 0.001 & 0.240 $\pm$ 0.001 & 0.116 $\pm$ 0.004 & 0.132 $\pm$ 0.004 & 0.183 $\pm$ 0.011 \\
 & BiGSL & \underline{0.165 $\pm$ 0.004} & \underline{0.207 $\pm$ 0.004} & \underline{0.172 $\pm$ 0.004} & \underline{0.235 $\pm$ 0.004} & \underline{0.176 $\pm$ 0.004} & \underline{0.255 $\pm$ 0.004} & \underline{0.124 $\pm$ 0.006} & \underline{0.141 $\pm$ 0.004} & \underline{0.195 $\pm$ 0.012} \\
 & GNPR-SID & 0.157 $\pm$ 0.003 & 0.196 $\pm$ 0.003 & 0.164 $\pm$ 0.003 & 0.222 $\pm$ 0.003 & 0.168 $\pm$ 0.003 & 0.240 $\pm$ 0.003 & 0.115 $\pm$ 0.004 & 0.132 $\pm$ 0.006 & 0.179 $\pm$ 0.011 \\
 & CAGNN & 0.153 $\pm$ 0.001 & 0.192 $\pm$ 0.003 & 0.160 $\pm$ 0.001 & 0.218 $\pm$ 0.003 & 0.163 $\pm$ 0.001 & 0.236 $\pm$ 0.003 & 0.107 $\pm$ 0.001 & 0.128 $\pm$ 0.005 & 0.173 $\pm$ 0.008 \\
 & \textbf{EviRec} & \textbf{0.175 $\pm$ 0.001} & \textbf{0.219 $\pm$ 0.001} & \textbf{0.182 $\pm$ 0.001} & \textbf{0.246 $\pm$ 0.001} & \textbf{0.186 $\pm$ 0.002} & \textbf{0.265 $\pm$ 0.002} & \textbf{0.156 $\pm$ 0.002} & \textbf{0.152 $\pm$ 0.002} & \textbf{0.232 $\pm$ 0.003} \\
\midrule
2025-11 & SASRec & 0.020 $\pm$ 0.000 & 0.026 $\pm$ 0.000 & 0.021 $\pm$ 0.000 & 0.031 $\pm$ 0.000 & 0.021 $\pm$ 0.000 & 0.034 $\pm$ 0.000 & 0.004 $\pm$ 0.000 & 0.021 $\pm$ 0.000 & 0.006 $\pm$ 0.000 \\
 & GETNext & 0.139 $\pm$ 0.000 & 0.176 $\pm$ 0.000 & 0.145 $\pm$ 0.000 & 0.201 $\pm$ 0.000 & 0.149 $\pm$ 0.000 & 0.218 $\pm$ 0.000 & 0.097 $\pm$ 0.000 & 0.121 $\pm$ 0.000 & 0.129 $\pm$ 0.000 \\
 & KBGNN & 0.138 $\pm$ 0.001 & 0.174 $\pm$ 0.002 & 0.145 $\pm$ 0.001 & 0.199 $\pm$ 0.001 & 0.148 $\pm$ 0.001 & 0.216 $\pm$ 0.002 & 0.103 $\pm$ 0.005 & 0.124 $\pm$ 0.002 & 0.136 $\pm$ 0.008 \\
 & DisenPOI & 0.135 $\pm$ 0.004 & 0.170 $\pm$ 0.004 & 0.141 $\pm$ 0.004 & 0.196 $\pm$ 0.004 & 0.145 $\pm$ 0.004 & 0.215 $\pm$ 0.004 & 0.103 $\pm$ 0.005 & 0.117 $\pm$ 0.003 & 0.134 $\pm$ 0.007 \\
 & Diff-POI & 0.136 $\pm$ 0.001 & 0.172 $\pm$ 0.001 & 0.142 $\pm$ 0.001 & 0.197 $\pm$ 0.002 & 0.146 $\pm$ 0.001 & 0.214 $\pm$ 0.002 & 0.107 $\pm$ 0.005 & 0.123 $\pm$ 0.001 & 0.144 $\pm$ 0.006 \\
 & BiGSL & \underline{0.144 $\pm$ 0.002} & \underline{0.182 $\pm$ 0.002} & \underline{0.150 $\pm$ 0.002} & \underline{0.208 $\pm$ 0.002} & \underline{0.154 $\pm$ 0.002} & \underline{0.227 $\pm$ 0.002} & \underline{0.113 $\pm$ 0.003} & \underline{0.129 $\pm$ 0.002} & \underline{0.146 $\pm$ 0.004} \\
 & GNPR-SID & 0.139 $\pm$ 0.002 & 0.175 $\pm$ 0.003 & 0.145 $\pm$ 0.003 & 0.199 $\pm$ 0.003 & 0.148 $\pm$ 0.003 & 0.215 $\pm$ 0.004 & 0.108 $\pm$ 0.005 & 0.126 $\pm$ 0.002 & 0.142 $\pm$ 0.007 \\
 & CAGNN & 0.137 $\pm$ 0.006 & 0.174 $\pm$ 0.007 & 0.143 $\pm$ 0.006 & 0.198 $\pm$ 0.007 & 0.147 $\pm$ 0.006 & 0.215 $\pm$ 0.007 & 0.105 $\pm$ 0.009 & 0.124 $\pm$ 0.008 & 0.140 $\pm$ 0.019 \\
 & \textbf{EviRec} & \textbf{0.154 $\pm$ 0.003} & \textbf{0.193 $\pm$ 0.003} & \textbf{0.160 $\pm$ 0.003} & \textbf{0.219 $\pm$ 0.004} & \textbf{0.164 $\pm$ 0.003} & \textbf{0.237 $\pm$ 0.004} & \textbf{0.143 $\pm$ 0.007} & \textbf{0.139 $\pm$ 0.004} & \textbf{0.181 $\pm$ 0.006} \\
\midrule
2025-12 & SASRec & 0.041 $\pm$ 0.000 & 0.053 $\pm$ 0.000 & 0.043 $\pm$ 0.000 & 0.061 $\pm$ 0.000 & 0.044 $\pm$ 0.000 & 0.066 $\pm$ 0.000 & 0.006 $\pm$ 0.000 & 0.034 $\pm$ 0.000 & 0.012 $\pm$ 0.000 \\
 & GETNext & 0.195 $\pm$ 0.000 & 0.238 $\pm$ 0.000 & 0.203 $\pm$ 0.000 & 0.270 $\pm$ 0.000 & 0.207 $\pm$ 0.000 & 0.289 $\pm$ 0.000 & 0.047 $\pm$ 0.000 & 0.121 $\pm$ 0.000 & 0.090 $\pm$ 0.000 \\
 & KBGNN & 0.195 $\pm$ 0.005 & 0.238 $\pm$ 0.006 & 0.202 $\pm$ 0.005 & 0.266 $\pm$ 0.006 & 0.206 $\pm$ 0.005 & 0.286 $\pm$ 0.005 & 0.052 $\pm$ 0.007 & 0.124 $\pm$ 0.007 & 0.093 $\pm$ 0.009 \\
 & DisenPOI & 0.191 $\pm$ 0.005 & 0.234 $\pm$ 0.008 & 0.198 $\pm$ 0.005 & 0.261 $\pm$ 0.007 & 0.202 $\pm$ 0.005 & 0.279 $\pm$ 0.007 & 0.049 $\pm$ 0.004 & 0.115 $\pm$ 0.004 & 0.086 $\pm$ 0.006 \\
 & Diff-POI & 0.194 $\pm$ 0.002 & 0.239 $\pm$ 0.004 & 0.200 $\pm$ 0.002 & 0.265 $\pm$ 0.003 & 0.204 $\pm$ 0.002 & 0.285 $\pm$ 0.002 & 0.057 $\pm$ 0.002 & \underline{0.127 $\pm$ 0.004} & \underline{0.106 $\pm$ 0.003} \\
 & BiGSL & \underline{0.201 $\pm$ 0.004} & \underline{0.246 $\pm$ 0.005} & \underline{0.208 $\pm$ 0.004} & \underline{0.275 $\pm$ 0.005} & \underline{0.212 $\pm$ 0.003} & \underline{0.294 $\pm$ 0.003} & \underline{0.058 $\pm$ 0.002} & 0.126 $\pm$ 0.005 & 0.105 $\pm$ 0.008 \\
 & GNPR-SID & 0.193 $\pm$ 0.004 & 0.236 $\pm$ 0.004 & 0.200 $\pm$ 0.003 & 0.261 $\pm$ 0.004 & 0.204 $\pm$ 0.004 & 0.281 $\pm$ 0.004 & 0.054 $\pm$ 0.004 & 0.124 $\pm$ 0.006 & 0.097 $\pm$ 0.005 \\
 & CAGNN & 0.192 $\pm$ 0.005 & 0.234 $\pm$ 0.005 & 0.199 $\pm$ 0.006 & 0.260 $\pm$ 0.007 & 0.203 $\pm$ 0.006 & 0.280 $\pm$ 0.008 & 0.051 $\pm$ 0.006 & 0.121 $\pm$ 0.004 & 0.093 $\pm$ 0.009 \\
 & \textbf{EviRec} & \textbf{0.211 $\pm$ 0.004} & \textbf{0.259 $\pm$ 0.006} & \textbf{0.218 $\pm$ 0.004} & \textbf{0.286 $\pm$ 0.006} & \textbf{0.222 $\pm$ 0.004} & \textbf{0.305 $\pm$ 0.004} & \textbf{0.081 $\pm$ 0.003} & \textbf{0.137 $\pm$ 0.005} & \textbf{0.136 $\pm$ 0.004} \\
\bottomrule
\end{tabular}}
\end{table*}

\begin{table*}[p]
\centering
\tiny
\caption{Monthly performance in Los Angeles.}
\label{tab:appendix-los-angeles-part1}
\resizebox{\textwidth}{!}{%
\begin{tabular}{llccccccccc}
\toprule
Month & Method & All N@10 & All R@10 & All N@20 & All R@20 & All N@30 & All R@30 & POI-New N@10 & User-New N@10 & Dual-New N@10 \\
\midrule
2025-05 & SASRec & 0.104 $\pm$ 0.000 & 0.128 $\pm$ 0.000 & 0.107 $\pm$ 0.000 & 0.142 $\pm$ 0.000 & 0.109 $\pm$ 0.000 & 0.150 $\pm$ 0.000 & 0.029 $\pm$ 0.000 & 0.132 $\pm$ 0.000 & 0.056 $\pm$ 0.000 \\
 & GETNext & \underline{0.237 $\pm$ 0.000} & \underline{0.287 $\pm$ 0.000} & \underline{0.245 $\pm$ 0.000} & \underline{0.319 $\pm$ 0.000} & \underline{0.250 $\pm$ 0.000} & \underline{0.341 $\pm$ 0.000} & 0.099 $\pm$ 0.000 & \underline{0.254 $\pm$ 0.000} & \underline{0.150 $\pm$ 0.000} \\
 & KBGNN & 0.228 $\pm$ 0.010 & 0.277 $\pm$ 0.012 & 0.235 $\pm$ 0.010 & 0.307 $\pm$ 0.012 & 0.240 $\pm$ 0.011 & 0.329 $\pm$ 0.014 & 0.097 $\pm$ 0.006 & 0.240 $\pm$ 0.015 & 0.140 $\pm$ 0.011 \\
 & DisenPOI & 0.202 $\pm$ 0.008 & 0.255 $\pm$ 0.006 & 0.211 $\pm$ 0.008 & 0.290 $\pm$ 0.005 & 0.216 $\pm$ 0.007 & 0.314 $\pm$ 0.003 & 0.087 $\pm$ 0.006 & 0.223 $\pm$ 0.009 & 0.126 $\pm$ 0.008 \\
 & Diff-POI & 0.219 $\pm$ 0.004 & 0.269 $\pm$ 0.006 & 0.228 $\pm$ 0.004 & 0.304 $\pm$ 0.007 & 0.233 $\pm$ 0.004 & 0.328 $\pm$ 0.006 & 0.093 $\pm$ 0.006 & 0.243 $\pm$ 0.003 & 0.140 $\pm$ 0.007 \\
 & BiGSL & 0.221 $\pm$ 0.001 & 0.274 $\pm$ 0.000 & 0.230 $\pm$ 0.001 & 0.310 $\pm$ 0.001 & 0.236 $\pm$ 0.001 & 0.335 $\pm$ 0.000 & 0.094 $\pm$ 0.002 & 0.239 $\pm$ 0.003 & 0.134 $\pm$ 0.003 \\
 & GNPR-SID & 0.228 $\pm$ 0.009 & 0.277 $\pm$ 0.009 & 0.235 $\pm$ 0.009 & 0.308 $\pm$ 0.009 & 0.240 $\pm$ 0.009 & 0.330 $\pm$ 0.009 & \underline{0.100 $\pm$ 0.007} & 0.243 $\pm$ 0.011 & 0.144 $\pm$ 0.007 \\
 & CAGNN & 0.223 $\pm$ 0.001 & 0.273 $\pm$ 0.001 & 0.231 $\pm$ 0.001 & 0.307 $\pm$ 0.002 & 0.236 $\pm$ 0.001 & 0.328 $\pm$ 0.001 & 0.093 $\pm$ 0.001 & 0.235 $\pm$ 0.004 & 0.135 $\pm$ 0.002 \\
 & \textbf{EviRec} & \textbf{0.255 $\pm$ 0.006} & \textbf{0.306 $\pm$ 0.005} & \textbf{0.263 $\pm$ 0.006} & \textbf{0.340 $\pm$ 0.005} & \textbf{0.268 $\pm$ 0.006} & \textbf{0.362 $\pm$ 0.005} & \textbf{0.144 $\pm$ 0.005} & \textbf{0.274 $\pm$ 0.006} & \textbf{0.188 $\pm$ 0.008} \\
\midrule
2025-06 & SASRec & 0.109 $\pm$ 0.000 & 0.132 $\pm$ 0.000 & 0.113 $\pm$ 0.000 & 0.146 $\pm$ 0.000 & 0.114 $\pm$ 0.000 & 0.153 $\pm$ 0.000 & 0.027 $\pm$ 0.000 & 0.092 $\pm$ 0.000 & 0.045 $\pm$ 0.000 \\
 & GETNext & 0.243 $\pm$ 0.000 & 0.292 $\pm$ 0.000 & 0.250 $\pm$ 0.000 & 0.322 $\pm$ 0.000 & 0.255 $\pm$ 0.000 & 0.342 $\pm$ 0.000 & 0.103 $\pm$ 0.000 & 0.217 $\pm$ 0.000 & 0.143 $\pm$ 0.000 \\
 & KBGNN & 0.236 $\pm$ 0.007 & 0.284 $\pm$ 0.008 & 0.244 $\pm$ 0.007 & 0.313 $\pm$ 0.009 & 0.248 $\pm$ 0.007 & 0.334 $\pm$ 0.009 & 0.102 $\pm$ 0.001 & 0.210 $\pm$ 0.006 & 0.138 $\pm$ 0.005 \\
 & DisenPOI & 0.217 $\pm$ 0.007 & 0.265 $\pm$ 0.008 & 0.225 $\pm$ 0.007 & 0.297 $\pm$ 0.008 & 0.230 $\pm$ 0.007 & 0.319 $\pm$ 0.009 & 0.092 $\pm$ 0.005 & 0.184 $\pm$ 0.005 & 0.122 $\pm$ 0.005 \\
 & Diff-POI & 0.223 $\pm$ 0.003 & 0.271 $\pm$ 0.003 & 0.231 $\pm$ 0.003 & 0.304 $\pm$ 0.003 & 0.236 $\pm$ 0.003 & 0.325 $\pm$ 0.002 & 0.095 $\pm$ 0.005 & 0.190 $\pm$ 0.002 & 0.126 $\pm$ 0.002 \\
 & BiGSL & \underline{0.245 $\pm$ 0.003} & \underline{0.296 $\pm$ 0.004} & \underline{0.253 $\pm$ 0.003} & \underline{0.327 $\pm$ 0.004} & \underline{0.257 $\pm$ 0.003} & \underline{0.349 $\pm$ 0.004} & \underline{0.110 $\pm$ 0.004} & 0.217 $\pm$ 0.002 & 0.145 $\pm$ 0.003 \\
 & GNPR-SID & 0.241 $\pm$ 0.006 & 0.288 $\pm$ 0.005 & 0.249 $\pm$ 0.006 & 0.319 $\pm$ 0.005 & 0.253 $\pm$ 0.006 & 0.339 $\pm$ 0.004 & 0.107 $\pm$ 0.002 & \underline{0.217 $\pm$ 0.006} & \underline{0.147 $\pm$ 0.003} \\
 & CAGNN & 0.237 $\pm$ 0.005 & 0.286 $\pm$ 0.005 & 0.244 $\pm$ 0.005 & 0.316 $\pm$ 0.003 & 0.249 $\pm$ 0.005 & 0.337 $\pm$ 0.002 & 0.104 $\pm$ 0.003 & 0.213 $\pm$ 0.007 & 0.140 $\pm$ 0.003 \\
 & \textbf{EviRec} & \textbf{0.261 $\pm$ 0.002} & \textbf{0.313 $\pm$ 0.003} & \textbf{0.269 $\pm$ 0.002} & \textbf{0.345 $\pm$ 0.005} & \textbf{0.273 $\pm$ 0.002} & \textbf{0.366 $\pm$ 0.004} & \textbf{0.149 $\pm$ 0.006} & \textbf{0.233 $\pm$ 0.005} & \textbf{0.186 $\pm$ 0.001} \\
\midrule
2025-07 & SASRec & 0.102 $\pm$ 0.000 & 0.125 $\pm$ 0.000 & 0.105 $\pm$ 0.000 & 0.137 $\pm$ 0.000 & 0.106 $\pm$ 0.000 & 0.144 $\pm$ 0.000 & 0.011 $\pm$ 0.000 & 0.078 $\pm$ 0.000 & 0.027 $\pm$ 0.000 \\
 & GETNext & 0.275 $\pm$ 0.000 & 0.325 $\pm$ 0.000 & 0.284 $\pm$ 0.000 & 0.359 $\pm$ 0.000 & 0.289 $\pm$ 0.000 & 0.381 $\pm$ 0.000 & 0.098 $\pm$ 0.000 & 0.236 $\pm$ 0.000 & 0.171 $\pm$ 0.000 \\
 & KBGNN & 0.277 $\pm$ 0.005 & 0.327 $\pm$ 0.004 & 0.285 $\pm$ 0.004 & 0.358 $\pm$ 0.003 & 0.289 $\pm$ 0.004 & 0.378 $\pm$ 0.003 & 0.101 $\pm$ 0.003 & 0.238 $\pm$ 0.004 & 0.174 $\pm$ 0.008 \\
 & DisenPOI & 0.254 $\pm$ 0.003 & 0.306 $\pm$ 0.003 & 0.263 $\pm$ 0.003 & 0.339 $\pm$ 0.002 & 0.267 $\pm$ 0.003 & 0.360 $\pm$ 0.003 & 0.089 $\pm$ 0.003 & 0.213 $\pm$ 0.004 & 0.129 $\pm$ 0.006 \\
 & Diff-POI & 0.259 $\pm$ 0.003 & 0.312 $\pm$ 0.003 & 0.267 $\pm$ 0.002 & 0.345 $\pm$ 0.001 & 0.271 $\pm$ 0.002 & 0.366 $\pm$ 0.002 & 0.086 $\pm$ 0.002 & 0.222 $\pm$ 0.002 & 0.134 $\pm$ 0.004 \\
 & BiGSL & \underline{0.279 $\pm$ 0.007} & \underline{0.334 $\pm$ 0.006} & \underline{0.286 $\pm$ 0.007} & \underline{0.366 $\pm$ 0.005} & \underline{0.291 $\pm$ 0.007} & \underline{0.389 $\pm$ 0.006} & \underline{0.107 $\pm$ 0.007} & 0.240 $\pm$ 0.009 & 0.175 $\pm$ 0.013 \\
 & GNPR-SID & 0.275 $\pm$ 0.004 & 0.326 $\pm$ 0.005 & 0.283 $\pm$ 0.004 & 0.356 $\pm$ 0.005 & 0.287 $\pm$ 0.004 & 0.375 $\pm$ 0.005 & 0.105 $\pm$ 0.005 & \underline{0.240 $\pm$ 0.007} & \underline{0.180 $\pm$ 0.017} \\
 & CAGNN & 0.271 $\pm$ 0.007 & 0.323 $\pm$ 0.007 & 0.279 $\pm$ 0.007 & 0.355 $\pm$ 0.007 & 0.284 $\pm$ 0.007 & 0.377 $\pm$ 0.006 & 0.099 $\pm$ 0.004 & 0.235 $\pm$ 0.005 & 0.170 $\pm$ 0.011 \\
 & \textbf{EviRec} & \textbf{0.301 $\pm$ 0.003} & \textbf{0.357 $\pm$ 0.004} & \textbf{0.309 $\pm$ 0.003} & \textbf{0.387 $\pm$ 0.004} & \textbf{0.313 $\pm$ 0.003} & \textbf{0.407 $\pm$ 0.005} & \textbf{0.138 $\pm$ 0.002} & \textbf{0.257 $\pm$ 0.005} & \textbf{0.213 $\pm$ 0.008} \\
\midrule
2025-08 & SASRec & 0.157 $\pm$ 0.000 & 0.187 $\pm$ 0.000 & 0.161 $\pm$ 0.000 & 0.204 $\pm$ 0.000 & 0.162 $\pm$ 0.000 & 0.212 $\pm$ 0.000 & 0.018 $\pm$ 0.000 & 0.143 $\pm$ 0.000 & 0.029 $\pm$ 0.000 \\
 & GETNext & 0.344 $\pm$ 0.000 & 0.398 $\pm$ 0.000 & 0.351 $\pm$ 0.000 & 0.427 $\pm$ 0.000 & 0.355 $\pm$ 0.000 & 0.445 $\pm$ 0.000 & 0.154 $\pm$ 0.000 & 0.251 $\pm$ 0.000 & 0.083 $\pm$ 0.000 \\
 & KBGNN & 0.341 $\pm$ 0.007 & 0.391 $\pm$ 0.010 & 0.348 $\pm$ 0.007 & 0.421 $\pm$ 0.010 & 0.353 $\pm$ 0.006 & 0.441 $\pm$ 0.008 & 0.146 $\pm$ 0.007 & 0.252 $\pm$ 0.002 & 0.089 $\pm$ 0.001 \\
 & DisenPOI & 0.337 $\pm$ 0.003 & 0.385 $\pm$ 0.003 & 0.344 $\pm$ 0.003 & 0.413 $\pm$ 0.002 & 0.349 $\pm$ 0.003 & 0.433 $\pm$ 0.003 & 0.146 $\pm$ 0.007 & 0.241 $\pm$ 0.007 & 0.089 $\pm$ 0.011 \\
 & Diff-POI & 0.338 $\pm$ 0.002 & 0.389 $\pm$ 0.001 & 0.346 $\pm$ 0.001 & 0.419 $\pm$ 0.003 & 0.350 $\pm$ 0.001 & 0.436 $\pm$ 0.004 & 0.143 $\pm$ 0.002 & \underline{0.256 $\pm$ 0.004} & 0.095 $\pm$ 0.005 \\
 & BiGSL & \underline{0.352 $\pm$ 0.005} & \underline{0.405 $\pm$ 0.008} & \underline{0.360 $\pm$ 0.005} & \underline{0.434 $\pm$ 0.005} & \underline{0.364 $\pm$ 0.005} & \underline{0.454 $\pm$ 0.007} & \underline{0.159 $\pm$ 0.005} & 0.253 $\pm$ 0.004 & \underline{0.099 $\pm$ 0.007} \\
 & GNPR-SID & 0.339 $\pm$ 0.003 & 0.388 $\pm$ 0.005 & 0.346 $\pm$ 0.003 & 0.418 $\pm$ 0.004 & 0.350 $\pm$ 0.003 & 0.438 $\pm$ 0.005 & 0.146 $\pm$ 0.004 & 0.253 $\pm$ 0.003 & 0.097 $\pm$ 0.009 \\
 & CAGNN & 0.332 $\pm$ 0.007 & 0.383 $\pm$ 0.006 & 0.339 $\pm$ 0.007 & 0.412 $\pm$ 0.007 & 0.344 $\pm$ 0.007 & 0.433 $\pm$ 0.007 & 0.143 $\pm$ 0.006 & 0.247 $\pm$ 0.006 & 0.094 $\pm$ 0.003 \\
 & \textbf{EviRec} & \textbf{0.373 $\pm$ 0.002} & \textbf{0.429 $\pm$ 0.004} & \textbf{0.381 $\pm$ 0.001} & \textbf{0.459 $\pm$ 0.003} & \textbf{0.384 $\pm$ 0.001} & \textbf{0.476 $\pm$ 0.003} & \textbf{0.193 $\pm$ 0.006} & \textbf{0.269 $\pm$ 0.001} & \textbf{0.122 $\pm$ 0.006} \\
\midrule
2025-09 & SASRec & 0.122 $\pm$ 0.000 & 0.156 $\pm$ 0.000 & 0.128 $\pm$ 0.000 & 0.177 $\pm$ 0.000 & 0.130 $\pm$ 0.000 & 0.190 $\pm$ 0.000 & 0.034 $\pm$ 0.000 & 0.119 $\pm$ 0.000 & 0.049 $\pm$ 0.000 \\
 & GETNext & 0.355 $\pm$ 0.000 & 0.421 $\pm$ 0.000 & 0.364 $\pm$ 0.000 & 0.455 $\pm$ 0.000 & 0.369 $\pm$ 0.000 & 0.477 $\pm$ 0.000 & 0.240 $\pm$ 0.000 & 0.296 $\pm$ 0.000 & 0.241 $\pm$ 0.000 \\
 & KBGNN & 0.359 $\pm$ 0.007 & 0.423 $\pm$ 0.006 & 0.368 $\pm$ 0.007 & 0.458 $\pm$ 0.005 & 0.373 $\pm$ 0.006 & 0.480 $\pm$ 0.004 & 0.240 $\pm$ 0.007 & 0.296 $\pm$ 0.008 & 0.234 $\pm$ 0.011 \\
 & DisenPOI & \textbf{0.402 $\pm$ 0.004} & \underline{0.463 $\pm$ 0.004} & \textbf{0.409 $\pm$ 0.004} & \underline{0.493 $\pm$ 0.004} & \textbf{0.413 $\pm$ 0.004} & \underline{0.512 $\pm$ 0.004} & \underline{0.306 $\pm$ 0.008} & \underline{0.331 $\pm$ 0.004} & \underline{0.276 $\pm$ 0.008} \\
 & Diff-POI & 0.365 $\pm$ 0.004 & 0.430 $\pm$ 0.003 & 0.373 $\pm$ 0.003 & 0.465 $\pm$ 0.002 & 0.378 $\pm$ 0.003 & 0.486 $\pm$ 0.002 & 0.249 $\pm$ 0.004 & 0.306 $\pm$ 0.003 & 0.242 $\pm$ 0.003 \\
 & BiGSL & 0.380 $\pm$ 0.004 & 0.443 $\pm$ 0.005 & 0.389 $\pm$ 0.004 & 0.477 $\pm$ 0.005 & 0.393 $\pm$ 0.004 & 0.497 $\pm$ 0.005 & 0.270 $\pm$ 0.005 & 0.318 $\pm$ 0.007 & 0.255 $\pm$ 0.008 \\
 & GNPR-SID & 0.360 $\pm$ 0.005 & 0.423 $\pm$ 0.004 & 0.368 $\pm$ 0.005 & 0.456 $\pm$ 0.004 & 0.373 $\pm$ 0.005 & 0.477 $\pm$ 0.003 & 0.243 $\pm$ 0.007 & 0.300 $\pm$ 0.007 & 0.237 $\pm$ 0.014 \\
 & CAGNN & 0.347 $\pm$ 0.012 & 0.411 $\pm$ 0.011 & 0.356 $\pm$ 0.012 & 0.445 $\pm$ 0.011 & 0.361 $\pm$ 0.012 & 0.468 $\pm$ 0.011 & 0.231 $\pm$ 0.013 & 0.289 $\pm$ 0.012 & 0.224 $\pm$ 0.017 \\
 & \textbf{EviRec} & \underline{0.396 $\pm$ 0.005} & \textbf{0.464 $\pm$ 0.006} & \underline{0.404 $\pm$ 0.005} & \textbf{0.497 $\pm$ 0.007} & \underline{0.408 $\pm$ 0.005} & \textbf{0.516 $\pm$ 0.006} & \textbf{0.319 $\pm$ 0.012} & \textbf{0.331 $\pm$ 0.002} & \textbf{0.298 $\pm$ 0.006} \\
\midrule
2025-10 & SASRec & 0.057 $\pm$ 0.000 & 0.070 $\pm$ 0.000 & 0.059 $\pm$ 0.000 & 0.078 $\pm$ 0.000 & 0.060 $\pm$ 0.000 & 0.084 $\pm$ 0.000 & 0.007 $\pm$ 0.000 & 0.054 $\pm$ 0.000 & 0.013 $\pm$ 0.000 \\
 & GETNext & 0.202 $\pm$ 0.000 & 0.244 $\pm$ 0.000 & 0.208 $\pm$ 0.000 & 0.270 $\pm$ 0.000 & 0.212 $\pm$ 0.000 & 0.286 $\pm$ 0.000 & 0.083 $\pm$ 0.000 & 0.174 $\pm$ 0.000 & 0.098 $\pm$ 0.000 \\
 & KBGNN & 0.198 $\pm$ 0.007 & 0.237 $\pm$ 0.009 & 0.204 $\pm$ 0.007 & 0.262 $\pm$ 0.010 & 0.208 $\pm$ 0.007 & 0.279 $\pm$ 0.009 & 0.081 $\pm$ 0.005 & 0.167 $\pm$ 0.009 & 0.092 $\pm$ 0.011 \\
 & DisenPOI & 0.202 $\pm$ 0.003 & 0.241 $\pm$ 0.005 & 0.209 $\pm$ 0.003 & 0.268 $\pm$ 0.004 & 0.213 $\pm$ 0.003 & 0.285 $\pm$ 0.005 & 0.089 $\pm$ 0.004 & 0.169 $\pm$ 0.002 & 0.091 $\pm$ 0.000 \\
 & Diff-POI & 0.199 $\pm$ 0.002 & 0.237 $\pm$ 0.001 & 0.205 $\pm$ 0.002 & 0.263 $\pm$ 0.003 & 0.209 $\pm$ 0.002 & 0.280 $\pm$ 0.003 & 0.083 $\pm$ 0.001 & 0.171 $\pm$ 0.002 & 0.093 $\pm$ 0.003 \\
 & BiGSL & \underline{0.209 $\pm$ 0.003} & \underline{0.253 $\pm$ 0.004} & \underline{0.216 $\pm$ 0.003} & \underline{0.279 $\pm$ 0.004} & \underline{0.220 $\pm$ 0.003} & \underline{0.297 $\pm$ 0.004} & \underline{0.091 $\pm$ 0.003} & \underline{0.176 $\pm$ 0.003} & \underline{0.100 $\pm$ 0.007} \\
 & GNPR-SID & 0.196 $\pm$ 0.003 & 0.235 $\pm$ 0.004 & 0.203 $\pm$ 0.003 & 0.260 $\pm$ 0.004 & 0.206 $\pm$ 0.002 & 0.277 $\pm$ 0.003 & 0.081 $\pm$ 0.003 & 0.166 $\pm$ 0.005 & 0.093 $\pm$ 0.005 \\
 & CAGNN & 0.193 $\pm$ 0.001 & 0.232 $\pm$ 0.002 & 0.200 $\pm$ 0.001 & 0.259 $\pm$ 0.002 & 0.204 $\pm$ 0.001 & 0.276 $\pm$ 0.003 & 0.077 $\pm$ 0.001 & 0.163 $\pm$ 0.002 & 0.087 $\pm$ 0.005 \\
 & \textbf{EviRec} & \textbf{0.217 $\pm$ 0.005} & \textbf{0.261 $\pm$ 0.006} & \textbf{0.223 $\pm$ 0.005} & \textbf{0.287 $\pm$ 0.005} & \textbf{0.227 $\pm$ 0.005} & \textbf{0.304 $\pm$ 0.003} & \textbf{0.122 $\pm$ 0.002} & \textbf{0.185 $\pm$ 0.002} & \textbf{0.132 $\pm$ 0.003} \\
\midrule
2025-11 & SASRec & 0.037 $\pm$ 0.000 & 0.049 $\pm$ 0.000 & 0.039 $\pm$ 0.000 & 0.056 $\pm$ 0.000 & 0.040 $\pm$ 0.000 & 0.061 $\pm$ 0.000 & 0.022 $\pm$ 0.000 & 0.026 $\pm$ 0.000 & 0.021 $\pm$ 0.000 \\
 & GETNext & 0.182 $\pm$ 0.000 & 0.219 $\pm$ 0.000 & 0.188 $\pm$ 0.000 & 0.242 $\pm$ 0.000 & 0.192 $\pm$ 0.000 & 0.259 $\pm$ 0.000 & 0.107 $\pm$ 0.000 & 0.134 $\pm$ 0.000 & \underline{0.125 $\pm$ 0.000} \\
 & KBGNN & 0.180 $\pm$ 0.002 & 0.218 $\pm$ 0.003 & 0.186 $\pm$ 0.002 & 0.240 $\pm$ 0.004 & 0.189 $\pm$ 0.002 & 0.254 $\pm$ 0.004 & 0.105 $\pm$ 0.010 & 0.131 $\pm$ 0.006 & 0.118 $\pm$ 0.014 \\
 & DisenPOI & 0.188 $\pm$ 0.003 & 0.225 $\pm$ 0.003 & 0.193 $\pm$ 0.003 & 0.248 $\pm$ 0.003 & 0.197 $\pm$ 0.003 & 0.263 $\pm$ 0.004 & \underline{0.110 $\pm$ 0.010} & 0.131 $\pm$ 0.005 & 0.116 $\pm$ 0.010 \\
 & Diff-POI & 0.181 $\pm$ 0.003 & 0.219 $\pm$ 0.003 & 0.187 $\pm$ 0.003 & 0.243 $\pm$ 0.003 & 0.191 $\pm$ 0.003 & 0.259 $\pm$ 0.002 & 0.108 $\pm$ 0.008 & 0.135 $\pm$ 0.005 & 0.123 $\pm$ 0.012 \\
 & BiGSL & \underline{0.189 $\pm$ 0.005} & \underline{0.228 $\pm$ 0.004} & \underline{0.195 $\pm$ 0.005} & \underline{0.252 $\pm$ 0.004} & \underline{0.198 $\pm$ 0.005} & \underline{0.269 $\pm$ 0.005} & 0.109 $\pm$ 0.006 & \underline{0.137 $\pm$ 0.004} & 0.121 $\pm$ 0.009 \\
 & GNPR-SID & 0.180 $\pm$ 0.002 & 0.216 $\pm$ 0.003 & 0.186 $\pm$ 0.002 & 0.238 $\pm$ 0.003 & 0.189 $\pm$ 0.002 & 0.254 $\pm$ 0.003 & 0.106 $\pm$ 0.009 & 0.131 $\pm$ 0.006 & 0.118 $\pm$ 0.013 \\
 & CAGNN & 0.178 $\pm$ 0.003 & 0.215 $\pm$ 0.004 & 0.184 $\pm$ 0.003 & 0.238 $\pm$ 0.005 & 0.187 $\pm$ 0.003 & 0.253 $\pm$ 0.005 & 0.102 $\pm$ 0.005 & 0.131 $\pm$ 0.003 & 0.120 $\pm$ 0.009 \\
 & \textbf{EviRec} & \textbf{0.201 $\pm$ 0.005} & \textbf{0.241 $\pm$ 0.006} & \textbf{0.208 $\pm$ 0.005} & \textbf{0.267 $\pm$ 0.005} & \textbf{0.211 $\pm$ 0.005} & \textbf{0.282 $\pm$ 0.006} & \textbf{0.140 $\pm$ 0.009} & \textbf{0.149 $\pm$ 0.007} & \textbf{0.152 $\pm$ 0.012} \\
\midrule
2025-12 & SASRec & 0.085 $\pm$ 0.000 & 0.106 $\pm$ 0.000 & 0.088 $\pm$ 0.000 & 0.117 $\pm$ 0.000 & 0.089 $\pm$ 0.000 & 0.123 $\pm$ 0.000 & 0.026 $\pm$ 0.000 & 0.069 $\pm$ 0.000 & 0.055 $\pm$ 0.000 \\
 & GETNext & 0.274 $\pm$ 0.000 & 0.323 $\pm$ 0.000 & 0.281 $\pm$ 0.000 & \underline{0.351 $\pm$ 0.000} & 0.284 $\pm$ 0.000 & 0.367 $\pm$ 0.000 & 0.123 $\pm$ 0.000 & 0.188 $\pm$ 0.000 & \underline{0.236 $\pm$ 0.000} \\
 & KBGNN & 0.268 $\pm$ 0.005 & 0.312 $\pm$ 0.009 & 0.274 $\pm$ 0.005 & 0.338 $\pm$ 0.009 & 0.278 $\pm$ 0.005 & 0.357 $\pm$ 0.008 & 0.125 $\pm$ 0.009 & 0.178 $\pm$ 0.005 & 0.221 $\pm$ 0.007 \\
 & DisenPOI & \underline{0.280 $\pm$ 0.003} & 0.324 $\pm$ 0.003 & \underline{0.286 $\pm$ 0.003} & 0.348 $\pm$ 0.002 & \underline{0.290 $\pm$ 0.002} & 0.366 $\pm$ 0.001 & 0.117 $\pm$ 0.003 & 0.178 $\pm$ 0.005 & 0.209 $\pm$ 0.009 \\
 & Diff-POI & 0.273 $\pm$ 0.003 & 0.320 $\pm$ 0.005 & 0.279 $\pm$ 0.003 & 0.343 $\pm$ 0.008 & 0.282 $\pm$ 0.003 & 0.360 $\pm$ 0.005 & 0.128 $\pm$ 0.003 & \underline{0.188 $\pm$ 0.006} & 0.228 $\pm$ 0.006 \\
 & BiGSL & 0.277 $\pm$ 0.001 & \underline{0.325 $\pm$ 0.003} & 0.284 $\pm$ 0.000 & 0.350 $\pm$ 0.005 & 0.288 $\pm$ 0.001 & \underline{0.369 $\pm$ 0.006} & 0.125 $\pm$ 0.004 & 0.185 $\pm$ 0.006 & 0.215 $\pm$ 0.012 \\
 & GNPR-SID & 0.268 $\pm$ 0.004 & 0.310 $\pm$ 0.004 & 0.275 $\pm$ 0.004 & 0.338 $\pm$ 0.005 & 0.278 $\pm$ 0.004 & 0.355 $\pm$ 0.007 & \underline{0.137 $\pm$ 0.010} & 0.181 $\pm$ 0.001 & 0.231 $\pm$ 0.005 \\
 & CAGNN & 0.262 $\pm$ 0.005 & 0.309 $\pm$ 0.002 & 0.269 $\pm$ 0.005 & 0.336 $\pm$ 0.005 & 0.273 $\pm$ 0.005 & 0.354 $\pm$ 0.005 & 0.123 $\pm$ 0.009 & 0.181 $\pm$ 0.005 & 0.219 $\pm$ 0.015 \\
 & \textbf{EviRec} & \textbf{0.290 $\pm$ 0.008} & \textbf{0.337 $\pm$ 0.008} & \textbf{0.297 $\pm$ 0.008} & \textbf{0.363 $\pm$ 0.009} & \textbf{0.301 $\pm$ 0.008} & \textbf{0.382 $\pm$ 0.007} & \textbf{0.159 $\pm$ 0.007} & \textbf{0.198 $\pm$ 0.006} & \textbf{0.258 $\pm$ 0.005} \\
\bottomrule
\end{tabular}}
\end{table*}

\begin{table*}[p]
\centering
\tiny
\caption{Monthly performance in New York.}
\label{tab:appendix-new-york-part1}
\resizebox{\textwidth}{!}{%
\begin{tabular}{llccccccccc}
\toprule
Month & Method & All N@10 & All R@10 & All N@20 & All R@20 & All N@30 & All R@30 & POI-New N@10 & User-New N@10 & Dual-New N@10 \\
\midrule
2025-05 & SASRec & 0.076 $\pm$ 0.000 & 0.093 $\pm$ 0.000 & 0.078 $\pm$ 0.000 & 0.103 $\pm$ 0.000 & 0.080 $\pm$ 0.000 & 0.110 $\pm$ 0.000 & 0.031 $\pm$ 0.000 & 0.086 $\pm$ 0.000 & 0.054 $\pm$ 0.000 \\
 & GETNext & 0.137 $\pm$ 0.000 & 0.165 $\pm$ 0.000 & 0.143 $\pm$ 0.000 & 0.186 $\pm$ 0.000 & 0.145 $\pm$ 0.000 & 0.199 $\pm$ 0.000 & 0.061 $\pm$ 0.000 & 0.132 $\pm$ 0.000 & 0.090 $\pm$ 0.000 \\
 & KBGNN & 0.134 $\pm$ 0.003 & 0.163 $\pm$ 0.003 & 0.139 $\pm$ 0.003 & 0.183 $\pm$ 0.003 & 0.142 $\pm$ 0.003 & 0.199 $\pm$ 0.001 & 0.061 $\pm$ 0.003 & 0.128 $\pm$ 0.008 & 0.085 $\pm$ 0.005 \\
 & DisenPOI & 0.131 $\pm$ 0.003 & 0.166 $\pm$ 0.003 & 0.137 $\pm$ 0.003 & 0.189 $\pm$ 0.003 & 0.140 $\pm$ 0.003 & 0.205 $\pm$ 0.004 & 0.057 $\pm$ 0.003 & 0.123 $\pm$ 0.005 & 0.076 $\pm$ 0.005 \\
 & Diff-POI & 0.144 $\pm$ 0.004 & 0.175 $\pm$ 0.004 & 0.149 $\pm$ 0.004 & 0.195 $\pm$ 0.004 & 0.152 $\pm$ 0.004 & 0.210 $\pm$ 0.004 & \underline{0.068 $\pm$ 0.003} & \underline{0.139 $\pm$ 0.005} & \underline{0.093 $\pm$ 0.004} \\
 & BiGSL & \underline{0.145 $\pm$ 0.001} & \underline{0.178 $\pm$ 0.002} & \underline{0.150 $\pm$ 0.001} & \underline{0.200 $\pm$ 0.002} & \underline{0.154 $\pm$ 0.001} & \underline{0.216 $\pm$ 0.003} & 0.065 $\pm$ 0.004 & 0.136 $\pm$ 0.000 & 0.089 $\pm$ 0.003 \\
 & GNPR-SID & 0.135 $\pm$ 0.005 & 0.165 $\pm$ 0.005 & 0.140 $\pm$ 0.005 & 0.185 $\pm$ 0.004 & 0.143 $\pm$ 0.005 & 0.201 $\pm$ 0.004 & 0.063 $\pm$ 0.001 & 0.129 $\pm$ 0.009 & 0.086 $\pm$ 0.004 \\
 & CAGNN & 0.131 $\pm$ 0.001 & 0.161 $\pm$ 0.003 & 0.137 $\pm$ 0.002 & 0.183 $\pm$ 0.005 & 0.140 $\pm$ 0.002 & 0.197 $\pm$ 0.005 & 0.060 $\pm$ 0.002 & 0.124 $\pm$ 0.006 & 0.082 $\pm$ 0.004 \\
 & \textbf{EviRec} & \textbf{0.160 $\pm$ 0.008} & \textbf{0.191 $\pm$ 0.008} & \textbf{0.165 $\pm$ 0.008} & \textbf{0.213 $\pm$ 0.010} & \textbf{0.169 $\pm$ 0.007} & \textbf{0.229 $\pm$ 0.009} & \textbf{0.098 $\pm$ 0.006} & \textbf{0.152 $\pm$ 0.009} & \textbf{0.119 $\pm$ 0.010} \\
\midrule
2025-06 & SASRec & 0.053 $\pm$ 0.000 & 0.066 $\pm$ 0.000 & 0.054 $\pm$ 0.000 & 0.072 $\pm$ 0.000 & 0.055 $\pm$ 0.000 & 0.076 $\pm$ 0.000 & 0.020 $\pm$ 0.000 & 0.034 $\pm$ 0.000 & 0.022 $\pm$ 0.000 \\
 & GETNext & 0.121 $\pm$ 0.000 & 0.151 $\pm$ 0.000 & 0.126 $\pm$ 0.000 & 0.172 $\pm$ 0.000 & 0.130 $\pm$ 0.000 & 0.188 $\pm$ 0.000 & 0.065 $\pm$ 0.000 & 0.099 $\pm$ 0.000 & 0.080 $\pm$ 0.000 \\
 & KBGNN & 0.114 $\pm$ 0.002 & 0.142 $\pm$ 0.003 & 0.119 $\pm$ 0.003 & 0.162 $\pm$ 0.006 & 0.123 $\pm$ 0.003 & 0.177 $\pm$ 0.007 & 0.059 $\pm$ 0.005 & 0.091 $\pm$ 0.004 & 0.074 $\pm$ 0.008 \\
 & DisenPOI & 0.106 $\pm$ 0.005 & 0.132 $\pm$ 0.004 & 0.111 $\pm$ 0.005 & 0.153 $\pm$ 0.003 & 0.114 $\pm$ 0.005 & 0.168 $\pm$ 0.003 & 0.049 $\pm$ 0.004 & 0.077 $\pm$ 0.004 & 0.052 $\pm$ 0.002 \\
 & Diff-POI & 0.115 $\pm$ 0.003 & 0.142 $\pm$ 0.004 & 0.120 $\pm$ 0.003 & 0.164 $\pm$ 0.005 & 0.123 $\pm$ 0.003 & 0.177 $\pm$ 0.006 & 0.058 $\pm$ 0.004 & 0.090 $\pm$ 0.004 & 0.066 $\pm$ 0.005 \\
 & BiGSL & \underline{0.130 $\pm$ 0.005} & \underline{0.160 $\pm$ 0.008} & \underline{0.135 $\pm$ 0.005} & \underline{0.182 $\pm$ 0.007} & \underline{0.139 $\pm$ 0.005} & \underline{0.197 $\pm$ 0.006} & \underline{0.073 $\pm$ 0.008} & \underline{0.109 $\pm$ 0.007} & \underline{0.091 $\pm$ 0.009} \\
 & GNPR-SID & 0.118 $\pm$ 0.007 & 0.146 $\pm$ 0.008 & 0.123 $\pm$ 0.007 & 0.167 $\pm$ 0.006 & 0.126 $\pm$ 0.007 & 0.182 $\pm$ 0.007 & 0.064 $\pm$ 0.006 & 0.098 $\pm$ 0.008 & 0.081 $\pm$ 0.009 \\
 & CAGNN & 0.115 $\pm$ 0.008 & 0.142 $\pm$ 0.009 & 0.120 $\pm$ 0.008 & 0.163 $\pm$ 0.010 & 0.123 $\pm$ 0.008 & 0.177 $\pm$ 0.010 & 0.060 $\pm$ 0.005 & 0.093 $\pm$ 0.010 & 0.076 $\pm$ 0.010 \\
 & \textbf{EviRec} & \textbf{0.138 $\pm$ 0.008} & \textbf{0.169 $\pm$ 0.006} & \textbf{0.144 $\pm$ 0.007} & \textbf{0.191 $\pm$ 0.005} & \textbf{0.147 $\pm$ 0.007} & \textbf{0.207 $\pm$ 0.002} & \textbf{0.099 $\pm$ 0.010} & \textbf{0.119 $\pm$ 0.004} & \textbf{0.118 $\pm$ 0.012} \\
\midrule
2025-07 & SASRec & 0.033 $\pm$ 0.000 & 0.041 $\pm$ 0.000 & 0.035 $\pm$ 0.000 & 0.047 $\pm$ 0.000 & 0.036 $\pm$ 0.000 & 0.051 $\pm$ 0.000 & 0.005 $\pm$ 0.000 & 0.040 $\pm$ 0.000 & 0.019 $\pm$ 0.000 \\
 & GETNext & 0.108 $\pm$ 0.000 & 0.143 $\pm$ 0.000 & 0.113 $\pm$ 0.000 & 0.164 $\pm$ 0.000 & 0.117 $\pm$ 0.000 & 0.185 $\pm$ 0.000 & 0.099 $\pm$ 0.000 & 0.112 $\pm$ 0.000 & 0.083 $\pm$ 0.000 \\
 & KBGNN & 0.111 $\pm$ 0.004 & 0.143 $\pm$ 0.003 & 0.118 $\pm$ 0.004 & 0.169 $\pm$ 0.003 & 0.122 $\pm$ 0.004 & 0.186 $\pm$ 0.003 & 0.098 $\pm$ 0.012 & 0.111 $\pm$ 0.004 & 0.084 $\pm$ 0.001 \\
 & DisenPOI & 0.095 $\pm$ 0.003 & 0.119 $\pm$ 0.003 & 0.101 $\pm$ 0.002 & 0.143 $\pm$ 0.002 & 0.105 $\pm$ 0.003 & 0.160 $\pm$ 0.004 & 0.053 $\pm$ 0.002 & 0.107 $\pm$ 0.009 & 0.083 $\pm$ 0.007 \\
 & Diff-POI & 0.110 $\pm$ 0.004 & 0.142 $\pm$ 0.003 & 0.117 $\pm$ 0.004 & 0.167 $\pm$ 0.005 & 0.120 $\pm$ 0.004 & 0.182 $\pm$ 0.005 & 0.088 $\pm$ 0.026 & \underline{0.118 $\pm$ 0.000} & 0.086 $\pm$ 0.011 \\
 & BiGSL & \underline{0.120 $\pm$ 0.003} & \underline{0.153 $\pm$ 0.003} & \underline{0.126 $\pm$ 0.003} & \underline{0.177 $\pm$ 0.005} & \underline{0.130 $\pm$ 0.003} & \underline{0.196 $\pm$ 0.005} & 0.097 $\pm$ 0.003 & 0.118 $\pm$ 0.005 & 0.084 $\pm$ 0.021 \\
 & GNPR-SID & 0.111 $\pm$ 0.002 & 0.144 $\pm$ 0.003 & 0.117 $\pm$ 0.002 & 0.167 $\pm$ 0.004 & 0.120 $\pm$ 0.002 & 0.183 $\pm$ 0.006 & \underline{0.101 $\pm$ 0.022} & 0.115 $\pm$ 0.004 & \underline{0.090 $\pm$ 0.010} \\
 & CAGNN & 0.102 $\pm$ 0.006 & 0.134 $\pm$ 0.008 & 0.108 $\pm$ 0.006 & 0.158 $\pm$ 0.008 & 0.111 $\pm$ 0.006 & 0.175 $\pm$ 0.010 & 0.076 $\pm$ 0.009 & 0.111 $\pm$ 0.004 & 0.081 $\pm$ 0.011 \\
 & \textbf{EviRec} & \textbf{0.144 $\pm$ 0.006} & \textbf{0.182 $\pm$ 0.005} & \textbf{0.151 $\pm$ 0.006} & \textbf{0.208 $\pm$ 0.005} & \textbf{0.154 $\pm$ 0.005} & \textbf{0.225 $\pm$ 0.004} & \textbf{0.161 $\pm$ 0.025} & \textbf{0.136 $\pm$ 0.007} & \textbf{0.119 $\pm$ 0.025} \\
\midrule
2025-08 & SASRec & 0.046 $\pm$ 0.000 & 0.059 $\pm$ 0.000 & 0.047 $\pm$ 0.000 & 0.063 $\pm$ 0.000 & 0.048 $\pm$ 0.000 & 0.066 $\pm$ 0.000 & 0.012 $\pm$ 0.000 & 0.050 $\pm$ 0.000 & 0.022 $\pm$ 0.000 \\
 & GETNext & 0.109 $\pm$ 0.000 & 0.130 $\pm$ 0.000 & 0.112 $\pm$ 0.000 & 0.145 $\pm$ 0.000 & 0.115 $\pm$ 0.000 & 0.155 $\pm$ 0.000 & \underline{0.055 $\pm$ 0.000} & 0.099 $\pm$ 0.000 & 0.063 $\pm$ 0.000 \\
 & KBGNN & 0.109 $\pm$ 0.002 & 0.128 $\pm$ 0.001 & 0.113 $\pm$ 0.002 & 0.144 $\pm$ 0.001 & 0.115 $\pm$ 0.002 & 0.154 $\pm$ 0.001 & 0.051 $\pm$ 0.004 & 0.103 $\pm$ 0.003 & 0.067 $\pm$ 0.009 \\
 & DisenPOI & 0.117 $\pm$ 0.001 & 0.138 $\pm$ 0.004 & 0.121 $\pm$ 0.002 & 0.154 $\pm$ 0.006 & 0.123 $\pm$ 0.002 & 0.165 $\pm$ 0.007 & 0.053 $\pm$ 0.001 & 0.099 $\pm$ 0.003 & 0.060 $\pm$ 0.001 \\
 & Diff-POI & 0.118 $\pm$ 0.002 & 0.140 $\pm$ 0.003 & 0.123 $\pm$ 0.002 & 0.156 $\pm$ 0.001 & 0.124 $\pm$ 0.003 & 0.165 $\pm$ 0.003 & 0.052 $\pm$ 0.005 & \underline{0.109 $\pm$ 0.004} & 0.067 $\pm$ 0.008 \\
 & BiGSL & \underline{0.120 $\pm$ 0.003} & \underline{0.142 $\pm$ 0.001} & \underline{0.124 $\pm$ 0.002} & \underline{0.158 $\pm$ 0.002} & \underline{0.126 $\pm$ 0.002} & \underline{0.169 $\pm$ 0.001} & 0.053 $\pm$ 0.005 & 0.108 $\pm$ 0.006 & \underline{0.069 $\pm$ 0.012} \\
 & GNPR-SID & 0.110 $\pm$ 0.002 & 0.129 $\pm$ 0.003 & 0.114 $\pm$ 0.002 & 0.144 $\pm$ 0.004 & 0.116 $\pm$ 0.002 & 0.154 $\pm$ 0.004 & 0.051 $\pm$ 0.005 & 0.102 $\pm$ 0.005 & 0.066 $\pm$ 0.010 \\
 & CAGNN & 0.106 $\pm$ 0.005 & 0.126 $\pm$ 0.004 & 0.110 $\pm$ 0.004 & 0.141 $\pm$ 0.002 & 0.112 $\pm$ 0.004 & 0.152 $\pm$ 0.003 & 0.045 $\pm$ 0.007 & 0.097 $\pm$ 0.006 & 0.062 $\pm$ 0.011 \\
 & \textbf{EviRec} & \textbf{0.131 $\pm$ 0.002} & \textbf{0.151 $\pm$ 0.003} & \textbf{0.135 $\pm$ 0.001} & \textbf{0.167 $\pm$ 0.001} & \textbf{0.138 $\pm$ 0.001} & \textbf{0.179 $\pm$ 0.001} & \textbf{0.076 $\pm$ 0.005} & \textbf{0.119 $\pm$ 0.003} & \textbf{0.091 $\pm$ 0.008} \\
\midrule
2025-09 & SASRec & 0.101 $\pm$ 0.000 & 0.132 $\pm$ 0.000 & 0.106 $\pm$ 0.000 & 0.154 $\pm$ 0.000 & 0.108 $\pm$ 0.000 & 0.165 $\pm$ 0.000 & 0.049 $\pm$ 0.000 & 0.118 $\pm$ 0.000 & 0.072 $\pm$ 0.000 \\
 & GETNext & 0.308 $\pm$ 0.000 & 0.364 $\pm$ 0.000 & 0.316 $\pm$ 0.000 & 0.393 $\pm$ 0.000 & 0.320 $\pm$ 0.000 & 0.415 $\pm$ 0.000 & 0.234 $\pm$ 0.000 & 0.324 $\pm$ 0.000 & 0.289 $\pm$ 0.000 \\
 & KBGNN & 0.302 $\pm$ 0.003 & 0.359 $\pm$ 0.003 & 0.311 $\pm$ 0.002 & 0.392 $\pm$ 0.003 & 0.315 $\pm$ 0.002 & 0.412 $\pm$ 0.003 & 0.230 $\pm$ 0.003 & 0.316 $\pm$ 0.004 & 0.280 $\pm$ 0.002 \\
 & DisenPOI & \textbf{0.368 $\pm$ 0.011} & \textbf{0.428 $\pm$ 0.011} & \textbf{0.376 $\pm$ 0.011} & \textbf{0.460 $\pm$ 0.010} & \textbf{0.380 $\pm$ 0.011} & \textbf{0.477 $\pm$ 0.008} & \underline{0.308 $\pm$ 0.015} & \textbf{0.374 $\pm$ 0.011} & \underline{0.344 $\pm$ 0.017} \\
 & Diff-POI & 0.334 $\pm$ 0.002 & 0.395 $\pm$ 0.004 & 0.342 $\pm$ 0.002 & 0.428 $\pm$ 0.004 & 0.347 $\pm$ 0.002 & 0.449 $\pm$ 0.004 & 0.262 $\pm$ 0.003 & 0.345 $\pm$ 0.006 & 0.302 $\pm$ 0.006 \\
 & BiGSL & 0.340 $\pm$ 0.006 & 0.395 $\pm$ 0.006 & 0.348 $\pm$ 0.006 & 0.427 $\pm$ 0.006 & 0.352 $\pm$ 0.006 & 0.446 $\pm$ 0.005 & 0.267 $\pm$ 0.008 & 0.350 $\pm$ 0.005 & 0.313 $\pm$ 0.007 \\
 & GNPR-SID & 0.312 $\pm$ 0.010 & 0.370 $\pm$ 0.008 & 0.319 $\pm$ 0.010 & 0.400 $\pm$ 0.007 & 0.324 $\pm$ 0.009 & 0.421 $\pm$ 0.006 & 0.238 $\pm$ 0.008 & 0.325 $\pm$ 0.009 & 0.288 $\pm$ 0.011 \\
 & CAGNN & 0.293 $\pm$ 0.009 & 0.349 $\pm$ 0.010 & 0.302 $\pm$ 0.009 & 0.383 $\pm$ 0.011 & 0.306 $\pm$ 0.009 & 0.404 $\pm$ 0.009 & 0.224 $\pm$ 0.008 & 0.308 $\pm$ 0.010 & 0.271 $\pm$ 0.012 \\
 & \textbf{EviRec} & \underline{0.359 $\pm$ 0.012} & \underline{0.419 $\pm$ 0.012} & \underline{0.366 $\pm$ 0.012} & \underline{0.450 $\pm$ 0.011} & \underline{0.370 $\pm$ 0.012} & \underline{0.468 $\pm$ 0.009} & \textbf{0.326 $\pm$ 0.014} & \underline{0.370 $\pm$ 0.009} & \textbf{0.365 $\pm$ 0.012} \\
\midrule
2025-10 & SASRec & 0.052 $\pm$ 0.000 & 0.065 $\pm$ 0.000 & 0.054 $\pm$ 0.000 & 0.073 $\pm$ 0.000 & 0.055 $\pm$ 0.000 & 0.078 $\pm$ 0.000 & 0.006 $\pm$ 0.000 & 0.041 $\pm$ 0.000 & 0.010 $\pm$ 0.000 \\
 & GETNext & 0.146 $\pm$ 0.000 & 0.170 $\pm$ 0.000 & 0.150 $\pm$ 0.000 & 0.187 $\pm$ 0.000 & 0.153 $\pm$ 0.000 & 0.198 $\pm$ 0.000 & 0.049 $\pm$ 0.000 & 0.108 $\pm$ 0.000 & 0.061 $\pm$ 0.000 \\
 & KBGNN & 0.147 $\pm$ 0.001 & 0.171 $\pm$ 0.001 & 0.151 $\pm$ 0.001 & 0.186 $\pm$ 0.001 & 0.154 $\pm$ 0.001 & 0.196 $\pm$ 0.002 & 0.053 $\pm$ 0.002 & 0.109 $\pm$ 0.000 & 0.062 $\pm$ 0.000 \\
 & DisenPOI & \underline{0.164 $\pm$ 0.002} & \underline{0.187 $\pm$ 0.002} & \underline{0.168 $\pm$ 0.003} & \underline{0.203 $\pm$ 0.003} & \underline{0.170 $\pm$ 0.002} & 0.213 $\pm$ 0.003 & \underline{0.067 $\pm$ 0.003} & 0.119 $\pm$ 0.001 & 0.071 $\pm$ 0.003 \\
 & Diff-POI & 0.162 $\pm$ 0.001 & 0.185 $\pm$ 0.001 & 0.165 $\pm$ 0.001 & 0.198 $\pm$ 0.001 & 0.167 $\pm$ 0.001 & 0.208 $\pm$ 0.001 & 0.064 $\pm$ 0.004 & \underline{0.122 $\pm$ 0.003} & \underline{0.074 $\pm$ 0.004} \\
 & BiGSL & 0.160 $\pm$ 0.002 & 0.186 $\pm$ 0.003 & 0.164 $\pm$ 0.002 & 0.202 $\pm$ 0.003 & 0.166 $\pm$ 0.002 & \underline{0.213 $\pm$ 0.004} & 0.062 $\pm$ 0.002 & 0.119 $\pm$ 0.002 & 0.070 $\pm$ 0.002 \\
 & GNPR-SID & 0.147 $\pm$ 0.001 & 0.169 $\pm$ 0.001 & 0.151 $\pm$ 0.000 & 0.185 $\pm$ 0.001 & 0.153 $\pm$ 0.000 & 0.195 $\pm$ 0.001 & 0.054 $\pm$ 0.003 & 0.109 $\pm$ 0.002 & 0.061 $\pm$ 0.002 \\
 & CAGNN & 0.144 $\pm$ 0.002 & 0.168 $\pm$ 0.001 & 0.148 $\pm$ 0.002 & 0.183 $\pm$ 0.001 & 0.150 $\pm$ 0.002 & 0.192 $\pm$ 0.002 & 0.051 $\pm$ 0.002 & 0.106 $\pm$ 0.001 & 0.061 $\pm$ 0.001 \\
 & \textbf{EviRec} & \textbf{0.165 $\pm$ 0.005} & \textbf{0.191 $\pm$ 0.004} & \textbf{0.170 $\pm$ 0.005} & \textbf{0.207 $\pm$ 0.006} & \textbf{0.172 $\pm$ 0.005} & \textbf{0.218 $\pm$ 0.006} & \textbf{0.080 $\pm$ 0.003} & \textbf{0.124 $\pm$ 0.003} & \textbf{0.085 $\pm$ 0.004} \\
\midrule
2025-11 & SASRec & 0.032 $\pm$ 0.000 & 0.040 $\pm$ 0.000 & 0.033 $\pm$ 0.000 & 0.046 $\pm$ 0.000 & 0.034 $\pm$ 0.000 & 0.050 $\pm$ 0.000 & 0.009 $\pm$ 0.000 & 0.034 $\pm$ 0.000 & 0.018 $\pm$ 0.000 \\
 & GETNext & 0.116 $\pm$ 0.000 & 0.139 $\pm$ 0.000 & 0.120 $\pm$ 0.000 & 0.154 $\pm$ 0.000 & 0.122 $\pm$ 0.000 & 0.164 $\pm$ 0.000 & 0.061 $\pm$ 0.000 & 0.111 $\pm$ 0.000 & 0.085 $\pm$ 0.000 \\
 & KBGNN & 0.120 $\pm$ 0.006 & 0.142 $\pm$ 0.005 & 0.124 $\pm$ 0.006 & 0.158 $\pm$ 0.006 & 0.126 $\pm$ 0.006 & 0.167 $\pm$ 0.005 & 0.063 $\pm$ 0.004 & 0.117 $\pm$ 0.006 & 0.091 $\pm$ 0.004 \\
 & DisenPOI & 0.136 $\pm$ 0.002 & 0.160 $\pm$ 0.002 & 0.140 $\pm$ 0.002 & 0.174 $\pm$ 0.001 & 0.142 $\pm$ 0.002 & 0.184 $\pm$ 0.002 & 0.070 $\pm$ 0.004 & 0.125 $\pm$ 0.002 & 0.095 $\pm$ 0.008 \\
 & Diff-POI & \underline{0.137 $\pm$ 0.001} & \underline{0.160 $\pm$ 0.001} & \underline{0.140 $\pm$ 0.001} & 0.175 $\pm$ 0.002 & \underline{0.142 $\pm$ 0.001} & 0.184 $\pm$ 0.002 & \underline{0.074 $\pm$ 0.002} & \underline{0.129 $\pm$ 0.002} & \underline{0.103 $\pm$ 0.003} \\
 & BiGSL & 0.136 $\pm$ 0.003 & 0.160 $\pm$ 0.004 & 0.140 $\pm$ 0.003 & \underline{0.176 $\pm$ 0.003} & 0.142 $\pm$ 0.003 & \underline{0.186 $\pm$ 0.002} & 0.072 $\pm$ 0.005 & 0.129 $\pm$ 0.003 & 0.099 $\pm$ 0.002 \\
 & GNPR-SID & 0.121 $\pm$ 0.004 & 0.144 $\pm$ 0.005 & 0.125 $\pm$ 0.004 & 0.158 $\pm$ 0.006 & 0.127 $\pm$ 0.004 & 0.168 $\pm$ 0.004 & 0.064 $\pm$ 0.004 & 0.118 $\pm$ 0.004 & 0.092 $\pm$ 0.005 \\
 & CAGNN & 0.120 $\pm$ 0.004 & 0.143 $\pm$ 0.004 & 0.124 $\pm$ 0.004 & 0.156 $\pm$ 0.005 & 0.126 $\pm$ 0.004 & 0.165 $\pm$ 0.003 & 0.064 $\pm$ 0.002 & 0.119 $\pm$ 0.004 & 0.094 $\pm$ 0.006 \\
 & \textbf{EviRec} & \textbf{0.144 $\pm$ 0.003} & \textbf{0.170 $\pm$ 0.004} & \textbf{0.148 $\pm$ 0.002} & \textbf{0.185 $\pm$ 0.004} & \textbf{0.150 $\pm$ 0.002} & \textbf{0.196 $\pm$ 0.002} & \textbf{0.093 $\pm$ 0.000} & \textbf{0.139 $\pm$ 0.003} & \textbf{0.124 $\pm$ 0.004} \\
\midrule
2025-12 & SASRec & 0.065 $\pm$ 0.000 & 0.079 $\pm$ 0.000 & 0.067 $\pm$ 0.000 & 0.087 $\pm$ 0.000 & 0.069 $\pm$ 0.000 & 0.095 $\pm$ 0.000 & 0.000 $\pm$ 0.000 & 0.062 $\pm$ 0.000 & 0.000 $\pm$ 0.000 \\
 & GETNext & 0.198 $\pm$ 0.000 & 0.225 $\pm$ 0.000 & 0.203 $\pm$ 0.000 & 0.246 $\pm$ 0.000 & 0.205 $\pm$ 0.000 & 0.256 $\pm$ 0.000 & 0.007 $\pm$ 0.000 & 0.145 $\pm$ 0.000 & \underline{0.016 $\pm$ 0.000} \\
 & KBGNN & 0.201 $\pm$ 0.004 & 0.229 $\pm$ 0.005 & 0.206 $\pm$ 0.004 & 0.248 $\pm$ 0.006 & 0.208 $\pm$ 0.004 & 0.261 $\pm$ 0.007 & 0.010 $\pm$ 0.006 & 0.142 $\pm$ 0.003 & 0.014 $\pm$ 0.003 \\
 & DisenPOI & \underline{0.219 $\pm$ 0.003} & \underline{0.249 $\pm$ 0.004} & \underline{0.223 $\pm$ 0.002} & \underline{0.267 $\pm$ 0.002} & \underline{0.225 $\pm$ 0.002} & 0.277 $\pm$ 0.002 & \underline{0.019 $\pm$ 0.008} & 0.147 $\pm$ 0.002 & 0.012 $\pm$ 0.009 \\
 & Diff-POI & 0.217 $\pm$ 0.003 & 0.249 $\pm$ 0.002 & 0.221 $\pm$ 0.002 & 0.266 $\pm$ 0.003 & 0.224 $\pm$ 0.002 & \underline{0.279 $\pm$ 0.002} & 0.015 $\pm$ 0.004 & \underline{0.158 $\pm$ 0.006} & 0.016 $\pm$ 0.005 \\
 & BiGSL & 0.217 $\pm$ 0.006 & 0.248 $\pm$ 0.008 & 0.221 $\pm$ 0.006 & 0.266 $\pm$ 0.008 & 0.224 $\pm$ 0.005 & 0.278 $\pm$ 0.003 & 0.012 $\pm$ 0.006 & 0.150 $\pm$ 0.001 & 0.014 $\pm$ 0.008 \\
 & GNPR-SID & 0.202 $\pm$ 0.002 & 0.232 $\pm$ 0.004 & 0.207 $\pm$ 0.001 & 0.251 $\pm$ 0.006 & 0.209 $\pm$ 0.001 & 0.260 $\pm$ 0.005 & 0.014 $\pm$ 0.008 & 0.143 $\pm$ 0.005 & 0.016 $\pm$ 0.000 \\
 & CAGNN & 0.199 $\pm$ 0.005 & 0.229 $\pm$ 0.007 & 0.203 $\pm$ 0.005 & 0.245 $\pm$ 0.008 & 0.206 $\pm$ 0.005 & 0.257 $\pm$ 0.009 & 0.012 $\pm$ 0.007 & 0.134 $\pm$ 0.005 & 0.016 $\pm$ 0.000 \\
 & \textbf{EviRec} & \textbf{0.225 $\pm$ 0.002} & \textbf{0.257 $\pm$ 0.003} & \textbf{0.230 $\pm$ 0.002} & \textbf{0.275 $\pm$ 0.005} & \textbf{0.232 $\pm$ 0.002} & \textbf{0.287 $\pm$ 0.003} & \textbf{0.027 $\pm$ 0.015} & \textbf{0.159 $\pm$ 0.004} & \textbf{0.024 $\pm$ 0.006} \\
\bottomrule
\end{tabular}}
\end{table*}